\documentclass[]{elsarticle}
\pdfoutput=1

\usepackage[margin=1.3in]{geometry}
\usepackage{wrapfig}
\usepackage{lscape}
\usepackage{rotating}
\usepackage{epstopdf}
\usepackage{float}
\usepackage{array}
\usepackage{algpseudocode}
\usepackage{algorithm}
\usepackage{multirow}
\usepackage[english]{babel} 
\usepackage{amsmath,amsfonts,amsthm,amssymb} 
\usepackage{url}
\usepackage{booktabs}
\usepackage{bm}
\usepackage{xfrac}
\usepackage{stfloats}
\usepackage{fixltx2e}
\usepackage{subfig}
\usepackage{xcolor}
\usepackage[export]{adjustbox}

\makeatletter
\algrenewcommand\ALG@beginalgorithmic{\footnotesize}
\makeatother
\usepackage{tikz}
\usetikzlibrary{decorations.pathreplacing,calc}
\newcommand{\tikzmark}[1]{\tikz[overlay,remember picture] \node (#1) {};}

\newcommand*{\AddNote}[4]{%
    \begin{tikzpicture}[overlay, remember picture]
        \draw [decoration={brace,amplitude=0.5em},decorate,thick,black]
            ($(#3)!(#1.north)!($(#3)-(0,1)$)$) --  
            ($(#3)!(#2.south)!($(#3)-(0,1)$)$)
                node [align=center, text width=2.5cm, pos=0.5, anchor=west] {#4};
    \end{tikzpicture}
}%

\newcommand\Train{\mathcal{T}}

\newcommand\Centr{\mathcal{R}}

\makeatletter
\def\ps@pprintTitle{%
 \let\@oddhead\@empty
 \let\@evenhead\@empty
 \def\@oddfoot{}%
 \let\@evenfoot\@oddfoot}
\makeatother

\begin{document}

\algnewcommand{\algorithmicgoto}{\textbf{go to}}%
\algnewcommand{\Goto}[1]{\algorithmicgoto~\ref{#1}}%
\algnewcommand\algorithmicinput{\textbf{Input:}}
\algnewcommand\Input{\item[\algorithmicinput]}
\algnewcommand\algorithmicoutput{\textbf{Output:}}
\algnewcommand\Output{\item[\algorithmicoutput]}

\begin{frontmatter}

\title{Online local pool generation for dynamic\\classifier selection: an extended version}



\author[mymainaddress]{Mariana A. Souza\corref{mycorrespondingauthor}}
\cortext[mycorrespondingauthor]{Corresponding author}
\ead{mas2@cin.ufpe.br}
\author[mymainaddress]{George D. C. Cavalcanti}
\ead{gdcc@cin.ufpe.br}

\author[mysecondaryaddress]{Rafael M. O. Cruz}
\ead{cruz@livia.etsmtl.ca}
\author[mysecondaryaddress]{Robert Sabourin}
\ead{robert.sabourin@etsmtl.ca}

\address[mymainaddress]{Centro de Inform\'{a}tica, Universidade Federal de Pernambuco, Recife, PE, Brazil}
\address[mysecondaryaddress]{\'{E}cole de Technologie Sup\'{e}rieure, Universit\'{e} du Qu\'{e}bec, Montreal, Quebec, Canada}

\begin{abstract}
Dynamic Classifier Selection (DCS) techniques have difficulty in selecting the most competent classifier in a pool, even when its presence is assured. Since the DCS techniques rely only on local data to estimate a classifier’s competence, the manner in which the pool is generated could affect the choice of the best classifier for a given instance. That is, the global perspective in which pools are generated may not help the DCS techniques in selecting a competent classifier for instances that are likely to be misclassified. Thus, it is proposed in this work an online pool generation method that produces a locally accurate pool for test samples in difficult regions of the feature space. The difficulty of a given area is determined by the estimated classification difficulty of the instances in it. That way, by using classifiers that were generated in a local scope, it could be easier for the DCS techniques to select the best one for those instances they would most probably misclassify. For the query samples surrounded by easy instances, a simple nearest neighbors rule is used in the proposed method. 
In the extended version of this work, a deep analysis on the correlation between instance hardness and the performance of DCS techniques is presented. 
An instance hardness measure that conveys the degree of local class overlap near a given sample is then used to identify in which cases the local pool is used in the proposed scheme. 
Experimental results show that the DCS techniques were more able to select the most competent classifier for difficult instances when using the proposed local pool than when using a globally generated pool. Moreover, the proposed technique yielded significantly greater recognition rates in comparison to a Bagging-generated pool and two other global generation schemes for all DCS techniques evaluated. The performance of the proposed technique was also significantly superior to three state-of-the-art classification models and was statistically equivalent to five of them. 
Furthermore, an extended analysis on the computational complexity of the proposed technique and of several DS techniques is presented in this version. 
We also provide the implementation of the proposed technique using the DESLib library on GitHub.
\end{abstract}


\end{frontmatter}


\section{Introduction}
\label{sec:intro}


\par Multiple Classifier Systems (MCS) aim to improve the overall performance of a pattern recognition system by combining numerous base classifiers \cite{surveyMCS,onCombin, kuncheva}. 
An MCS contains three phases \cite{DESsurvey}: (1) Generation, (2) Selection and (3) Integration. 
In the first phase, a pool of classifiers is generated using the training data. 
In the second phase, a non-empty subset of classifiers from the pool is selected to perform the classification task. 
In the third and last phase, the selected classifiers' predictions are combined to form the final system's output.  
There are two possible approaches in the Selection phase: Static Selection (SS), in which the same set of classifiers is used to label all unknown instances, or Dynamic Selection (DS), which selects certain classifiers from the pool according to each query sample. 

\par The DS techniques, which have been shown to outperform static ensembles, specially on ill-defined problems \cite{DESsurvey,metades}, are based on the idea that the classifiers in the pool are individually competent in different regions of the feature space. 
The aim of the selection scheme is, then, to choose the classifier(s) that is(are) best fit, according to some criterion, for classifying each unknown instance in particular \cite{DESsurvey}. 
The amount of classifiers singled out to label a given sample separates the DS schemes in two groups \cite{dinamSel}: Dynamic Classifier Selection (DCS) techniques, in which the classifier with highest estimated competence in the pool is selected, and Dynamic Ensemble Selection (DES) schemes, in which a locally accurate subset of classifiers from the pool is chosen and combined to label the test sample.

\par In the context of DCS, the Oracle \cite{oracle} can be defined as an abstract model that mimics the perfect selection scheme: it always selects the classifier that correctly labels a given instance, if the pool contains such classifier. 
Thus, the Oracle accuracy rate is the theoretical limit for DCS \looseness=-1 techniques. 

\par The behavior of the Oracle regarding pool generation for DCS techniques was characterized in a previous work \cite{mariana}. 
It was shown that even though the presence of one competent classifier was assured for a given instance, the DCS techniques still struggled to select it. 
This analysis was done using a pool generation method that guarantees an Oracle accuracy rate of 100\% on the training set. 
It was reasoned that the nature of the Oracle makes it not very well suited to guide the generation of a pool of classifiers for DCS since the model is performed globally, while DCS techniques use only local data to select the most competent classifier for each instance. 
Thus, the difference in perspectives between the generation and the selection may hinder the DCS techniques in the selection of a competent classifier, even when the latter is guaranteed to be in the pool. 

\par In addition to that, most works regarding DS use classical generation methods, which were designed for static ensembles \cite{cruz2017dynamic} and therefore do not take into account the regional aspect of the competence estimation performed by the DS techniques. 
Thus, since local information is not considered during the generation process, the presence of local experts is not guaranteed in the \looseness=-1 final pool. 

\par Based on these observations, it is proposed in this work, which is an extended version of \cite{SOUZA2019132}, an online pool generation method which attempts to explore the Oracle's properties on a local scope. 
Since the Oracle and DCS techniques view the problem from different perspectives, using the Oracle model in a local setting to match these perspectives may help the DCS techniques in the choice of the most competent classifier for a given instance. 
This work focus only on DCS techniques since their relationship to the Oracle was already characterized in \cite{mariana}, and so the results can be further analyzed based on certain aspects presented in the previous work. 

\par Thus, the main idea is to use the Oracle model to guide the generation of a pool specialized on the local region where a given unknown sample is, if that region is deemed difficult. 
In this context, a region is considered \textit{difficult} if it contains an instance likely to be misclassified, as indicated by an instance hardness measure. 
Therefore, if a query sample is located in a difficult region of the feature space, a local pool (LP) is generated on the fly so that its classifiers fully cover the surrounding area of that specific instance. 
Otherwise, a simple k-Nearest Neighbors (k-NN) rule is used to label the query sample, since the classification task is less complex.
Hence, whenever an unknown sample is located in a difficult region, the proposed method uses Oracle information in that area to generate locally accurate classifiers for that instance, in hopes that the best classifier among them will be more easily selected by a DCS technique than if the classifiers were generated with a \looseness=-1 global perspective.


\par To the best of our knowledge, there is no ensemble method designed to generate local experts for dynamic selection techniques. 
However, a local learning algorithm with a similar strategy to that of the proposed technique is presented in \cite{bottou1992local}. 
The learning algorithm consists of generating a linear classifier for each unknown instance using its surrounding training samples, and then labelling that instance with it. 
This model was used to analyze the trade-off between capacity and locality of the learning algorithms and its impact on their recognition rates. 
Although the learning algorithm provides a local perspective on the classification problem, its concept was not used in the context of producing a pool of locally accurate classifiers for DS techniques.

\par Other related works, such as the Mixture of Random Prototype-based Local Experts \cite{armano2010mixture} and the Forest of Local Trees \cite{armano2017building} techniques, explore the divide-to-conquer approach of MCS by locally training their base classifiers in different regions of the feature space and weighting the classifiers' votes based on the distance between the query sample and their assigned region. 
As opposed to these works, in which the pool generation is paired to a selection based on dynamic distance weighting, our approach consists of producing on the fly a locally accurate pool to be coupled with a DCS technique. 
Furthermore, the generation process of these approaches do not guarantee the presence of local experts in the vicinity of each borderline unknown sample, as the proposed \looseness=-1 method does. 

\par Thus, with our proposed approach, we aim to find out in this work whether the presence of locally generated pools is advantageous in DCS context. 
The research questions we intend to answer are: (1) does the use of locally generated pools aid the DCS techniques in selecting the best classifier for a given instance?, and (2) do the recognition rates improve as a result of this?.
To that end, the performances of the proposed scheme and of different ensemble methods that yield globally-generated pools are assessed using DCS techniques over 20 public datasets, and the results compared and analyzed. 
A comparative study with several state-of-the-art classification models is also performed afterwards. 


\par This work is organized as follows: in Section \ref{sec:ih-analysis}, an analysis on instance hardness for DCS techniques is performed in order to observe the correlation between an instance hardness measure and the mistakes made by these techniques. 
Once this relationship is established, Section \ref{sec:prop-tech} presents the proposed generation method, which explores the instance hardness information obtained in the previous section. 
In Section \ref{sec:exp} the proposed method is evaluated, and it is analyzed whether the use of specialist subpools in difficult regions is beneficial for DCS techniques. 
A comparative study with state-of-the-art classification models is also performed in Section \ref{sec:exp}. 
Lastly, in Section \ref{sec:concl} the results are summarized and future works are suggested.

\section{Instance Hardness Analysis}
\label{sec:ih-analysis}

\par Hardness is an aspect inherent to a problem that hinders a classifier, or a set of classifiers, in the classification task. 
\textit{Instance hardness} is then a characteristic of a sample's problem that conveys the likelihood of such sample being mislabelled by a classifier \cite{smith2014instance}. 
Hardness measures attempt to quantify this characteristic based on different sources of difficulty associated with data, as well as provide insights as to why some instances or problems are difficult for most learning algorithms. 
Many data hardness measures were proposed and also used to improve a vast number of methods in the literature \cite{DONG20031215,smith2016comparative,singh2003multiresolution}. 






\par In \cite{ho2002complexity}, the authors introduce a set of hardness measures obtained over the entire training set. 
They also identify aspects that lead to a problem, as a whole, being difficult for a classifier. 
The authors in \cite{Garcia2015108} propose a set of hardness measures, also over the whole dataset, and use them, together with the ones from \cite{ho2002complexity}, to identify and remove noisy data in the training set. 

\par In addition to characterizing the hardness of an entire set, efforts have been made to estimate the difficulty of classifying each individual instance from a dataset. 
In \cite{smith2011improving}, the authors propose several instance hardness measures and use a subset of them in the construction of noise filter which removes potentially noisy instances among the hard ones. 
A hardness analysis on an instance level was done in \cite{smith2014instance}, in which the authors identify the most influential causes for an instance to be hard for many diverse classification models. 
They also introduce more instance hardness measures and show the correlation between them and the misclassification of the classifiers analysed. 
Moreover, they suggest the integration of the error information of these classifiers in two different scenarios: in the training of a neural network, so that the weight of the hard instances are smaller, and in a noise filter, based on the same idea as the previous one. 
In \cite{de2014cost}, the authors propose two instance hardness measures that take into account misclassification costs. 
These measures are further used to define a measure of similarity between algorithms. 

\par Although in \cite{smith2011improving} a set of classifiers was used to evaluate the correlation between the hardness measures and the instance hardness itself, the authors did not investigate it for DCS techniques. 
Though an analysis on instance hardness regarding DS techniques was performed in \cite{cruz2017ipta}, its focus was on the comparison between these techniques and the k-NN classifier. 
Thus, an analysis on instance hardness in DCS context is done in this section. 
The purpose of such analysis is to understand the correlation between instance hardness measures and the errors made by DCS techniques, in order to identify in which cases the DCS techniques fail to choose a competent classifier for a given instance. 
This information will later be used to generate subpools specialized in the difficult regions of the training set.

\par The chosen instance hardness measure to be analyzed is the k-Disagreeing Neighbors (kDN) \cite{smith2014instance}, which is defined in Equation \ref{eq:kdn}, where $\mathbf{x_{i}}$ is the instance being evaluated, $\Train$ is the 
dataset that contains it, $kNN()$ is the k-Nearest Neighbors (k-NN) rule and $k_{h}$ is the neighborhood size of the hardness measure.
The kDN measure is the percentage of instances in an example's neighborhood that do not share the same label as itself. 
Therefore, a high kDN value means the instance is in a local overlap region, making it harder to label it. 
The reason for using this measure is because it denotes the most relevant source of instance hardness (overlap), according to \cite{smith2014instance}. 
Moreover, the kDN measure was the most correlated with instance hardness according to the same article. 

\begin{equation} \label{eq:kdn}
\centering
kDN(\mathbf{x_{i}},\Train,k_{h}) = \frac{\vert { \mathbf{x_{j}}: \mathbf{x_{j}} \in kNN(\mathbf{x_{i}},\Train,k_{h}) \wedge label(\mathbf{x_{j}}) \neq label(\mathbf{x_{i})}} \vert }{k_{h}} 
\end{equation}


\par The pool generation technique used in this analysis was the Self-generating Hyperplanes (SGH) method \cite{mariana}, a simple generation scheme which yielded a similar performance as Bagging \cite{bagging} for most DCS techniques. 
The SGH generation method is described in Algorithm~\ref{alg:sgh}. 
The input to the SGH method is only the training set $\Train$, and its output is the generated pool of classifiers ($C$). 
In each iteration (Step 3 to Step 15), the centroids of all classes in $\Train$ are obtained in Step 4 and stored in $\Centr$. 
The two centroids in $\Centr$ most distant from each other, $\mathbf{r_{i}}$ and $\mathbf{r_{j}}$, are selected in Step 5. 
Then, a hyperplane $c_{m}$ is placed between $\mathbf{r_{i}}$ and $\mathbf{r_{j}}$, dividing both points halfway from each other. 
The two-class linear classifier $c_{m}$ is then tested over the training set, and the instances it correctly labels are removed from $\Train$ (Step 7 to Step 12). 
Then, $c_{m}$ is added to $C$ in Step 13, and the loop is repeated until $\Train$ is completely empty. 
That is, the SGH method only stops generating hyperplanes when all training instances are correctly labelled by at least one classifier in $C$, i.e., the Oracle accuracy rate for the training set is 100\%.

\begin{algorithm}[!htb]
\centering
\footnotesize
\begin{algorithmic}[1]
\Input $\Train = \{\mathbf{x_{1}}, \mathbf{x_{2}}, ...,\mathbf{x_{N}}\}$ \Comment{Training dataset}
\Output $C$ \Comment{Final pool}
\State $C \gets \{\}$ \Comment{Pool initially empty}
\State $m \gets 1$  \Comment{Classifier count}
\While {$\Train \neq \{\}$ } 
	\State $ \Centr \gets getCentroids(\Train)$ \Comment{Calculate each class' centroid}
	\State $ \mathbf{r_{i}}, \mathbf{r_{j}} \gets selectCentroids(\Centr)$ \Comment{Select the most distant centroids}
	\State $ c_{m} \gets placeHyperplane(\mathbf{r_{i}},\mathbf{r_{j}})$ \Comment{Generate hyperplane between centroids $\mathbf{r_{i}}$ and $\mathbf{r_{j}}$}
	\For {every $\mathbf{x_{n}}$ in $\Train$} 
		\State $ \omega \gets c_{m}(\mathbf{x_{n}})$ \Comment{Test $c_{m}$ over training instance}
		\If{$ \omega = y_{n}$} 		
			\State $\Train \gets \Train - \{\mathbf{x_{n}}\}$ \Comment{Remove from $\Train$ correctly classified instance}
		\EndIf
	\EndFor
	\State $ C \gets C \cup \{c_{m}\} $ \Comment{Add $c_{m}$ to pool}
	\State $m \gets m + 1$
\EndWhile \\
\Return $C$
\end{algorithmic}
\caption{General procedure of the Self-generating Hyperplanes (SGH) method.}
\label{alg:sgh}
\end{algorithm}

\par In order to evaluate the correlation between the instance hardness measures and the accuracy of the DCS techniques, the hardness of each instance was computed using the entire dataset. 
Afterwards, the accuracy of the DCS techniques were obtained using 10 times 10-fold cross validation. 
The previous knowledge regarding each instance's hardness is then used to draw a relationship between this measure and the frequency at which the DCS techniques misclassifies it. 
That way, an evaluation of the correlation between the kDN measure and the error rate of the DCS techniques can be performed. 

\par The datasets used in this analysis are shown in Table \ref{table:datasets}. 
All of them are public datasets. 
Eleven from the UCI machine learning repository \cite{uci}, three from the Ludmila Kuncheva Collection \cite{lkc} of real medical data, three from the STATLOG project \cite{statlog}, two from the Knowledge Extraction based on Evolutionary Learning (KEEL) repository \cite{keel} and one from the Enhanced Learning for Evolutive Neural Architectures (ELENA) project \cite{elena}.
The DCS techniques used in this analysis were Overall Local Accuracy (OLA) \cite{ola} and Local Class Accuracy (LCA) \cite{ola}, which were the two best performing DCS techniques in a recent survey on dynamic selection of classifiers \cite{cruz2017dynamic}. 
 
\begin{table}[!htbp]
\centering
\caption{Main characteristics of the datasets used in the experiments.}
\label{table:datasets}
\scalebox{0.7}{
\begin{tabular}{|c|ccccc|}
\hline
Dataset & \begin{tabular}[c]{@{}c@{}}No. of\\ Instances\end{tabular} & \begin{tabular}[c]{@{}c@{}}No. of\\ Features\end{tabular} & \begin{tabular}[c]{@{}c@{}}No. of\\ Classes\end{tabular} & Class Sizes & Source \\ \hline
Adult					& 48842 & 14 & 2 & 383;307	& UCI \\
Blood Transfusion		& 748 & 4 & 2 	 & 570;178	& UCI \\
Cardiotocography (CTG)	& 2126 & 21 & 3  & 1655;295;176	& UCI \\
Steel Plate Faults		& 1941 & 27 & 7	 & 158;190;391;72;55;402;673	& UCI \\
German credit 			& 1000 & 20 & 2  & 700;300	& STATLOG \\
Glass			 		& 214 & 9 & 6 	 & 70;76;17;13;9;29	& UCI \\
Haberman's Survival		& 306 & 3 & 2 	 & 225;81	& UCI \\
Heart					& 270 & 13 & 2 	 & 150;120	& STATLOG \\ 
Ionosphere				& 315 & 34 & 2 	 & 126;225	& UCI \\
Laryngeal1				& 213 & 16 & 2 	 & 81;132	& LKC \\
Laryngeal3				& 353 & 16 & 3 	 & 53;218;82	& LKC \\
Liver Disorders			& 345 & 6 & 2 	 & 145;200	& UCI \\
Mammographic 			& 961 & 5 & 2 	 & 427;403	& KEEL \\
Monk2					& 4322 & 6 & 2 	 & 204;228	& KEEL \\
Phoneme					& 5404 & 6 & 2   & 3818;1586	& ELENA \\
Pima					& 768 & 8 & 2 	 & 500;268	& UCI \\
Sonar					& 208 & 60 & 2 	 & 97;111	& UCI \\
Vehicle					& 846 & 18 & 4 	 & 199;212;217;218	& STATLOG \\
Vertebral Column		& 310 & 6 & 2 	 & 204;96	& UCI \\
Weaning					& 302 & 17 & 2 	 & 151;151	& LKC \\ \hline
\end{tabular}}
\end{table}

\par Firstly, the pool generated by the SGH method was tested with OLA and LCA over the datasets from Table \ref{table:datasets}. Both DCS techniques had its neighborhood size $k_{s}$ varied in the set $\{3,5,7,9,11,13,15\}$ and the 
techniques' individual misclassifications for each value of $k_{s}$ were obtained. 
Afterwards, each instance had its hardness estimated using kDN, with the neighborhood size of the hardness measure $k_{h}$ also varying in the set $\{3,5,7,9,11,13,15\}$, and grouped according to four intervals: $kDN \in [0,0.25]$, $kDN \in (0.25,0.50]$, $kDN \in (0.50,0.75]$ and $kDN \in (0.75,1]$. 
Since seven hardness estimates were obtained for each instance, one for each $k_{h}$, an instance can belong to different groups depending on the value of $k_{h}$. 
Then, the error rate of the DCS techniques with respect to each group was calculated for each pair $(k_{s},k_{h})$. 
The purpose of evaluating the error rate by kDN range with varying neighborhood sizes $k_{s}$ and $k_{h}$ is to observe how the difference in the regions the DCS techniques and the hardness measure operate impacts the behavior of the error rate for a given kDN value.

\begin{figure*}[!htbp]
		\centering
		\centerline{
		\subfloat[]{
			\includegraphics[scale=0.6]{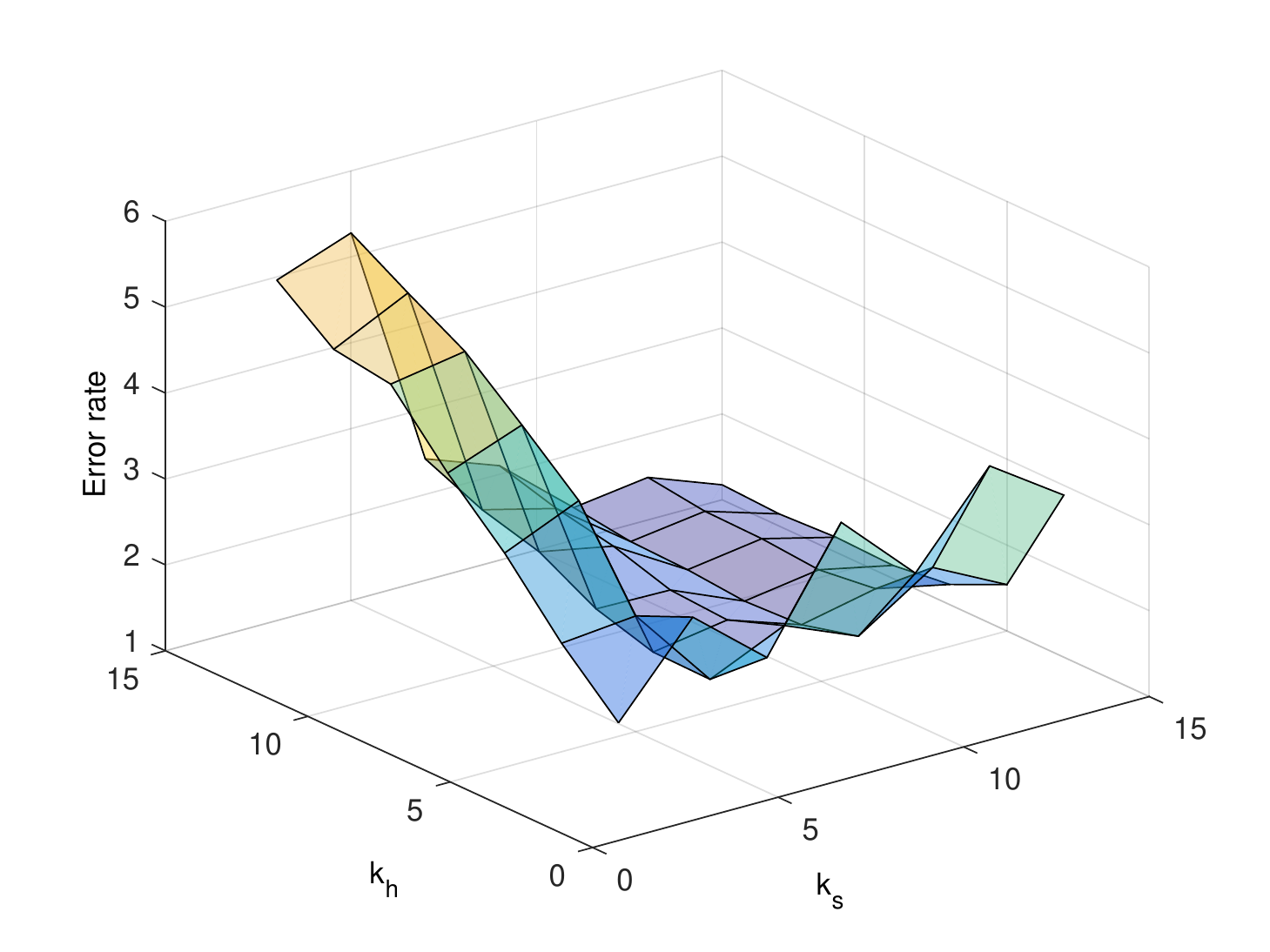}
		}
		\subfloat[]{
			\includegraphics[scale=0.6]{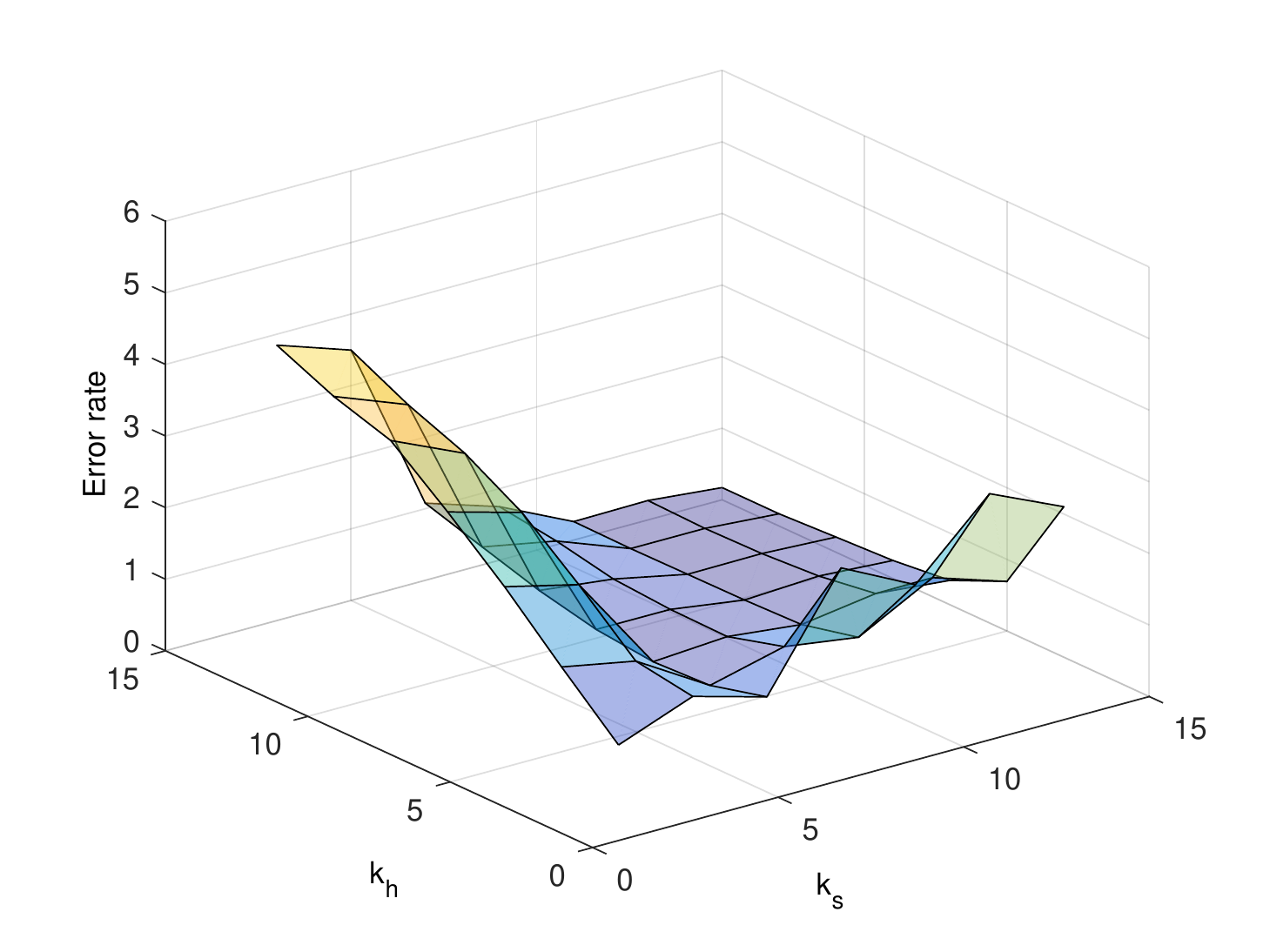}
		}}
		\caption{Mean error rate of (a) OLA and (b) LCA for instances with $kDN \in [0,0.25]$, for all datasets from Table \ref{table:datasets}. 
		$k_{h}$ and $k_{s}$ are the neighborhood sizes of kDN and the DCS technique, respectively.}
		\label{fig:kk-0-25}
\end{figure*}

\par Figure \ref{fig:kk-0-25} shows the mean error rate of OLA and LCA by neighborhood sizes $k_{h}$ and $k_{s}$ for the instances with $kDN \in [0,0.25]$, that is, for the easiest instances in the datasets. 
It is expected, then, that the error rate of the DCS techniques for those instances is very low. 
It can be observed that this is the case when $k_{h}$ and $k_{s}$ have similar values. 
More specifically, the error rate for the cases in which $k_{h} = k_{s}$ are almost identical, which may indicate a certain 
constancy in the way the kDN measure correlates with the DCS techniques' error rates. 
On the other hand, the error rate significantly increases as the difference between the neighborhood sizes increase, which 
suggests the estimation of hardness in much different regions of competence may hinder the hardness measure's ability of singling out the instances the DCS techniques find it easy or hard to classify.

\begin{figure*}[!htbp]
		\centering
		\centerline{
		\subfloat[]{
			\includegraphics[scale=0.6]{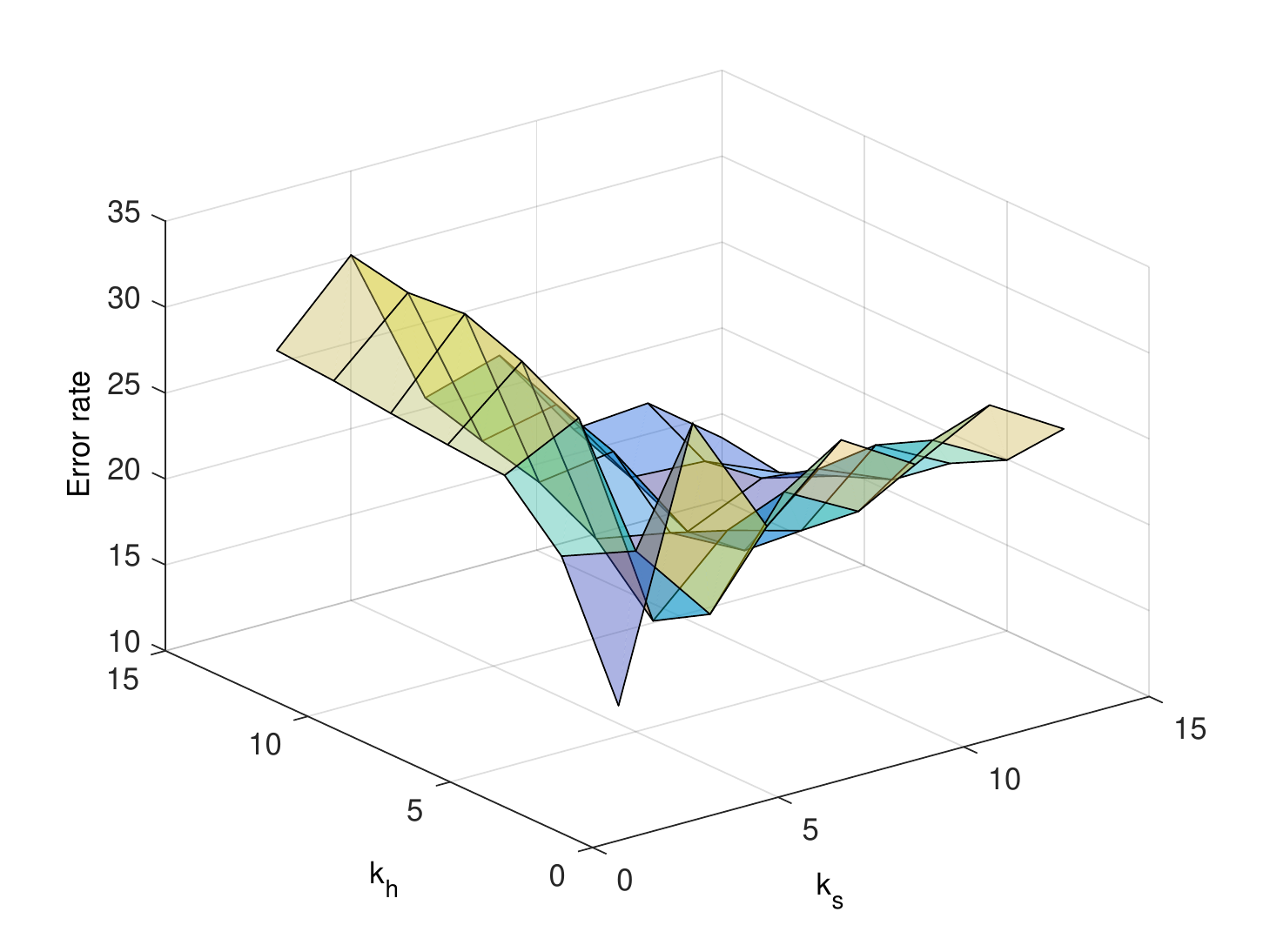}
		}
		\subfloat[]{
			\includegraphics[scale=0.6]{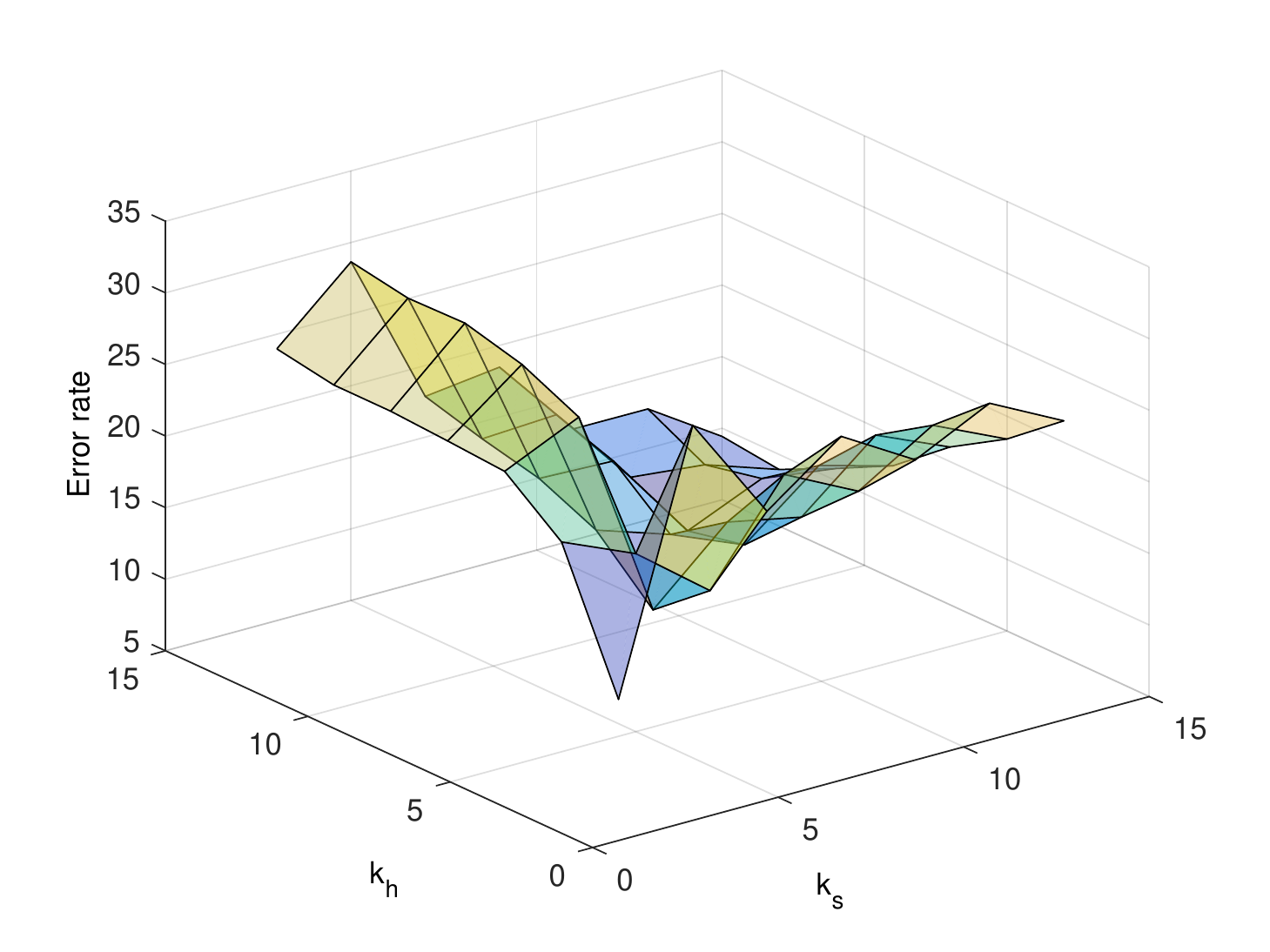}
		}}
		\caption{Mean error rate of (a) OLA and (b) LCA for instances with $kDN \in (0.25,0.50]$, for all datasets from Table \ref{table:datasets}. 
		$k_{h}$ and $k_{s}$ are the neighborhood sizes of kDN and the DCS technique, respectively.}
		\label{fig:kk-25-50}
\end{figure*}

\par The same behavior can be observed in Figure \ref{fig:kk-25-50}, where the mean error rate of OLA and LCA for instances with $kDN \in (0.25,0.50]$ is shown by neighborhood sizes $k_{h}$ and $k_{s}$. 
Since the instances within this kDN interval have most neighbors of the same class, the error rate of the DCS techniques should not be too high, and yet the more distant $k_{h}$ and $k_{s}$ become, the higher the mean error rate. 
On the other hand, the closer the neighborhood sizes, the more consistent with the hardness measure's interpretation the error rate becomes. 
Moreover, with the exception of the case in which $k_{h} = k_{s} = 5$, the error rates of the other cases with equal neighborhood sizes were very similar for both DCS techniques, further suggesting that having the same region of competence preserves the relationship between the hardness measure and the DCS techniques' misclassification rates.

\begin{figure*}[!htbp]
		\centering
		\centerline{
		\subfloat[]{
			\includegraphics[scale=0.6]{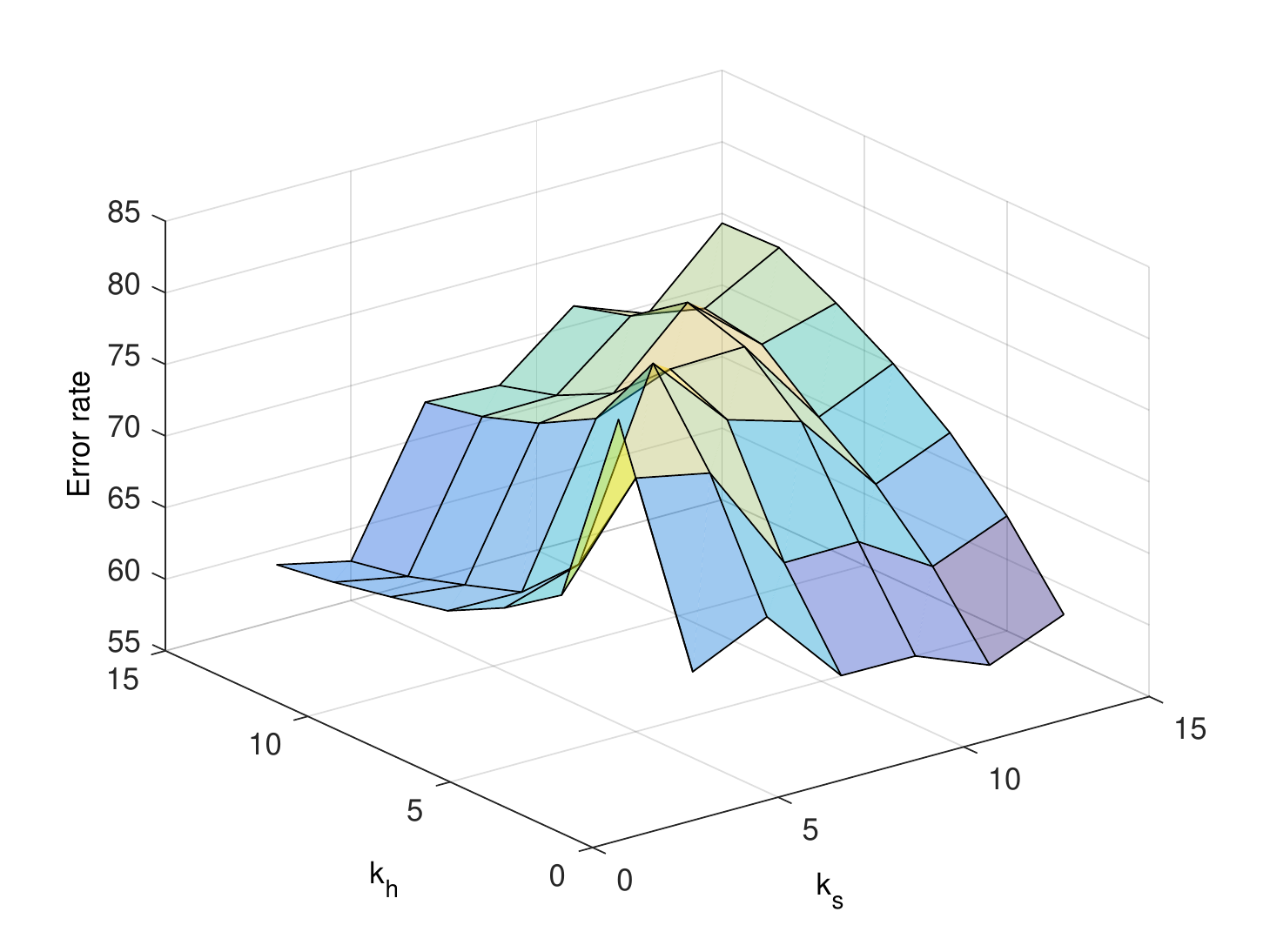}
		}
		\subfloat[]{
			\includegraphics[scale=0.6]{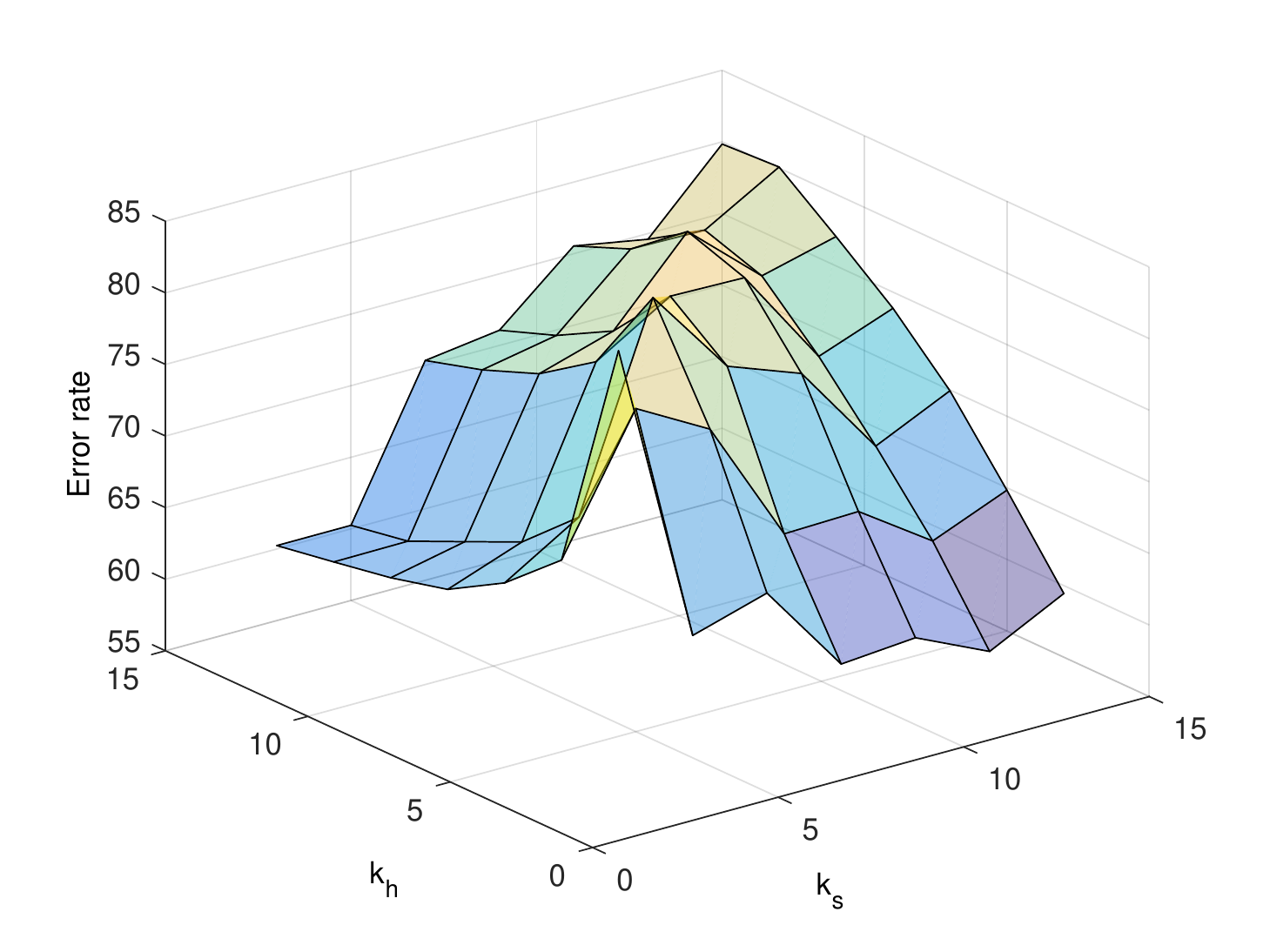}
		}}
		\caption{Mean error rate of (a) OLA and (b) LCA for instances with $kDN \in (0.50,0.75]$, for all datasets from Table \ref{table:datasets}. 
		$k_{h}$ and $k_{s}$ are the neighborhood sizes of kDN and the DCS technique, respectively.}
		\label{fig:kk-50-75}
\end{figure*}

\par For the instances with $kDN \in (0.50,0.75]$, which have more neighbors of a different class, the mean error rates of OLA and LCA are expected to be reasonably high, as Figure \ref{fig:kk-50-75} shows. 
Again, the closer the neighborhood sizes are, the more coherent the error rate is to the hardness measure values. 
Moreover, as with previous intervals of kDN, the cases in which $k_{h} = k_{s}$ also present similar error rates for both DCS techniques.

\begin{figure*}[!htbp]
		\centering
		\centerline{
		\subfloat[]{
			\includegraphics[scale=0.6]{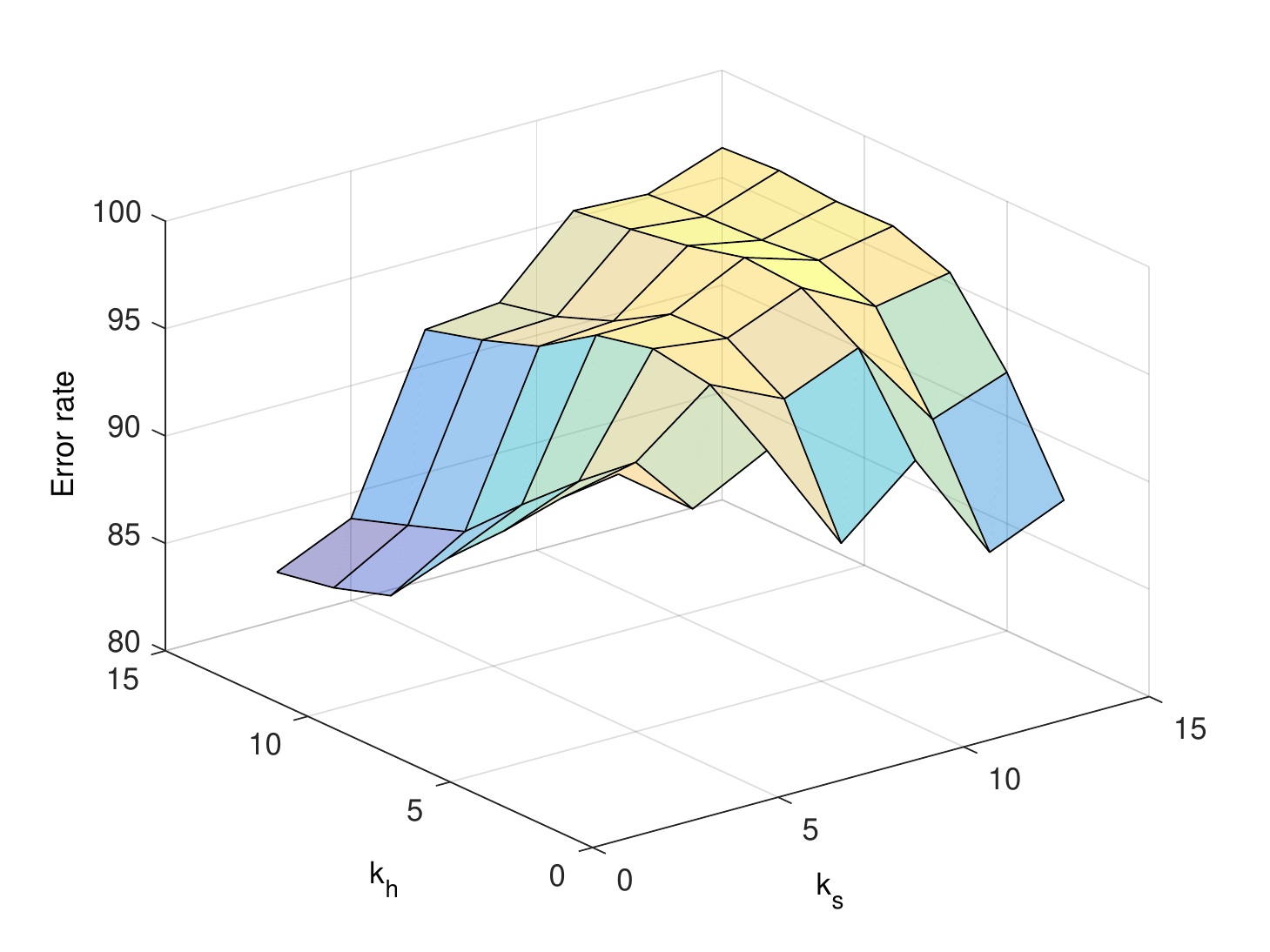}
		}
		\subfloat[]{
			\includegraphics[scale=0.6]{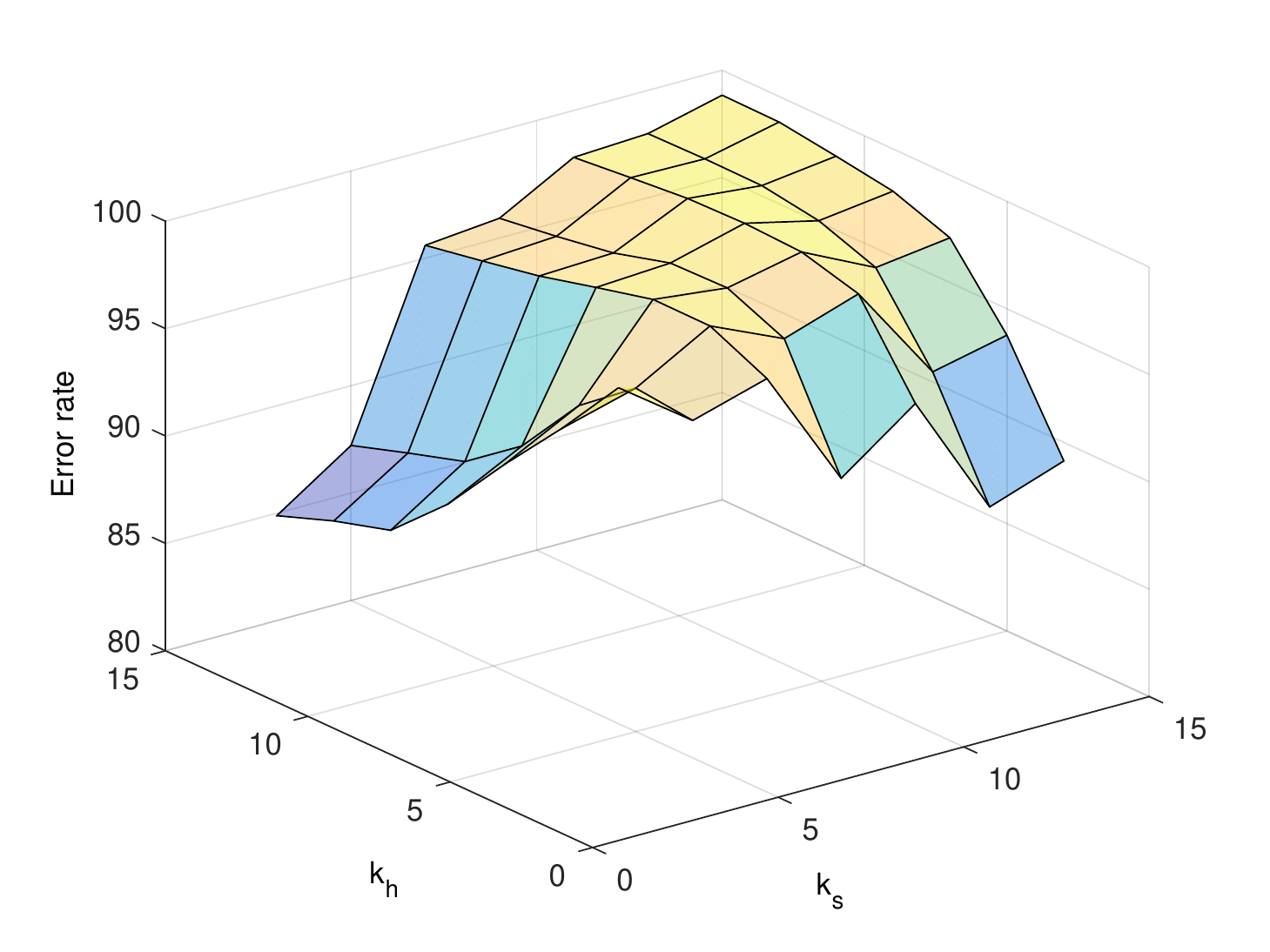}
		}}
		\caption{Mean error rate of (a) OLA and (b) LCA for instances with $kDN \in (0.75,1]$, for all datasets from Table \ref{table:datasets}. 
		$k_{h}$ and $k_{s}$ are the neighborhood sizes of kDN and the DCS technique, respectively.}
		\label{fig:kk-75-100}
\end{figure*}

\par Finally, the mean error rate for the most difficult instances, with $kDN \in (0.75,1]$, can be observed in Figure \ref{fig:kk-75-100} for both DCS techniques. 
For these instances, which are located in high overlap regions, the error rate is supposed to be quite high. 
It can be observed that similar neighborhood sizes correspond to the expected behavior of the error rate, as seen previously. 
On the other hand, the greater the difference between $k_{h}$ and $k_{s}$, the more degraded the relationship between the error rate and the hardness measure, further suggesting that the use of similar regions of competence yields a better characterization of this relationship. 
Also, the error rates using the same neighborhood sizes were quite similar in this case, with the exception of $k_{h} = k_{s} = 5$.

\par
\begin{sidewaysfigure}[!htpb]
		\centering
		\includegraphics[scale=0.7]{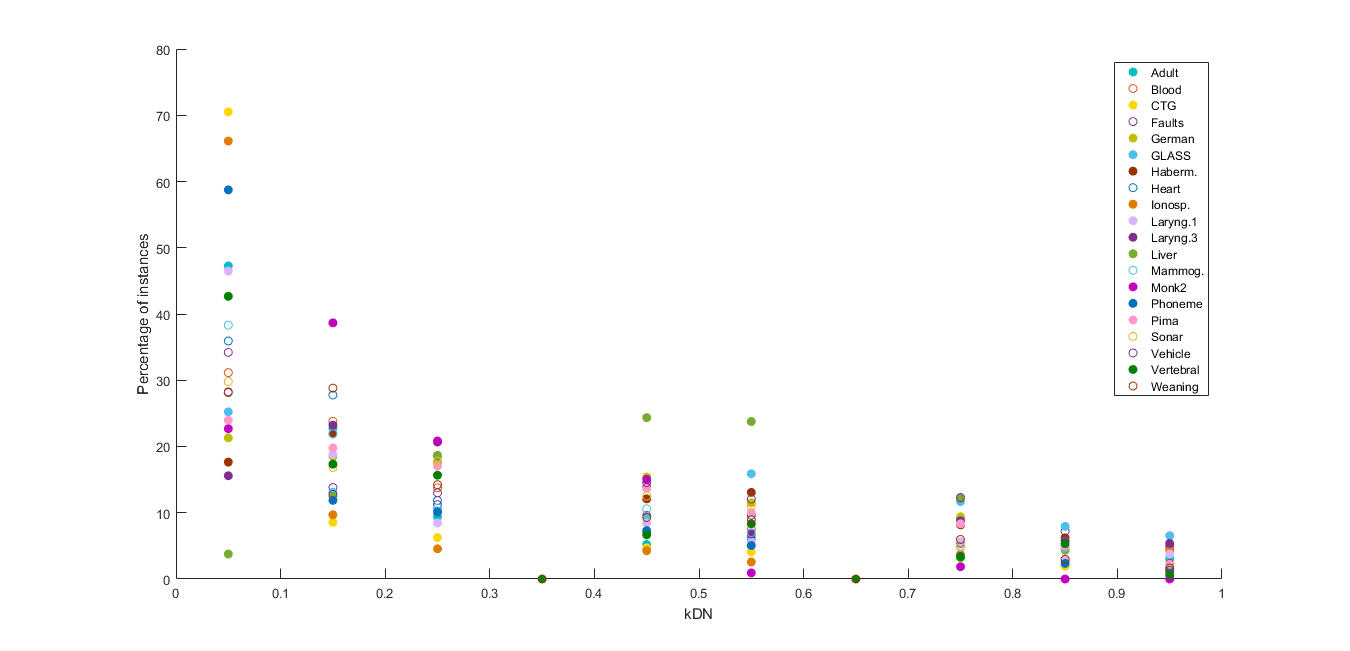}
		\caption{Percentage of instances by kDN range for all datasets from Table \ref{table:datasets}.
		The neighborhood size of the kDN measure is $k_{h} = 7.$}
		\label{fig:data-perc}
\end{sidewaysfigure}

 Since it was observed that having equal neighborhood sizes yields a rather stable correlation between the hardness measure and the errors committed by the DCS techniques, the case in which $k_{s} = k_{h} = 7$ was further analyzed in order to better investigate such correlation. 
The distribution of each datasets' instances by kDN range, with $k_{h} = 7$, can be observed in Figure \ref{fig:data-perc}. 

\par Table \ref{table:acc-dcs-ih} show the mean accuracy rates of the DCS techniques, with $k_{s} = 7$. 
The column kDN denotes the mean hardness of the instances the technique classified correctly (Hit) and misclassified (Miss). 
It can be observed that, on average, the instances the DCS techniques correctly classifies are the supposedly easy ones, which have kDN closer to zero. 
In order to assess whether the correctly classified and the misclassified instances have discernible kDN values, 
a Wilcoxon signed-rank test with a significance level of $\alpha = 0.05$ was performed on the difference between the mean hardness of the groups Hit and Miss (Wilcoxon row) for the two DCS techniques. 
Since both tests resulted in p-values smaller than $\alpha$, it can be concluded that there is a significant difference between the average hardness of the correctly classified and misclassified instances for both DCS techniques. 
This means the chosen hardness measure is indeed correlated to the error rate of the DCS techniques in such a way that they can effectively distinguish the groups of easy and hard 
instances for these techniques. 

\begin{table}[!htbp]
\centering
\caption{Mean and standard deviation of the accuracy rate and the instance hardness measures using (a) OLA and (b) LCA. 
The columns Hit and Miss denote the measures of the instances correctly and incorrectly classified by the DCS technique, respectively. 
The neighborhood sizes of the DCS techniques and the kDN measure are $k_{s} = k_{h} = 7$.
The Wilcoxon row shows the resulting p-value of a Wilcoxon signed rank test for the null hypothesis that the difference between the mean instance hardness of the pairs Hit/Miss comes from a distribution with zero median. The significance level was $\alpha = 0.05$.}
\label{table:acc-dcs-ih}
\subfloat[]{
\label{table:acc-ola}
\scalebox{0.7}{
\begin{tabular}{|c|c|cc|}
\hline
\multirow{2}{*}{\textbf{Dataset}} & \multirow{2}{*}{\textbf{Accuracy}} & \multicolumn{2}{c|}{\textbf{kDN}} \\ \cline{3-4} 
                                  &                                    & \textbf{Hit}    & \textbf{Miss}   \\ \hline
Adult 			& 87.98 (2.39) & 0.10 (0.14) & 0.59 (0.30) \\
Blood 			& 75.37 (2.26) & 0.13 (0.13) & 0.50 (0.34)  \\
CTG 			& 90.24 (0.77) & 0.05 (0.11) & 0.48 (0.30)  \\
Faults 			& 71.91 (1.60) & 0.14 (0.16) & 0.56 (0.33) \\
German 			& 71.00 (2.54) & 0.19 (0.17) & 0.38 (0.29)  \\
Glass 			& 67.17 (3.99) & 0.17 (0.16) & 0.58 (0.30) \\
Haberman 		& 74.21 (3.15) & 0.18 (0.14) & 0.47 (0.30)  \\
Heart 			& 87.35 (3.58) & 0.13 (0.15) & 0.57 (0.30)  \\
Ionosphere 		& 88.75 (2.76) & 0.07 (0.14) & 0.59 (0.37)  \\
Laryngeal1 		& 81.42 (3.53) & 0.10 (0.13) & 0.56 (0.32) \\
Laryngeal3 		& 71.63 (2.68) & 0.19 (0.13) & 0.50 (0.35) \\
Liver 			& 60.29 (2.90) & 0.24 (0.13) & 0.38 (0.26)  \\
Mammographic 	& 82.74 (2.56) & 0.12 (0.14) & 0.51 (0.30)  \\
Monk2 			& 81.71 (3.71) & 0.12 (0.10) & 0.24 (0.18) \\
Phoneme 		& 86.28 (0.78) & 0.08 (0.12) & 0.45 (0.32)  \\
Pima 			& 75.05 (4.86) & 0.17 (0.15) & 0.43 (0.30)  \\
Sonar 			& 73.37 (2.94) & 0.16 (0.16) & 0.33 (0.30)  \\
Vehicle 		& 69.50 (1.90) & 0.16 (0.15) & 0.46 (0.30) \\
Vertebral 		& 80.00 (2.33) & 0.11 (0.13) & 0.43 (0.31) \\
Weaning 		& 79.67 (4.53) & 0.16 (0.16) & 0.34 (0.31) \\ \hline
Average    		& 77.78         & 0.14         & 0.47            \\ \hline
Wilcoxon (p-value)   	& n/a            & \multicolumn{2}{c|}{$8.85 \times 10^{-5} $} \\ \hline
\end{tabular}}
}
\subfloat[]{
\label{table:acc-lca}
\scalebox{0.7}{
\begin{tabular}{|c|c|cc|}
\hline
\multirow{2}{*}{\textbf{Dataset}} & \multirow{2}{*}{\textbf{Accuracy}} & \multicolumn{2}{c|}{\textbf{kDN}} \\ \cline{3-4} 
                                  &                                    & \textbf{Hit}    & \textbf{Miss}   \\ \hline
Adult 			& 87.98 (1.89) & 0.10 (0.13) & 0.59 (0.29) \\
Blood 			& 75.85 (2.23) & 0.13 (0.13) & 0.52 (0.34)  \\
CTG 			& 90.30 (0.84) & 0.05 (0.11) & 0.49 (0.30)  \\
Faults 			& 71.99 (1.53) & 0.14 (0.15) & 0.54 (0.33) \\
German 			& 73.18 (2.84) & 0.18 (0.15) & 0.44 (0.30)  \\
Glass 			& 70.75 (3.44) & 0.17 (0.16) & 0.57 (0.30) \\
Haberman 		& 74.80 (5.06) & 0.18 (0.13) & 0.51 (0.30)  \\
Heart 			& 87.21 (3.20) & 0.13 (0.14) & 0.56 (0.29)  \\
Ionosphere 		& 86.88 (1.82) & 0.06 (0.12) & 0.69 (0.30)  \\
Laryngeal1 		& 82.36 (3.88) & 0.10 (0.13) & 0.57 (0.31) \\
Laryngeal3 		& 72.47 (3.25) & 0.19 (0.13) & 0.55 (0.34) \\
Liver 			& 60.06 (3.46) & 0.24 (0.12) & 0.41 (0.26)  \\
Mammographic 	& 83.10 (2.70) & 0.12 (0.13) & 0.51 (0.30)  \\
Monk2 			& 88.19 (3.60) & 0.15 (0.11) & 0.23 (0.22)  \\
Phoneme 		& 86.49 (0.81) & 0.08 (0.12) & 0.46 (0.31)  \\
Pima 			& 74.32 (3.43) & 0.16 (0.14) & 0.49 (0.30)  \\
Sonar 			& 72.31 (4.16) & 0.16 (0.15) & 0.39 (0.31)  \\
Vehicle 		& 70.12 (2.02) & 0.16 (0.15) & 0.47 (0.30)  \\
Vertebral 		& 81.35 (2.84) & 0.11 (0.13) & 0.47 (0.31) \\
Weaning 		& 80.99 (4.65) & 0.15 (0.14) & 0.42 (0.31) \\ \hline
Average    		& 78.53         & 0.14         & 0.49             \\ \hline
Wilcoxon (p-value)     	& n/a            & \multicolumn{2}{c|}{$8.81 \times 10^{-5}$} \\ \hline
\end{tabular}}
}
\end{table}

\begin{figure*}[!htbp]
		\centering
		\includegraphics[scale=0.55]{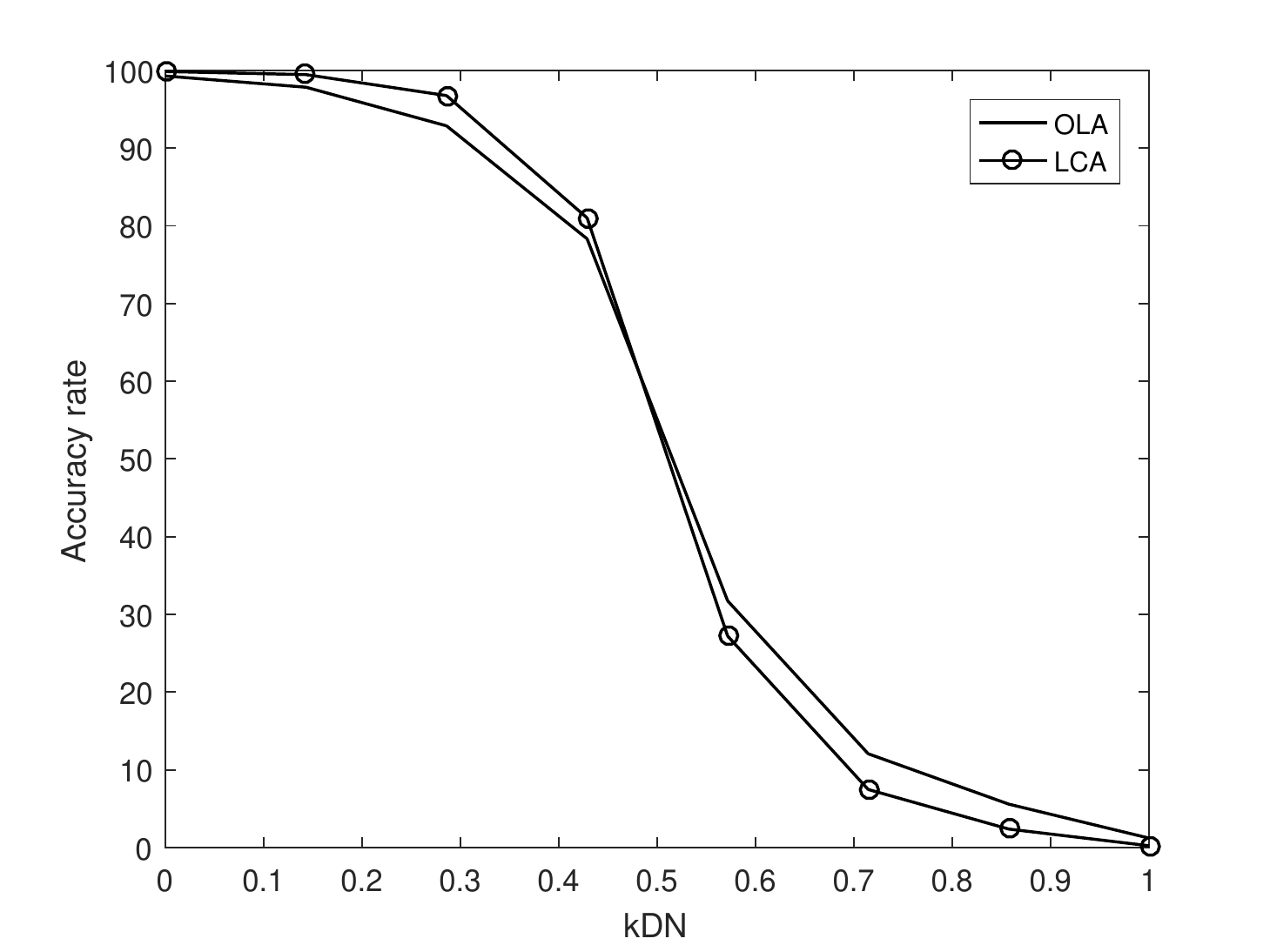}
		\caption{Mean accuracy rate of OLA and LCA for each group of kDN value, for all datasets from Table \ref{table:datasets}.
	The neighborhood sizes of the DCS techniques and the kDN measure are $k_{s} = k_{h} = 7$.}
		\label{fig:kDNmiss}
\end{figure*}

\par In order to further characterize the relationship between the kDN measure and the DCS techniques, all samples from each dataset were grouped by their true hardness value and the mean accuracy rate of both DCS techniques on each group was calculated. The results are summarized in Figure \ref{fig:kDNmiss}. 

\par It can be observed in Figure \ref{fig:kDNmiss} that the two DCS techniques misclassify the majority of all instances with kDN above 0.5, on average. 
It can also be seen a great difference between the accuracy rates of instances with kDN $ \in [0.4,0.5]$ and kDN $\in [0.5,0.6]$, which is reasonable since kDN values above 0.5 mean the majority of an instance's neighbors belong to a different class than its own class. 
This further shows the correlation between the kDN measure and the classification difficulty by the DCS techniques, which was somewhat expected since the measure and the techniques operate locally using the same mechanism of selecting the k nearest neighbors.  
Moreover, Figure \ref{fig:kDNmiss} shows that LCA correctly classifies a greater amount of the easiest instances ($kDN \leq 0.5$) than OLA, though it struggles more to correctly label the hardest instances ($kDN \geq 0.5$), on average. 

\section{The Proposed Method}
\label{sec:prop-tech}

\par In the previous section, it was shown the DCS techniques struggle to select a competent classifier for instances in regions with overlap between the problem's classes. 
Moreover, since the DCS techniques rely only on a small region, an instance's neighborhood, in order to select the most competent classifier for this instance, a global perspective in the search for a promising pool for DCS could be inadequate in such cases \cite{mariana}.

\par With that in mind, it is proposed the use of an Oracle-guided generation method on a local scope, so that the model's properties may be explored by the DCS techniques. 
The idea is to use a local pool (LP) consisted of specialized classifiers, 
each of which selected using a DCS technique from a local subpool that contains at least one competent classifier for each instance in class overlap regions of the feature space.
If the unknown instance's Region of Competence (RoC) is located in a difficult region, the LP is generated on the fly using its neighboring instances and then used to label the query sample. 
However, if the query instance is far from the classes' borders, no pool is generated and the output label is obtained using a simple nearest neighbors rule.

\par The reasoning behind the proposed approach is that using locally generated classifiers for instances in class overlap areas may be of help to the DCS techniques due to their high accuracy in these regions. 
Moreover, most works regarding DCS use classical generation methods, which were designed for static ensembles \cite{cruz2017dynamic} and therefore do not take into account the regional aspect of the competence estimation performed by the DS techniques. 
Thus, matching the perspectives of the generation and the selection stages may be advantageous for these techniques.

\par An overview of the proposed method is described in more detail in Section \ref{sec:overview}. 
Then, a step-by-step analysis of the proposed method is presented using a 2D toy problem in Section \ref{sec:toy-problem}.

\subsection{Overview}
\label{sec:overview}

\par The proposed technique is divided into two phases: 
\begin{enumerate}
  \item The \textit{offline} phase, in which the hardness estimation of the training instances is performed. 
  The hardness value of the training samples is used to identify the difficult regions of the \looseness=-1 feature space.   
  \item The \textit{online} phase, in which the RoC of the query sample is evaluated using a hardness measure in order to identify if the local area is difficult. 
  If it is not, the sample is labelled using a nearest neighbors rule. 
  Otherwise, a pool composed of the most locally accurate classifiers in that region, as indicated by a DCS technique, is generated and used to label the \looseness=-1 unknown instance.
\end{enumerate}

\par
\begin{figure*}[!htb]
		\centering
		\subfloat[]{
			\label{fig:overview-off}
			\includegraphics[width=0.4\textwidth]{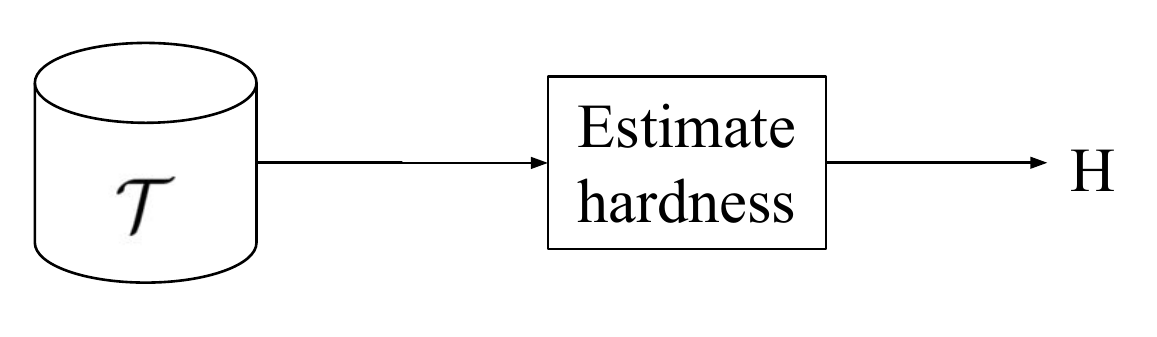}
		}
		\\
		\subfloat[]{
			\label{fig:overview-on}
			\includegraphics[width=1\textwidth]{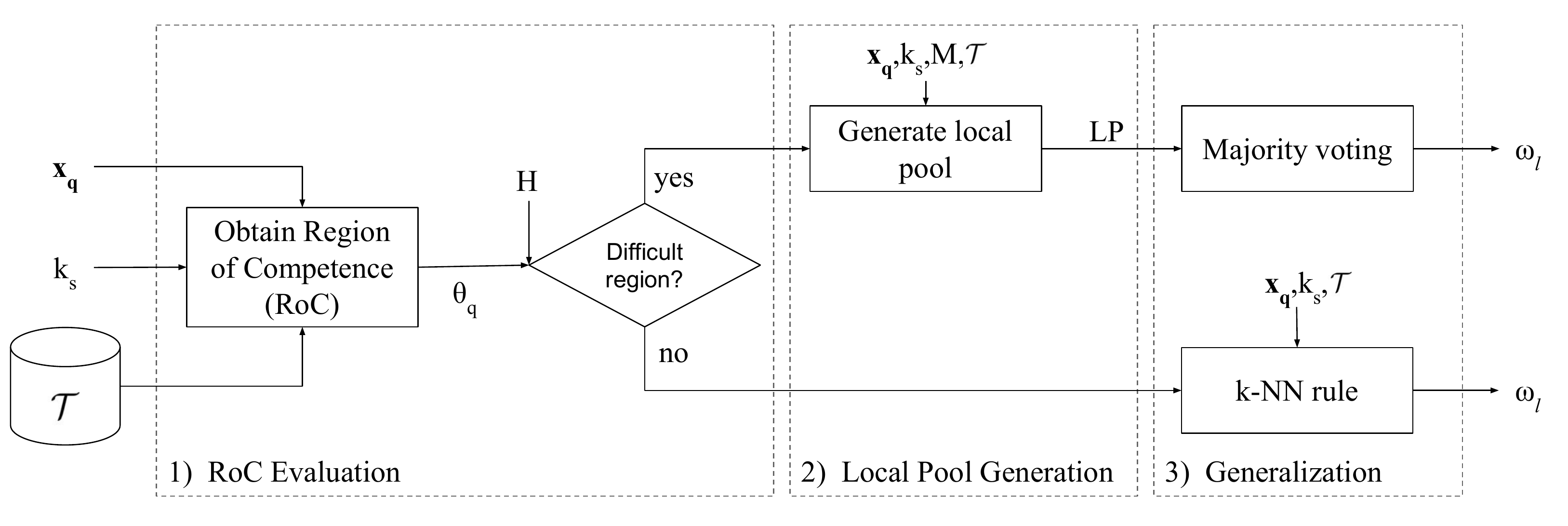}
		}
\caption{Overview of the (a) offline and (b) online phases of the proposed method.
$\Train$ is the training set, $H$ is the set of hardness estimates, $\mathbf{x_{q}}$ is the query sample, $\theta_{q}$ is its Region of Competence (RoC), $k_{s}$ is the size of $\theta_{q}$, $LP$ is the local pool, $M$ is the pool size of $LP$ and $\omega_{l}$ is the output label of $\mathbf{x_{q}}$. 
In the offline phase, the hardness value of all instances in $\Train$ is estimated and stored in $H$. 
In the online phase, $\theta_{q}$ is first obtained and evaluated based on the hardness values in $H$. If it only contains easy instances, the k-NN rule is used to label $\mathbf{x_{q}}$ in the last step. Otherwise, the local pool is generated in the second step, and $\mathbf{x_{q}}$ is labelled via majority voting of the classifiers in $LP$ in the third step.
}
\label{fig:overview}
\end{figure*}

 An overview of the proposed techniques' phases is depicted in Figure \ref{fig:overview}, in which $\Train$ is the training set, $H$ is the set of hardness estimates, $\mathbf{x_{q}}$ is the query sample, $\theta_{q}$ is its RoC, $k_{s}$ is the size of $\theta_{q}$, $LP$ is the local pool, $M$ is the pool size of $LP$ and $\omega_{l}$ is the output label of $\mathbf{x_{q}}$. 
In the offline phase (Figure \ref{fig:overview-off}), the hardness of each instance $\mathbf{x_{n}} \in \Train$ is estimated using a hardness measure, and its value is stored in $H$, to be later used in the online phase.
The online phase, in which the query sample $\mathbf{x_{q}}$ is labelled, is performed in three steps: RoC evaluation, local pool generation and generalization, as shown in Figure \ref{fig:overview-on}.

\par In the RoC evaluation step, the $k_{s}$ nearest neighbors in the training set $\Train$ of the query sample $\mathbf{x_{q}}$ are selected to form the query sample's RoC $\theta_{q}$. 
The dynamic selection dataset (DSEL), which is a set of labelled samples used for RoC definition in DS techniques \cite{cruz2017dynamic}, was not used in the proposed method because the SGH method did not present overfitting when used for RoC definition \cite{mariana}. 
Then, the instances that compose $\theta_{q}$ are analyzed. If all of them are not in an overlap region, that is, they all have hardness estimate $H = 0$, the method skips the local pool generation and goes directly to the generalization phase. 
The output class $\omega_{l}$ of $\mathbf{x_{q}}$ is then obtained using the k-NN rule with parameter $k_{s}$. 
However, if there is at least one instance in $\theta_{q}$ located in an identified class overlap area, the query sample's RoC is considered to be a difficult region. 
Thus, the local pool $LP$ containing $M$ classifiers is generated in the second step and used to label $\mathbf{x_{q}}$ in the generalization step via majority voting. 
The local pool generation step is explained next.

\begin{figure*}[!htb]
		\centering		
		\includegraphics[width=0.8\textwidth]{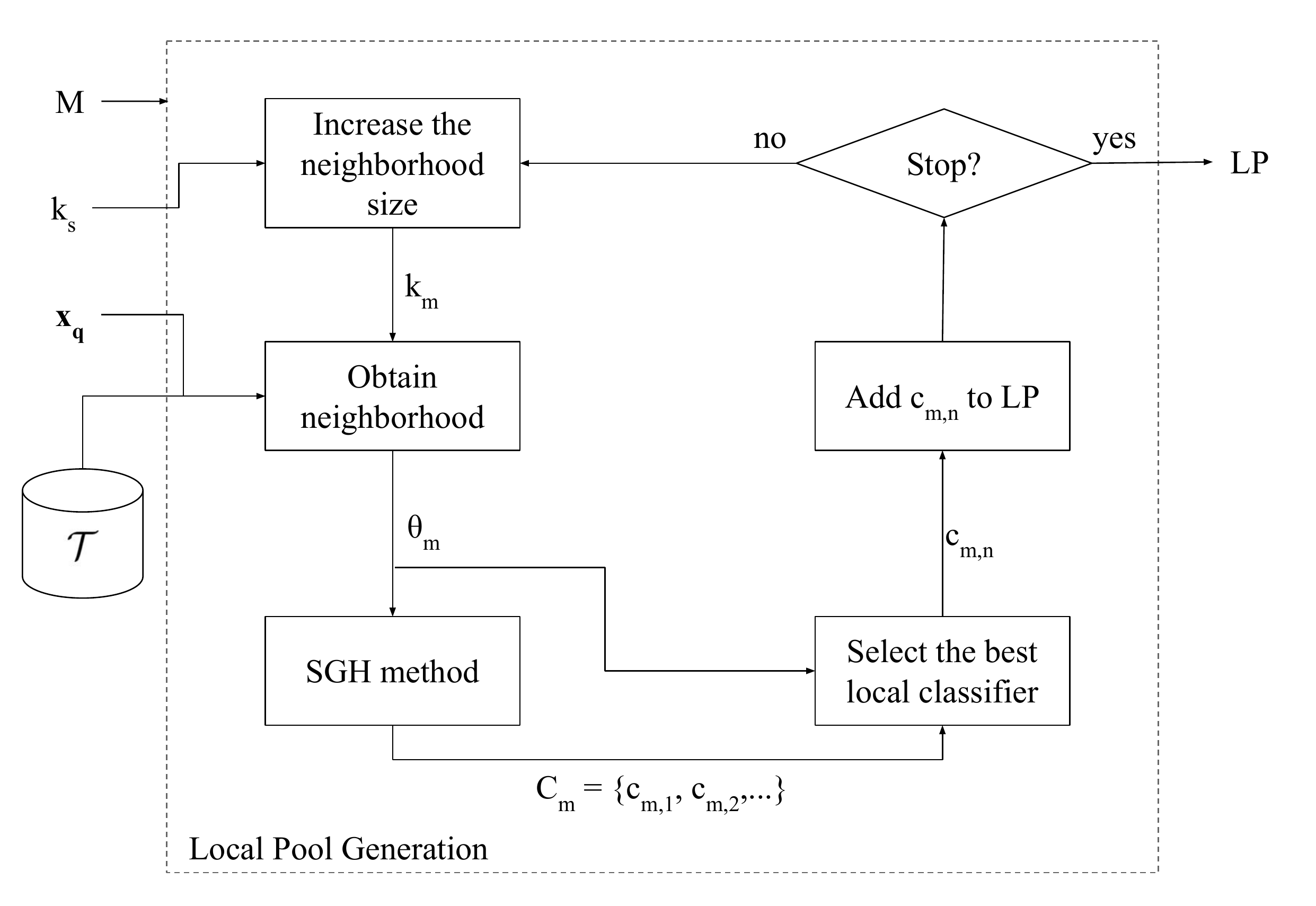}
		\caption{Local pool generation step. 
		The inputs to the generation scheme are the training set $\Train$, the query sample $\mathbf{x_{q}}$, the size $k_{s}$ of the query sample's RoC and the local pool size $M$. 
		The output is the local pool $LP$. 
		In the \textit{m}-th iteration, the query sample's neighborhood $\theta_{m}$ of size $k_{m}$ is obtained and used as input to the SGH method, which yields the subpool $C_{m}$. 
		The classifiers from $C_{m}$ are then evaluated over $\theta_{m}$ using a DCS technique. 
		The classifiers' notation refers a classifier $c_{m,k}$ as the \textit{k}-th classifier from the \textit{m}-th subpool ($C_{m}$). 
		The most competent classifier $c_{m,n}$ in subpool $C_{m}$ is then selected and added to the local pool $LP$. 
		This process is then repeated until $LP$ contains $M$ locally accurate classifiers.
		}
		\label{fig:gen-lp}
\end{figure*}

\par Figure \ref{fig:gen-lp} shows the generation procedure of the local pool $LP$. 
The pool size $M$ of the local pool is an input parameter. 
The other inputs are the training set ($\Train$), the query sample ($\mathbf{x_{q}}$) and the query sample's RoC size ($k_{s}$). 
The $LP$ is constructed iteratively. 
In the \textit{m}-th iteration, the query sample's neighboring instances in the training set are obtained using any nearest neighbors method, with parameter $k_{m}$. 
These neighboring instances form the query sample's neighborhood $\theta_{m}$, which is used as input to the SGH method (Algorithm~\ref{alg:sgh} from Section \ref{sec:ih-analysis}). 
The SGH method then returns a local subpool $C_{m}$ that fully covers the neighborhood $\theta_{m}$. 
That is, the presence of at least one competent classifier $c_{m,k} \in C_{m}$ for each instance in $\theta_{m}$ is guaranteed. 
The indexes in the classifiers' notation indicates that the classifier $c_{m,k}$ is the \textit{k}-th classifier from the \textit{m}-th subpool ($C_{m}$). 
Then, the most competent classifier $c_{m,n}$ from $C_{m}$ in the region delimited by the neighborhood $\theta_{q}$ is selected by a DCS technique and added to the local pool. 
The same procedure is performed in iteration $m+1$ with the neighborhood size $k_{m+1}$ increased by 2.  
This process is then repeated until the local pool contains $M$ locally accurate classifiers.

\begin{algorithm}[!htb]
\centering
\begin{algorithmic}[1]
\Input $\Train, k_{h}$ \Comment{Training dataset and kDN neighborhood size}
\Output $H$ \Comment{Estimated hardness values}
\For {every $\mathbf{x_{i}}$ in $\Train$} 
	\State $ H(i) \gets kDN(\mathbf{x_{i}},\Train,k_{h})$ \Comment{Calculate hardness (Equation \ref{eq:kdn})}
\EndFor \\
\Return $H$
\end{algorithmic}
\caption{Offline phase (proposed technique).}
\label{alg:off}
\end{algorithm}

\par The pseudocode of the offline phase of the proposed method is shown in Algorithm~\ref{alg:off}. 
Its inputs are the training set $\Train$ and the kDN parameter $k_{h}$, which denotes the neighborhood size of the hardness estimate. 
From Step 1 to Step 3, the hardness of each instance $\mathbf{x_{i}} \in \Train$ is calculated and stored in $H$, which is then returned in Step 4.

\begin{algorithm}[!htb]
\centering
\begin{algorithmic}[1]
\Input $\mathbf{x_{q}}, \Train, H$ \Comment{Query sample, training set and hardness estimates}
\Input $k_{s},M$ \Comment{RoC size and pool size of local pool $LP$}
\Output $\omega_{l}$ \Comment{Output label of $\mathbf{x_{q}}$ }
\State $\theta_{q} \gets obtainRoC(\mathbf{x_{q}},k_{s},\Train)$ \tikzmark{t1} \Comment{Obtain the query instance's RoC} 
\If {$ \{\exists \mathbf{x_{i}} \in \theta_{q} | H(i) > 0\} $} \tikzmark{b1}	
	\State $LP \gets \{\}$ \tikzmark{t2} \Comment{Local pool initially empty}
	\For {every $m$ in $\{1,2,...,M\}$}
		\State $k_{m} \gets k_{s} + 2 \times (m-1)$ \Comment{Increase neighborhood size by $2$}
		\State $\theta_{m} \gets obtainNeighborhood(\mathbf{x_{q}},k_{m},\Train)$ \tikzmark{r1} \tikzmark{r2} \tikzmark{r3} \Comment{Obtain neighborhood of $\mathbf{x_{q}}$}
		\State $C_{m} \gets generatePool(\theta_{m})$ \Comment{Generate local subpool $C_{m}$}
		\State $c_{m,n} \gets selectClassifier(\mathbf{x_{q}},\theta_{m},C_{m})$
		\Comment{Select best classifier in $C_{m}$}
		\State $LP \gets LP \cup \{c_{m,n}\}$ \Comment{Add  $c_{m,n}$ to $LP$}
	\EndFor \tikzmark{b2}
	\State $ \omega_{l} \gets majorityVoting(\mathbf{x_{q}}, LP)$  \tikzmark{t3} \Comment{Label $\mathbf{x_{i}}$ with majority voting on $LP$} 
\Else
	\State $ \omega_{l} \gets kNN(\mathbf{x_{q}},k_{s},\Train)$ \Comment{Label query sample using k-NN rule}
\EndIf \tikzmark{b3} \\ 
\Return $\omega_{l}$
\end{algorithmic}
\caption{Online phase (proposed technique).}
\label{alg:prop-met} 
\AddNote{t1}{b1}{r1}{1) RoC \\Evaluation}
\AddNote{t2}{b2}{r1}{2) Local Pool\\Generation}
\AddNote{t3}{b3}{r1}{3) Generalization}
\end{algorithm}

\par The online phase, on the other hand, is described in more detail in Algorithm~\ref{alg:prop-met}. 
Its inputs are the query sample $\mathbf{x_{q}}$, training set $\Train$, the set of hardness estimates $H$, the RoC size $k_{s}$ and the local pool size $M$. 
In Step 1, the query sample's RoC $\theta_{q}$ is obtained by selecting the $k_{s}$ closest samples to $\mathbf{x_{q}}$ in the training set. 
The RoC is then evaluated in Step 2. 
If all instances in $\theta_{q}$ are not located in a difficult region, that is, their hardness value is zero, the method goes straight to Step 13 and the query sample's output label $\omega_{l}$ is obtained using the k-NN rule with parameter $k_{s}$ and returned in Step 15.

\par However, if there is one instance $\mathbf{x_{i}}$ from $\theta_{q}$ whose hardness estimate $H(i)$ is above zero, the region is considered a difficult one and the method proceeds to Step 3. 
Each classifier in the local pool $LP$ is obtained in the loop that iterates $M$ times (Step 4 to Step 10). 
In each iteration, the neighborhood size $k_{m}$ is calculated in Step 5. 
Then, the query sample's neighborhood $\theta_{m}$ is obtained using a nearest neighbors method in Step 6. 
The subpool $C_{m}$ is then generated in Step 7 using the SGH method with $\theta_{m}$ as training set. 
In Step 8, a DCS technique is then used to select the most competent classifier $c_{m,n}$ in $C_{m}$. 
The classifier $c_{m,n}$ is added to $LP$ in Step 9, and then the loop continues until the local pool is complete. 
Finally, the query sample's label $\omega_{l}$ is obtained using majority voting over the locally accurate classifiers in $LP$ and returned in Step 15.

\subsection{Step-by-step Analysis}
\label{sec:toy-problem}

\par In order to better understand the generation process by the proposed technique, the latter was executed over a 2D toy problem dataset. 
The P2 Problem \cite{p2problem} was chosen for its complex borders. 
Since the P2 problem has no overlap between the classes, noise was added to the original distribution by randomly changing the labels of the samples near the class borders. 
The dataset used in this analysis contains 1000 instances, 75\% of which were used for training and the rest for test. 
The parameters used in this demonstration were $k_{h} = k_{s} = 7$ and $M = 7$. 
The method used for selecting the query instance's neighborhood in $getNeighborhood()$ (Step 6 of Algorithm~\ref{alg:prop-met}) for this example was the regular k-NN, and the DCS technique used to select the most competent classifier (Step 8 of Algorithm~\ref{alg:prop-met}) was OLA. 

\begin{figure*}[!htb]
		\centering
		\subfloat[]{
			\label{fig:train}
			\includegraphics[width=0.47\textwidth]{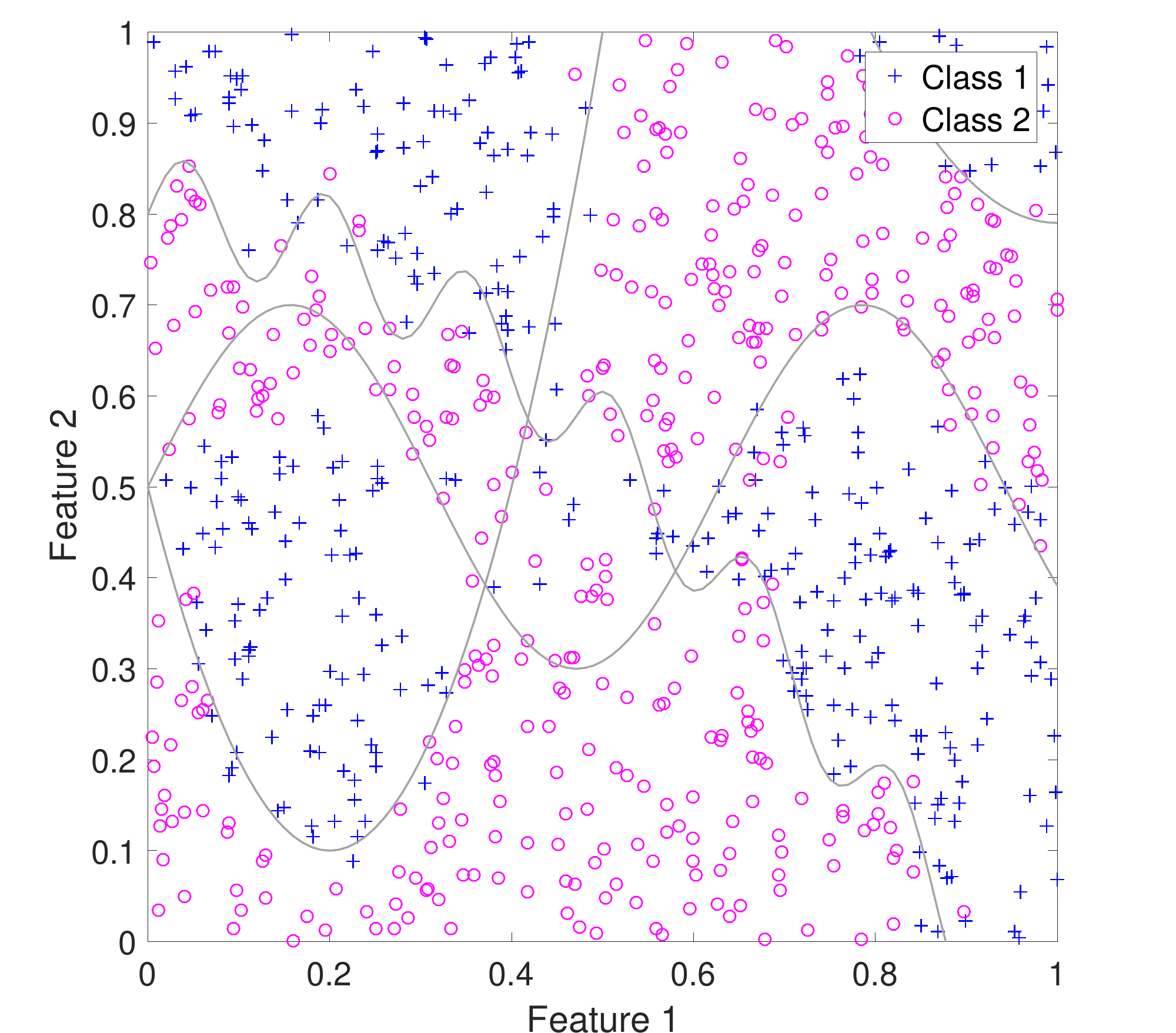}
		}
		\subfloat[]{
			\label{fig:train+kdn}
			\includegraphics[width=0.435\textwidth]{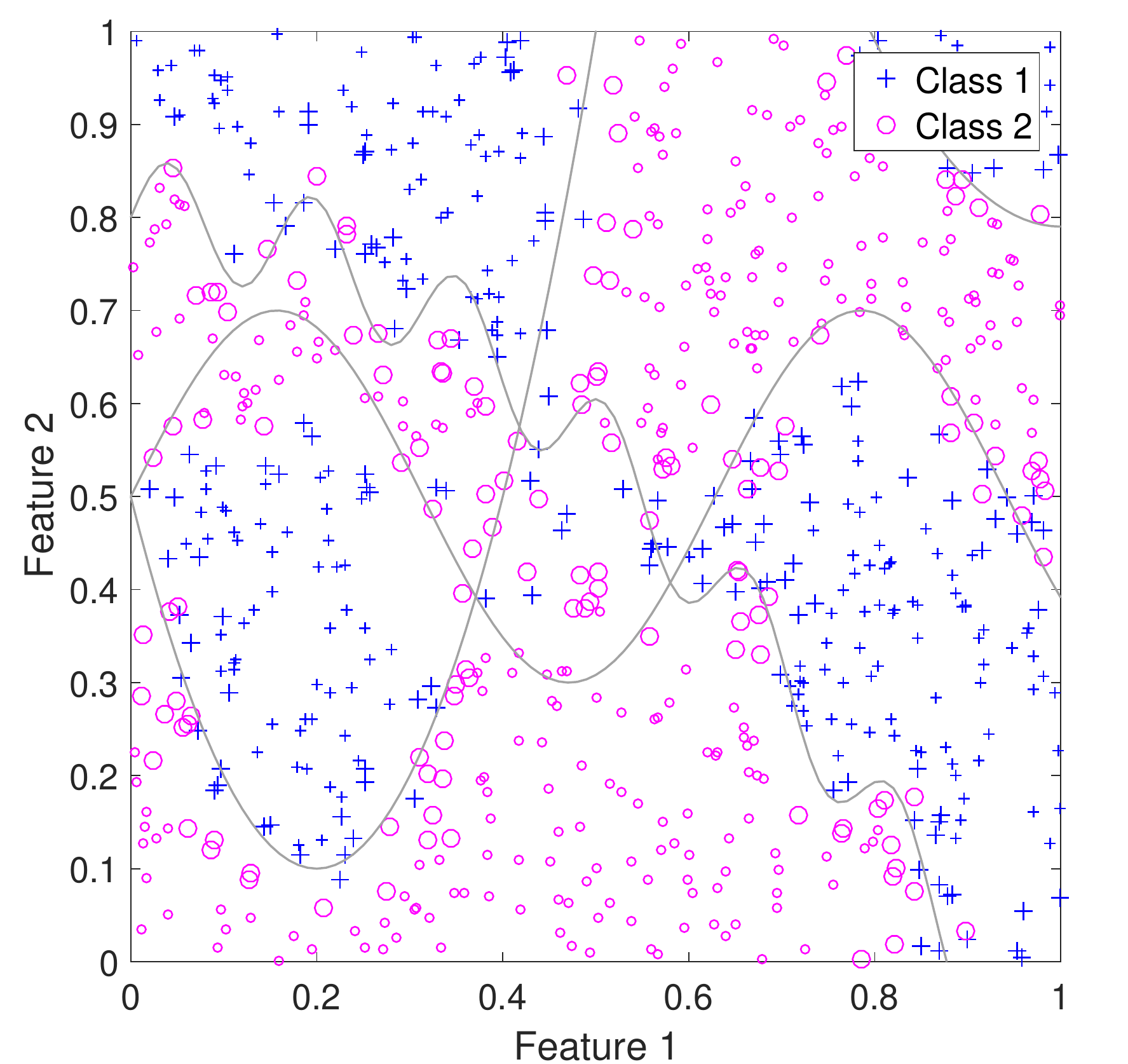}
		}
\caption{P2 Problem dataset, with theoretical decision boundaries in grey. The training set is depicted in (a), and in (b) the same set is shown with hard instances in large markers and easy instances in small ones.}
\end{figure*}	

\par The P2 Problem training set used in this analysis is shown in Figure \ref{fig:train}, with its theoretical decision boundaries in grey. 
The hardness estimation in Step 1 to Step 3 of Algorithm~\ref{alg:off} separates the training instances with estimated hardness above zero, that is, the instances closer to the border between classes, and the remaining ones. 
Figure \ref{fig:train+kdn} shows the training instances closer to the borders (large markers) and the instances with $H(i) = 0$ (small markers).

\par
\begin{figure*}[!htb]
		\centering
		\subfloat[]{
		\label{fig:xq-easy}
			\includegraphics[width=0.45\textwidth]{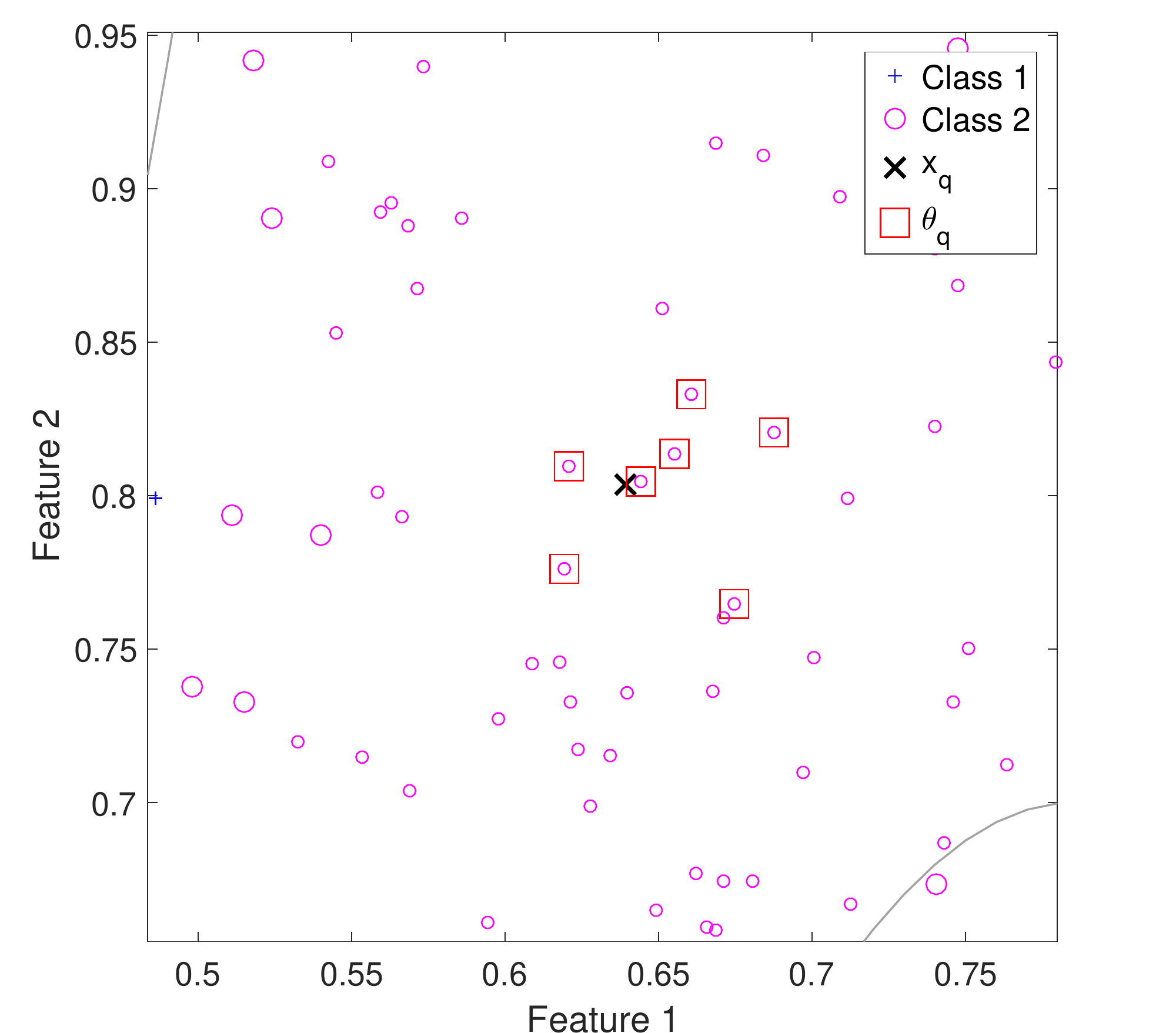}
		}
		\subfloat[]{
		\label{fig:xq-hard}
			\includegraphics[width=0.45\textwidth]{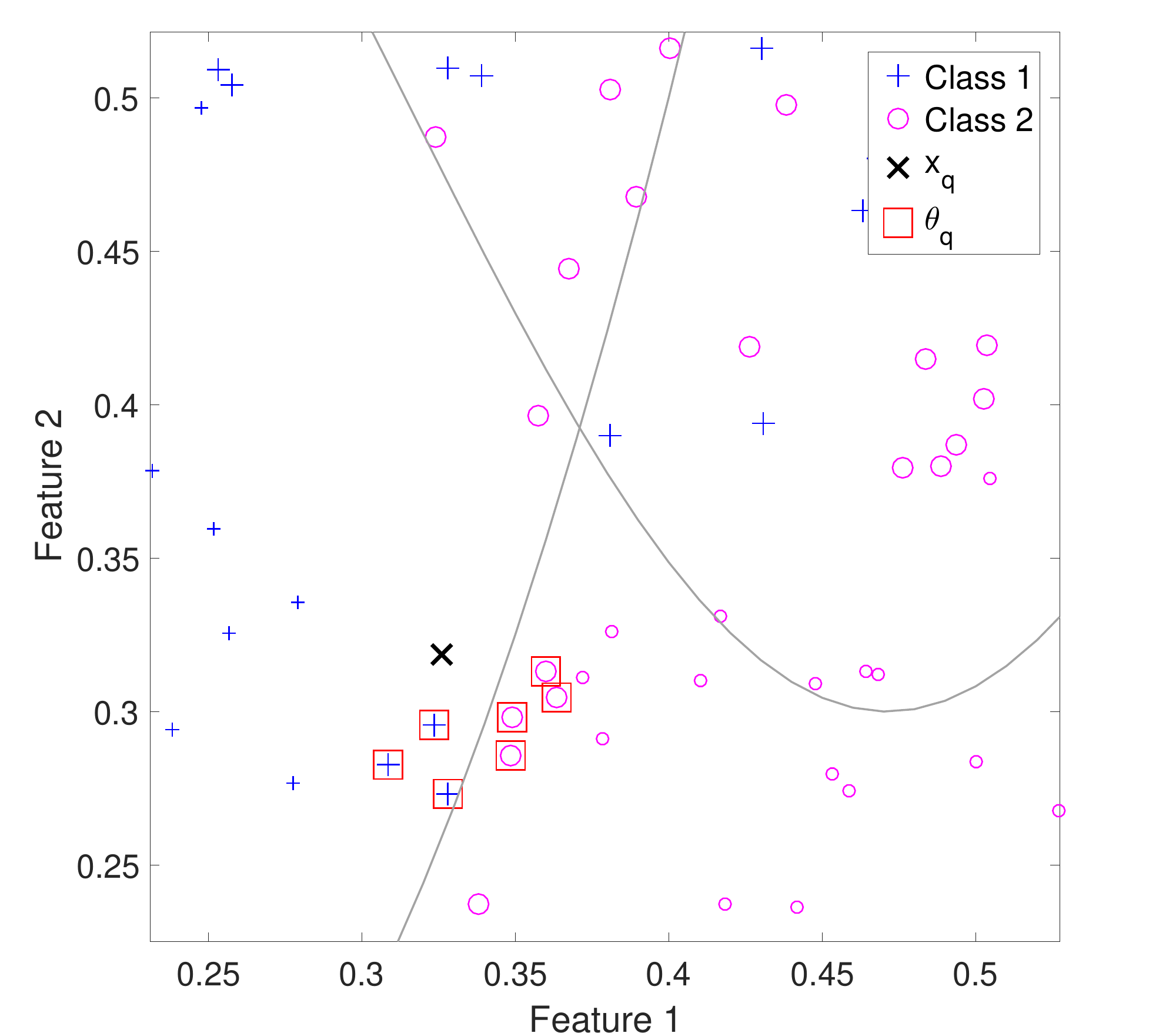}
		}
		\caption{Two different scenarios of the online phase. 
		In (a), the query instance $\mathbf{x_{q}}$ belongs to Class 2. 
		Since all instances in its neighborhood $\theta_{q}$ are easy (small markers), the k-NN rule is used to label $\mathbf{x_{q}}$. 
		On the other hand, all instances in the query sample's neighborhood $\theta_{q}$ in (b) are deemed hard (large markers). 
		Thus, the local pool $LP$ will label the query instance $\mathbf{x_{q}}$, which belongs to Class 1.
		}
\end{figure*}

 Two scenarios of the proposed scheme's online phase can be observed in Figure \ref{fig:xq-easy} and Figure \ref{fig:xq-hard}. 
In the first, the input query instance $\mathbf{x_{q}}$ of Algorithm~\ref{alg:prop-met} belongs to Class 2. 
The query sample's RoC $\theta_{q}$ is obtained selecting its k-nearest neighbors over the training set $\Train$ in Step 1. 
In this case, since all instances in $\theta_{q}$ have estimated hardness $H_{i} = 0$, represented in Figure \ref{fig:xq-easy} by small markers, $\mathbf{x_{q}}$ is considered to be in an easy region of the feature space. 
Therefore, the procedure goes to Step 13, in order to obtain the output label $\omega_{l}$ of $\mathbf{x_{q}}$ using the k-NN rule over the training set with parameter $k_{s}$. 
Then, the query sample's label is returned in Step 15.
In this case $\omega_{l} = 2$ since all $k_{s}$ neighbors of $\mathbf{x_{q}}$ belong to this class. 

\par In the second scenario, shown in Figure \ref{fig:xq-hard}, the query instance $\mathbf{x_{q}}$ of Algorithm~\ref{alg:prop-met} belongs to Class 1. 
Its RoC $\theta_{q}$ is obtained in Step 1, with more than half of its instances belonging to the opposite class. 
Thus, a simple k-NN rule would misclassify this query sample. 
The query instance's RoC $\theta_{q}$ is then analyzed in Step 2.
The hardness estimate $H_{i}$ of each neighbor is verified in Step 2 of Algorithm~\ref{alg:prop-met}, and since at least one of them is above zero, depicted in Figure \ref{fig:xq-hard} in large markers, the local pool (LP) will be generated and used from this step forward. 
Starting with an empty set (Step 3), each iteration from Step 4 to Step 10 adds a single classifier to LP. 


\par In the first iteration, the neighborhood size $k_{1}$ is set to 7 in Step 5, and then the $k_{1}$ nearest neighbors of $\mathbf{x_{q}}$ are selected to compose the query sample's neighborhood $\theta_{1}$ in Step 6. 
The local subpool $C_{1}$ is then generated using $\theta_{1}$ as the input dataset to the SGH method. 
The resulting pool, which guarantees an Oracle accuracy rate of 100\% in $\theta_{1}$, is shown in Figure \ref{fig:xq-hard-lp1}, containing only one classifier, $c_{1,1}$. 
Since there is only one classifier in $C_{1}$, $c_{1,1}$ is selected to compose $LP$ in Step 8 and Step 9. 

\par In the second iteration, the neighborhood parameter is increased by $2$ in Step 5, and the resulting neighborhood $\theta_{2}$ contains $k_{2} = 9$ instances, as shown in Figure \ref{fig:xq-hard-lp2}. 
Then, the local subpool $C_{2}$ is generated in Step 7, with $\theta_{2}$ as the input parameter of the SGH method. 
Since only one classifier was able to deliver an Oracle accuracy rate of 100\% over $\theta_{2}$, the resulting pool contains only $c_{2,1}$, which is selected in Step 8 to be added to $LP$ in Step 9. 


\par The neighborhood $\theta_{3}$, obtained in Step 6 of the third iteration, contains $k_{3} = 11$ instances, as Figure \ref{fig:xq-hard-lp3} shows. 
$C_{3}$ is then generated in Step 7 so that it fully covers $\theta_{3}$, resulting in only one classifier ($c_{3,1}$), which is later added to $LP$ in Step 9.


\par The fourth local subpool $C_{4}$, depicted in Figure \ref{fig:xq-hard-lp4}, is generated in Step 7 of the fourth iteration, with neighborhood $\theta_{4}$ of size $k_{4} = 13$ as input to the SGH method. 
The only classifier generated, $c_{4,1}$, is then added to $LP$ in Step 9.


\begin{figure*}[!htb]
		\centering
		\subfloat[]{
			\label{fig:xq-hard-lp1}
			\includegraphics[width=0.45\textwidth]{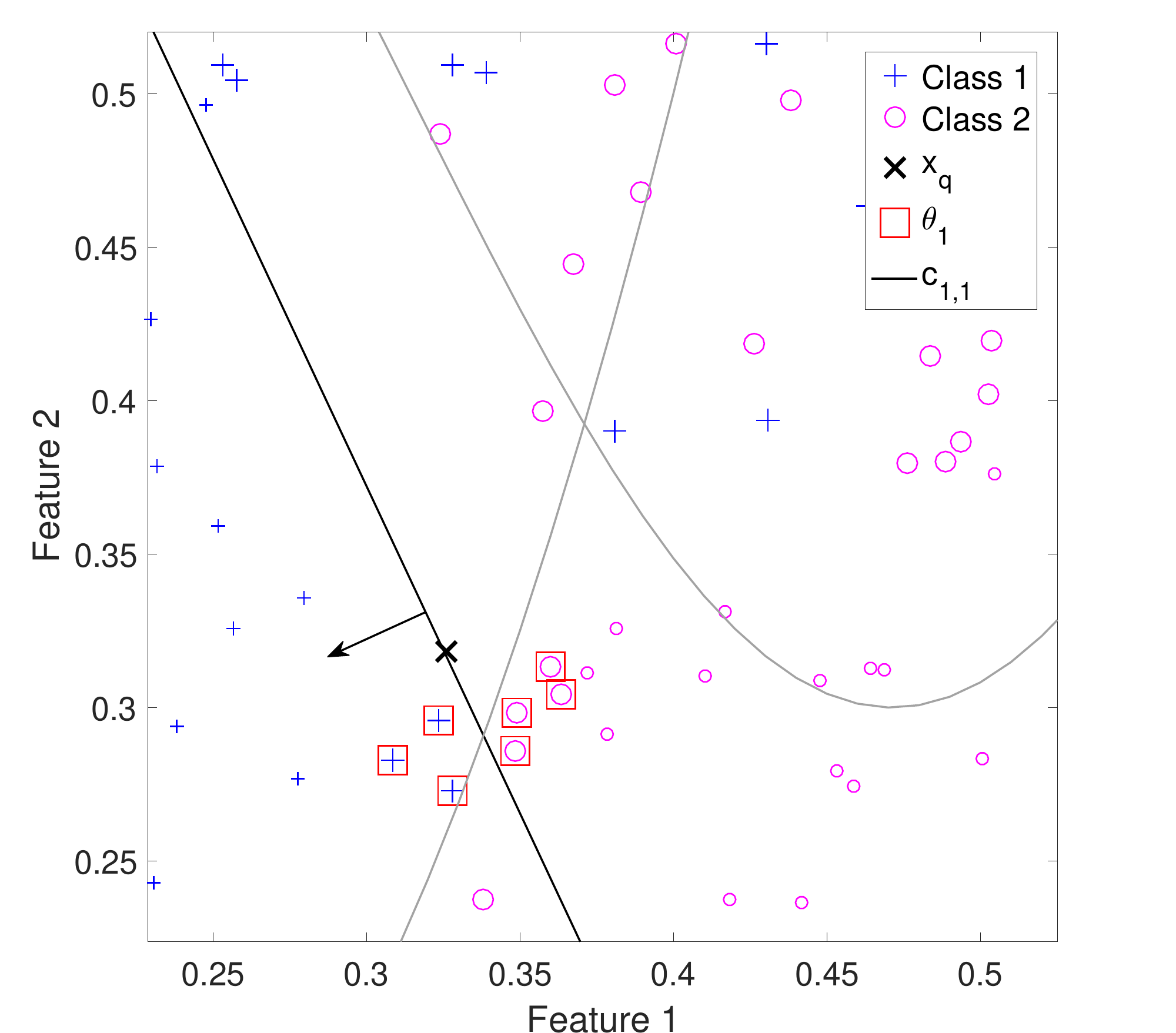}
		}
		\subfloat[]{
			\label{fig:xq-hard-lp2}
			\includegraphics[width=0.45\textwidth]{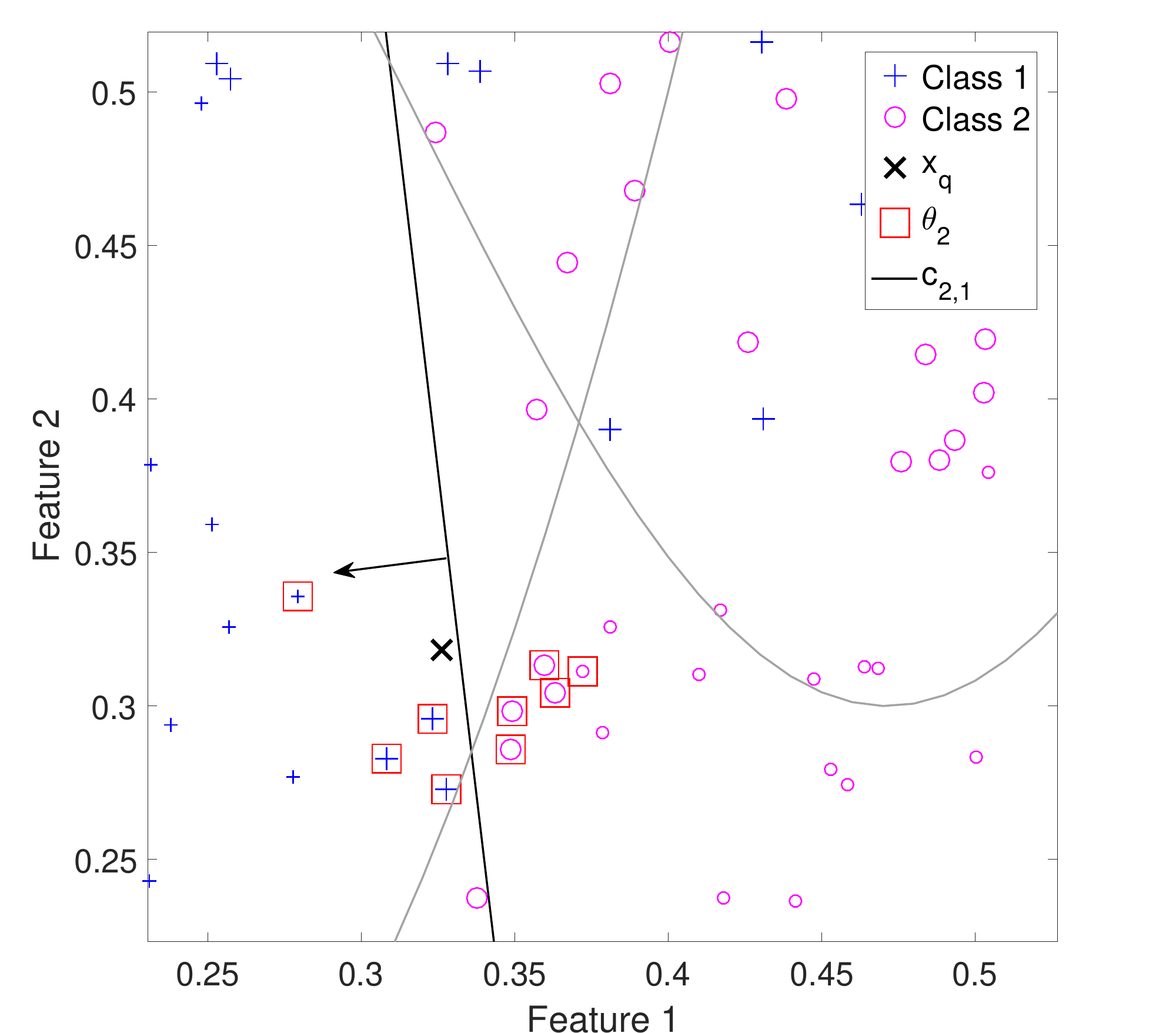}
		}
		\\
		\subfloat[]{
		\label{fig:xq-hard-lp3}
			\includegraphics[width=0.45\textwidth]{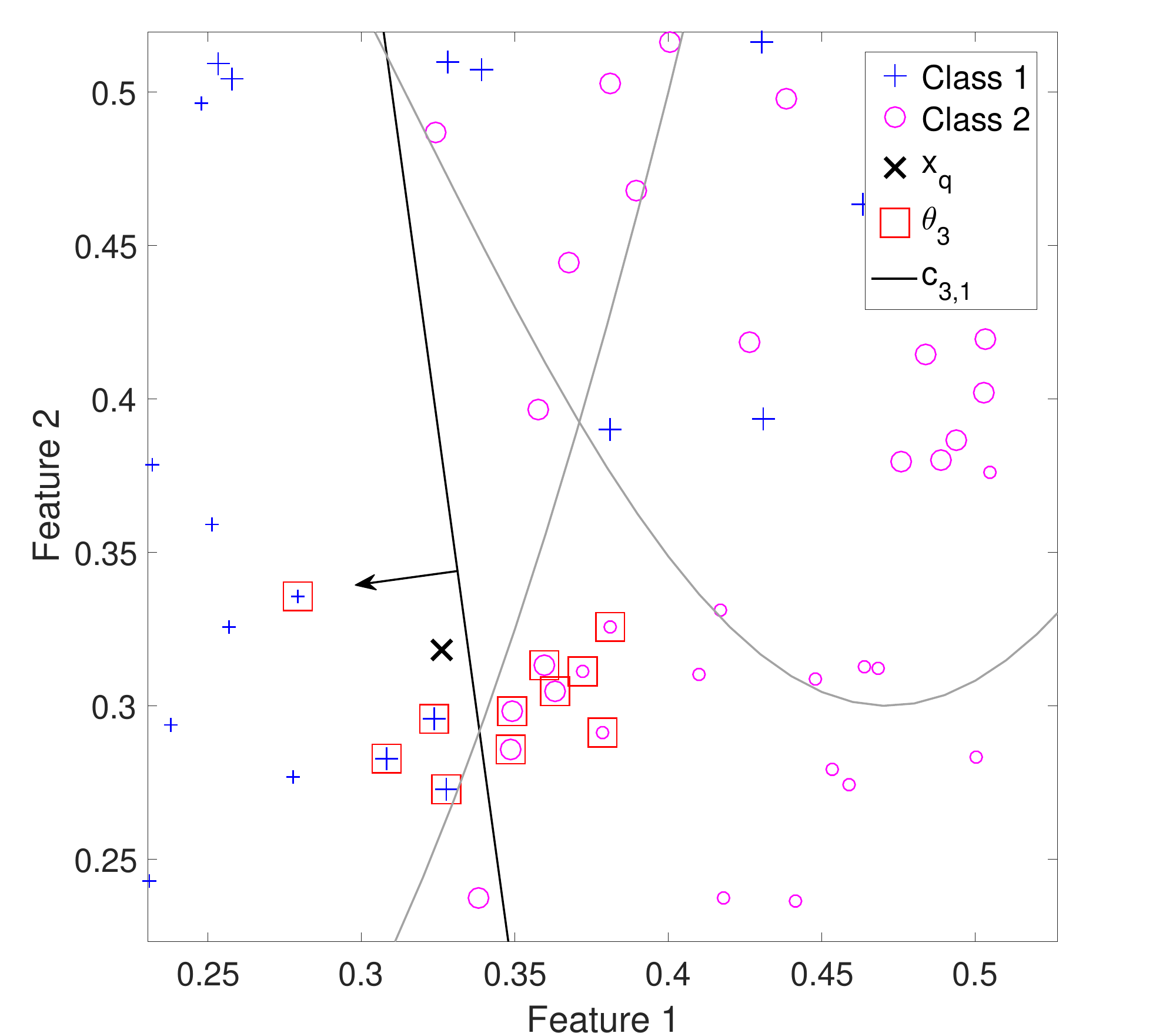}
		}
		\subfloat[]{
		\label{fig:xq-hard-lp4}
			\includegraphics[width=0.45\textwidth]{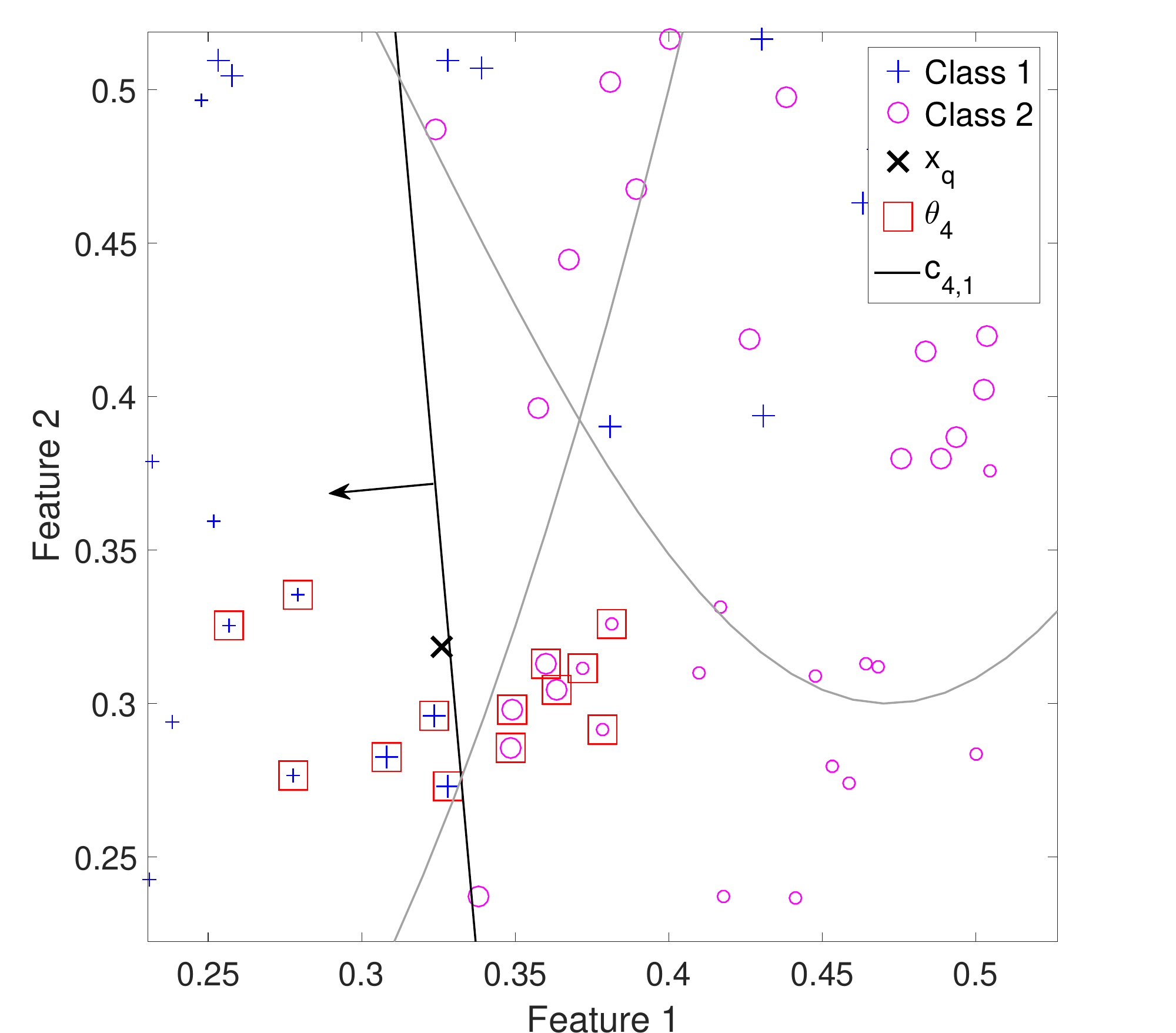}
		}
\caption{Local pool generation. 
		(a) First, (b) second, (c) third, (d) fourth, (e) fifth, (f) sixth and (g) seventh iteration of the method, with its respective neighborhoods ($\theta_{m}$) and generated local subpools $C_{m}$ formed by the depicted classifiers ($c_{m,k}$). 
		The arrows indicate in which part of the feature space the classifiers label as Class 1. 
		Each local subpool $C_{m}$ is obtained using the SGH method with its respective neighborhood $\theta_{m}$, which increases in each iteration, as input. 
		The final local pool $LP$, formed by the best classifiers in each subpool $C_{m}$, is shown in (h).  
		}
\end{figure*}

\begin{figure*}[!htb]
\ContinuedFloat
	  \centering	
		\subfloat[]{
		\label{fig:xq-hard-lp5}
			\includegraphics[width=0.45\textwidth]{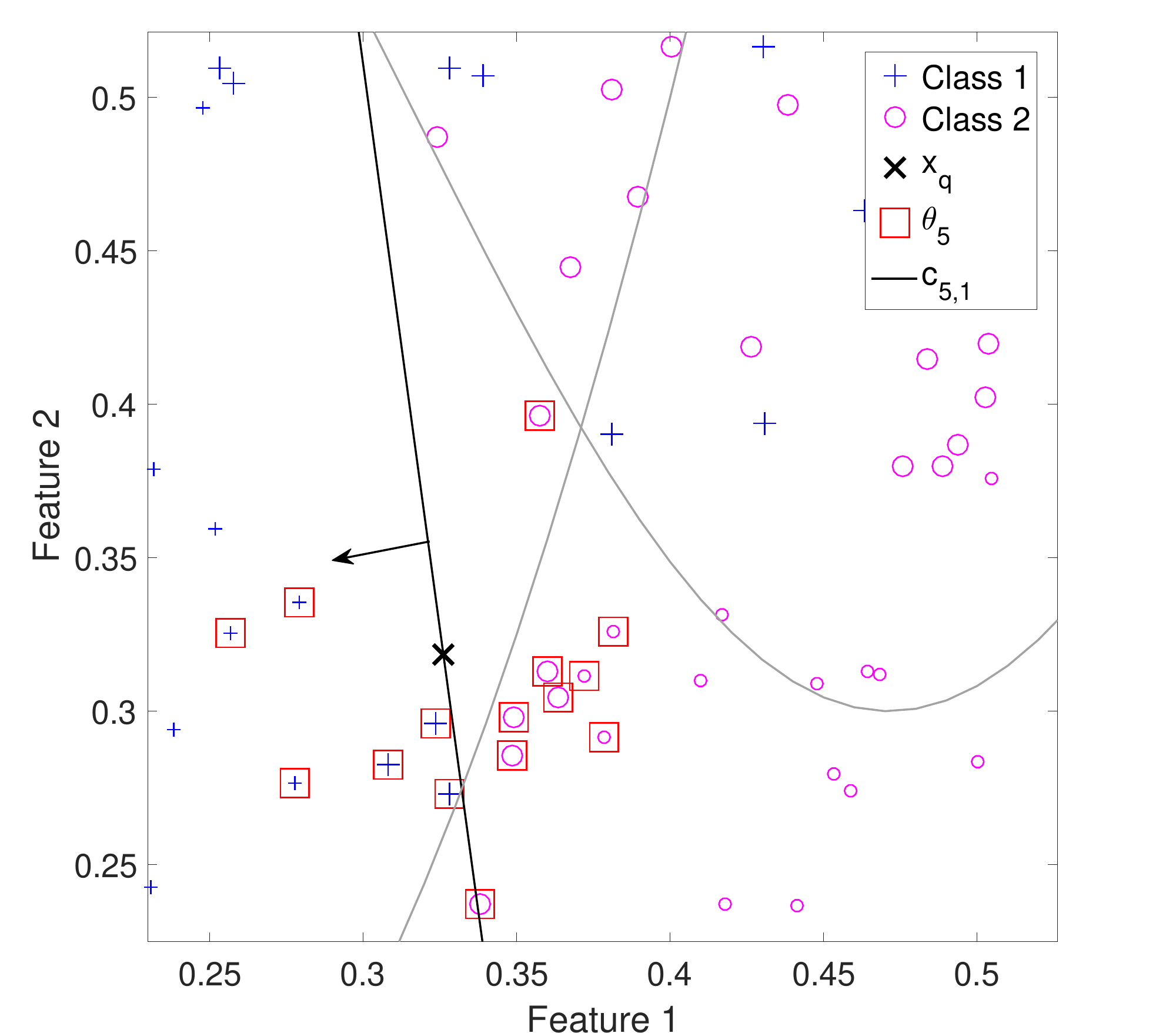}
		}
		\subfloat[]{
		\label{fig:xq-hard-lp6}
			\includegraphics[width=0.45\textwidth]{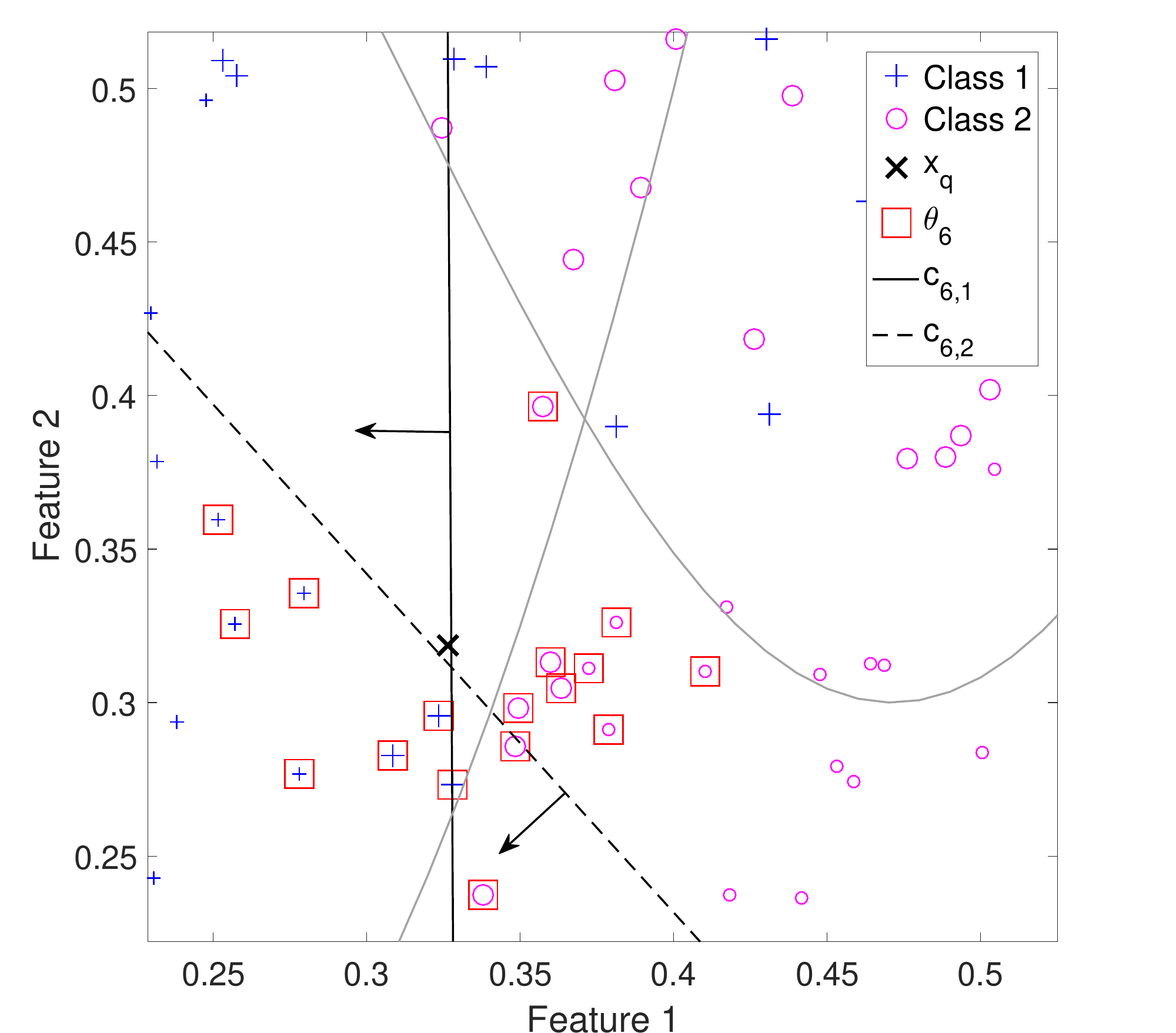}
		}
		\\
		\subfloat[]{
		\label{fig:xq-hard-lp7}
			\includegraphics[width=0.45\textwidth]{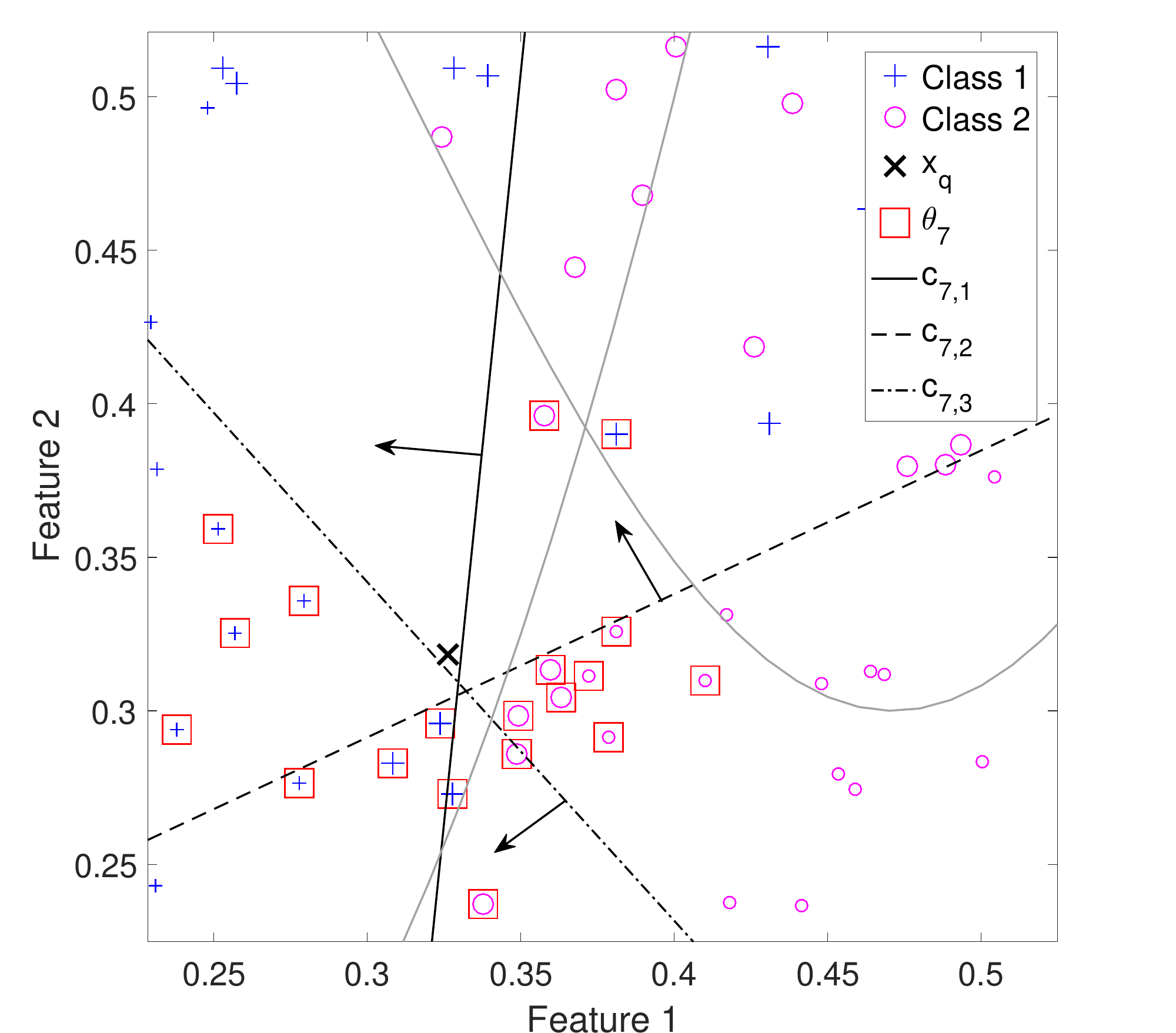}
		}
		\subfloat[]{
		\label{fig:lp-final}
			\includegraphics[width=0.45\textwidth]{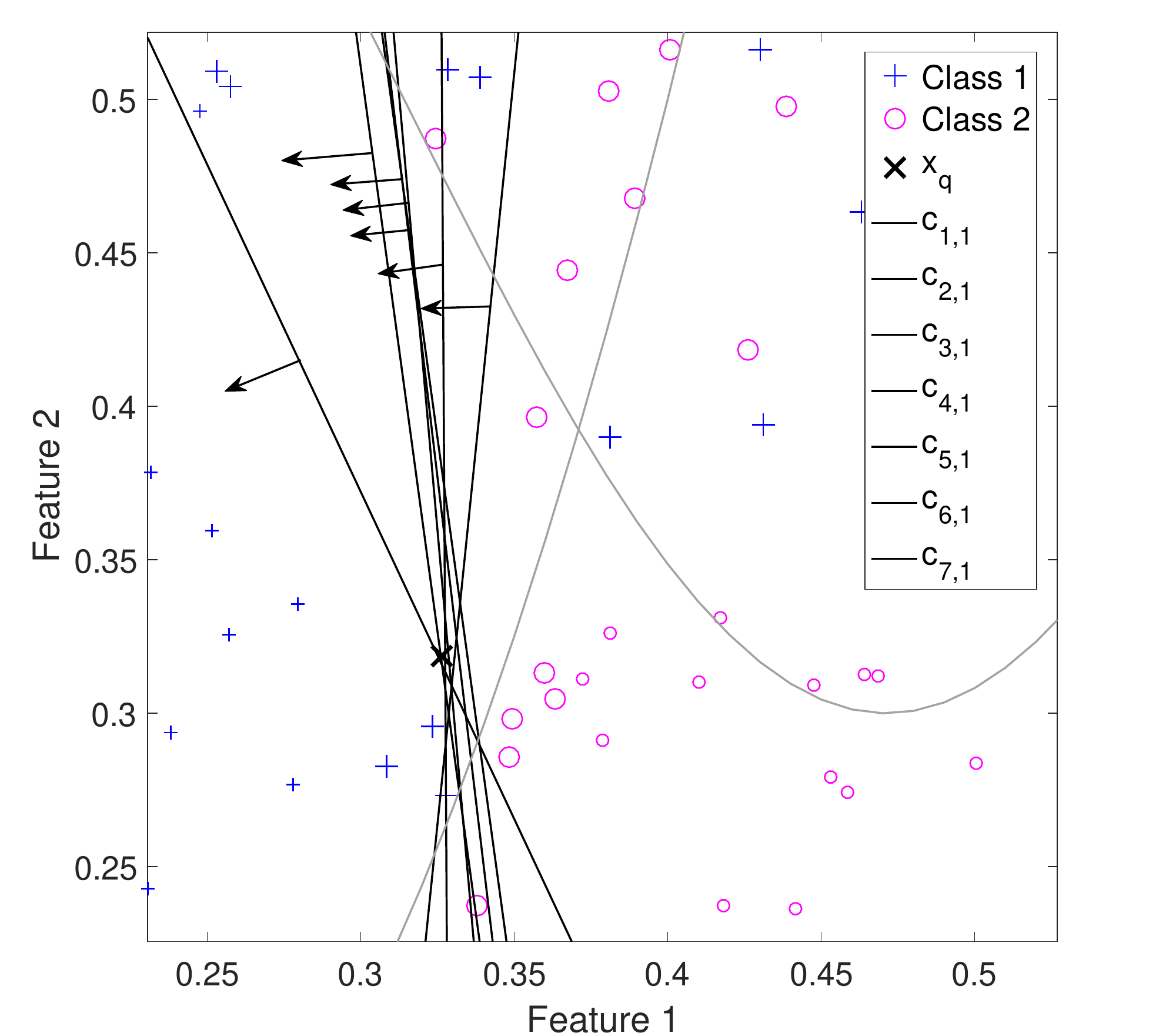}
		}
		\caption{Continued.
		}
\end{figure*}

\par In the fifth iteration, the neighborhood $\theta_{5}$ is obtained with parameter $k_{5} = 15$ in Step 6. 
In Step 7, the SGH method yields the local subpool $C_{5}$, depicted in Figure \ref{fig:xq-hard-lp5}. 
Afterwards, the single classifier $c_{5,1}$ in $C_{5}$ is added to $LP$. 


\par The neighborhood $\theta_{6}$ of the sixth iteration is obtained with $k_{6} = 17$ in Step 6. 
Then, the local subpool $C_{6}$ is generated in Step 7, resulting in two classifiers, $c_{6,1}$ and $c_{6,2}$, as shown in Figure \ref{fig:xq-hard-lp6}. 
In Step 8, both classifiers are evaluated over $\theta_{6}$ using a DCS technique, OLA in this case. 
The most accurate one ($c_{6,1}$) in $C_{6}$ is returned and added to $LP$ in Step 9. 


\begin{table}[!htb]
\centering
\caption{Majority voting of the classifiers from LP for the query instance from Figure \ref{fig:xq-hard}.}
\label{table:votes-toy-prob}
\scriptsize
\begin{tabular}{|c|ccccccc|c|}
\hline
\textbf{}        & \textbf{$c_{1,1}$} & \textbf{$c_{2,1}$} & \textbf{$c_{3,1}$} & \textbf{$c_{4,1}$} & \textbf{$c_{5,1}$} & \textbf{$c_{6,1}$} & \textbf{$c_{7,1}$} & \textbf{Total} \\ \hline
\textbf{Class 1} &                    & x                  & x                  & x                  &                    & x                  & x                  & 5              \\ \hline
\textbf{Class 2} & x                  &                    &                    &                    & x                  &                    &                    & 2              \\ \hline
\end{tabular}
\end{table}

\par In the last iteration, the local subpool $C_{7}$ is generated in Step 7 using the neighborhood $\theta_{7}$ with $k_{7} = 19$ instances. 
Then, the local subpool $C_{7}$ is generated, yielding three classifiers that fully cover $\theta_{7}$. 
Each classifier in $C_{7}$, shown in Figure \ref{fig:xq-hard-lp7}, is then evaluated using OLA, and the one that performs best over $\theta_{7}$ is selected. 
The selected classifier, $c_{7,1}$ in this case, is then added to the local pool, completing the generation process of $LP$, depicted in Figure \ref{fig:lp-final}.


\par After the generation process of the local pool, each classifier in it labels the query instance $\mathbf{x_{q}}$, and the final label is obtained by majority vote in Step 11. Table \ref{table:votes-toy-prob} shows the vote of each classifier in $LP$. 
The final label returned in Step 11 by the local pool is $\omega_{l} = 1$, which is the true class of \looseness=-1 $\mathbf{x_{q}}$.


\section{Experiments}
\label{sec:exp}


\par In order to analyse and evaluate the performance of the proposed method, experiments were conducted over the 20 datasets described in Table \ref{table:datasets}. 
All methods in the comparative study were evaluated using 20 replications of each dataset. 
For the configurations that used pools generated by the SGH method, each replication was randomly split into two parts: 75\% for training and 25\% for test. 
Since the SGH method did not present overfitting, both in global \cite{mariana} and local scope, the training set was used as the DSEL set. 
In the comparative study, however, the methods that use a DS technique were tested using a pool of 100 Perceptrons obtained using Bagging \cite{bagging}, as it is often done in DS works \cite{metades,cruz2015meta}.  
For these configurations, the validation set was used as the DSEL in order to avoid overfitting, so one third of the training set was randomly selected to compose the DSEL \looseness=-1 set.

\par This section is organized as follows.
A comparative study with regards to DCS techniques is performed in Section \ref{sec:comp-dcs}, with the purpose of analyzing whether the use of locally generated pools is in fact advantageous in this context. 
In Section \ref{sec:comp-models}, the performances of the proposed method and state-of-the-art models, including single models, static ensembles and DS techniques, are also compared and analyzed. 
Lastly, the computational complexity of the proposed method and the compared models are discussed in Section \ref{sec:cost}.

\subsection{Comparison with DCS techniques}
\label{sec:comp-dcs}

\par In this section, an experimental analysis on the proposed method is performed. 
The aim of these experiments is to observe whether the DCS techniques are more prone to selecting the best classifier in the pool when said pool is generated locally and whether the use of such pools increase classification rates, in comparison to globally generated pools. 

\par The DCS techniques chosen to evaluate the methods in these experiments were OLA, LCA and MCB, since they outperformed the other evaluated DCS techniques in \cite{cruz2017dynamic}. 
Implementation of these techniques and several other DS techniques, as well as the proposed method itself, can be found on DESLib \cite{deslib}, a dynamic ensemble learning library available at https://github.com/Menelau/DESlib. 
The RoC size $k_{s}$ for each of the DCS techniques is set to 7, since it yielded the best results in \cite{DESsurvey}. 

\par The parameters of the proposed method were set to $k_{s} = k_{h} = 7$ and $M = 7$. 
Moreover, the proposed scheme was tested with two neighborhood acquisition methods: the $LP$ configuration and the $LP^{e}$ configuration. 
In the first, the $getNeighborhood()$ method from Algorithm~\ref{alg:prop-met} (Step 6) used was the regular k-NN. 
In the second configuration, the $getNeighborhood()$ procedure used was a version of the k-Nearest Neighbor Equality (k-NNE) \cite{knne} in which the returned neighborhood contains an equal amount of instances from all classes, given that these classes are present in the query instance's RoC $\theta_{q}$ from Algorithm~\ref{alg:prop-met} (Step 1). 

\par The performance of the proposed method with regards to the DCS techniques is compared to three globally generated pool configurations. 
The baseline method used in the comparison is a Bagging-generated pool composed of 100 classifiers. 
The SGH method over the entire training set is also included in the comparative study, since it provides another global approach for generating classifiers. 
The pool generated by this technique is referenced as the global pool (GP). 
Lastly, another related method, though it is not a generation one, is used in the comparison with DCS techniques: the Frienemy Indecision Region Dynamic Ensemble Selection (FIRE-DES) framework \looseness=-1 \cite{dayvid}. 

\par In the FIRE-DES framework, when a query sample is in an \textit{indecision region}, that is, a neighborhood that contains more than one class, the classifiers that correctly label instances from different classes in the query sample's RoC are pre-selected to form the pool used in the DS technique. 
That is, if a border is detected in the query sample's RoC, the selection scheme searches only among the classifiers that cross this border. 
This is performed using the Dynamic Frienemy Pruning (DFP), an online pruning method for DS techniques. 
The FIRE-DES framework is designed for two-class problems, and it obtained a significant increase in accuracy for most DS techniques, specially for highly imbalanced datasets, in which cases the DFP method provided a considerable improvement in performance for those techniques. 

\par In the FIRE-DES context, an unknown sample in an indecision region has, by definition, a hardness value greater than zero, since at least one of its neighbors belongs to a different class, regardless of its label. 
In the proposed method, such an instance is labelled using the local pool, which is guaranteed to contain classifiers that cross the query sample's RoC due to its generation procedure (Figure \ref{fig:gen-lp}).
Thus, the same idea of using only locally accurate classifiers for instances in overlap regions from the FIRE-DES framework indirectly applies to the proposed method as well. 
Therefore, the FIRE-DES framework, coupled with the chosen DCS techniques, is also included in the comparative study that follows. 
The pool used by this framework in the experiments is the same as the one from the Bagging configuration, which contains 100 globally generated classifiers. 

\par The performance of the chosen configurations with regards to DCS techniques is evaluated in memorization, using the hit rate measure, in Section \ref{sec:hit-rate}, and in generalization, using the accuracy rate over the datasets from Table \ref{table:datasets}, in Section \ref{sec:acc-rate}. 

\subsubsection{Performance in Memorization}
\label{sec:hit-rate}

\par The proposed method was evaluated in memorization using the hit rate \cite{mariana}, which is a metric derived from the SGH method that indicates how well the generated pool integrates with the DCS techniques. 
In the SGH method, since the Oracle accuracy rate over the training set is 100\%, each training instance is assigned to a classifier in the pool that correctly labels it. 
The hit rate is then obtained using the training set as test set, and comparing the chosen classifier to the correct classifier indicated by the SGH method for each training instance. 
Thus, the hit rate is the rate at which the DCS technique selects the correct classifier for a given known instance. 


\par Since the hit rate is defined specifically for pools generated using the SGH method, the hit rate of the proposed method is only compared with the $GP$ configuration, which uses a pool generated by the SGH method with the entire training set as input. 
The hit rate of the proposed configurations are calculated the same way as the $GP$ configuration, with the only difference being for instances not in difficult regions. 
In this case, the accuracy rate of the k-NN rule is used to compute the measure. 
The comparison between the $GP$ and the $LP$ configurations is relevant because it provides the answer to whether or not the generation over a local region instead of over the entire problem is useful in the selection process of a DCS technique.

\par Table \ref{table:hit-rate} shows the mean hit rate for OLA, LCA and MCB of the three configurations that use the SHG method. 
In comparison with the $GP$ configuration, both $LP$ and $LP^{e}$ configurations obtained a greater overall hit rate for both DCS techniques. 
More specifically, for the $LP^{e}$ configuration, nearly half of the datasets yielded a hit rate above $90\%$, whilst for the $GP$ only two of them at most obtained a similar hit rate for the three DCS techniques.  

\par Moreover, a Wilcoxon signed rank test with a significance level of $\alpha = 0.05$ was performed between the hit rate results for the $GP$ and the two proposed configurations. 
It can be observed, from the Wilcoxon rows, that the proposed configurations yielded a significantly greater hit rate than the global configuration for LCA, and, in particular, the $LP^{e}$ configuration obtained a significant increase in the hit rate for OLA and MCB as well. 
This suggests that the use of the local pools indeed facilitates the DCS technique in choosing the correct classifier for instances in difficult \looseness=-1 regions.

\begin{table}[!htb]
	\centering
	\caption{Mean and standard deviation of the hit rate, i.e., the rate at which the right Perceptron is chosen by (a) OLA, (b) LCA and (c) MCB using the $GP$, $LP$ and $LP^{e}$ configurations. 
The row Wilcoxon shows the result of a Wilcoxon signed rank test over the mean hit rates of the GP configuration and the two proposed configurations. 
The significance level was $\alpha = 0.05$, and the symbols $+$, $-$ and $\sim$ indicate the method is significantly superior, inferior or not significantly different, respectively. 
Best results are in bold.}
\label{table:hit-rate}
		\subfloat[]{
		\scriptsize
		    \label{table:hit-rate-ola}
			\scalebox{0.9}{
\begin{tabular}{|c|ccc|}
\hline
\textbf{Dataset} & $\mathbf{GP}$ & $\mathbf{LP}$ & $\mathbf{LP^{e}}$ \\ \hline
 Adult 		 &          {86.91 (0.87)} &         {86.10 (1.13)} & \textbf{91.41 (0.87)}  \\
 Blood 		 &          {79.59 (0.51)} &         {73.11 (1.42)} & \textbf{80.96 (0.88)}  \\
 CTG 		     &          {92.50 (0.59)} &  		 {92.66 (0.83)} & \textbf{92.95 (0.92)}  \\
 Faults 		 &   \textbf{76.88 (1.26)} &         {74.85 (0.70)} &        {70.92 (0.95)}  \\
 German 		 &          {71.05 (1.44)} &         {89.25 (0.56)} & \textbf{91.86 (0.34)}  \\
 Glass 		 &   \textbf{76.21 (1.98)} &         {63.07 (0.75)} &        {69.71 (1.74)}  \\
 Haberman 	     &          {76.26 (1.10)} &         {69.14 (2.45)} & \textbf{77.55 (0.93)}  \\
 Heart 		 &          {84.06 (1.92)} &         {90.06 (1.24)} & \textbf{92.64 (0.54)}  \\
 Ionosphere 	 &          {86.46 (1.48)} &         {87.26 (1.59)} & \textbf{87.97 (1.19)}  \\
 Laryngeal1 	 &          {84.75 (2.07)} &         {87.16 (0.89)} & \textbf{89.82 (1.22)}  \\
 Laryngeal3 	 &          {74.81 (2.95)} &  \textbf{79.63 (1.32)} &        {76.44 (1.20)}  \\
 Liver 		 &          {67.22 (1.40)} &         {79.16 (1.28)} & \textbf{83.92 (0.89)}  \\
 Mammographic   &          {82.72 (0.64)} &         {70.32 (1.61)} & \textbf{84.02 (0.84)}  \\
 Monk2 		 &          {85.77 (3.60)} &         {95.85 (0.32)} & \textbf{96.47 (0.33)}  \\
 Phoneme 	     &          {87.40 (0.46)} &         {88.56 (0.23)} & \textbf{90.09 (0.29)}  \\
 Pima 		     &          {75.64 (1.55)} &         {83.00 (0.63)} & \textbf{87.65 (0.28)}  \\
 Sonar 		 &          {80.00 (3.62)} &         {92.48 (1.11)} & \textbf{93.81 (1.13)}  \\
 Vehicle 	     &          {76.14 (1.49)} &  \textbf{78.25 (1.00)} &        {77.32 (0.70)}  \\
 Vertebral 	 &          {82.39 (2.14)} &         {87.42 (1.27)} & \textbf{89.38 (1.02)}  \\
 Weaning 	     &          {83.45 (1.33)} &         {94.82 (0.45)} & \textbf{94.97 (0.35)}  \\ \hline
Average         &  80.51       &  83.11            &  \textbf{86.00}            \\ \hline
Wilcoxon        &  n/a      &  $\sim$    &  +            \\ \hline
\end{tabular}}
		}
	\subfloat[]{
	\scriptsize
		    \label{table:hit-rate-lca}
			\scalebox{0.9}{
\begin{tabular}{|c|ccc|}
\hline
\textbf{Dataset} & $\mathbf{GP}$ & $\mathbf{LP}$ & $\mathbf{LP^{e}}$ \\ \hline
 Adult 		 &        {86.77 (0.92)} &        {89.99 (0.60)} & \textbf{91.27 (0.89)}   \\
 Blood 		 &        {80.20 (0.35)} &        {79.43 (1.35)} & \textbf{80.27 (1.20)}   \\
 CTG 		     &        {92.63 (0.44)} & \textbf{94.03 (0.27)} &        {93.30 (0.34)}   \\
 Faults 		 & \textbf{76.84 (1.01)} &        {74.85 (0.39)} &        {71.25 (0.61)}   \\
 German 		 &        {75.75 (1.35)} &        {90.01 (0.63)} & \textbf{91.90 (0.32)}   \\
 Glass 		 & \textbf{77.95 (1.92)} &        {67.73 (1.37)} &        {69.17 (2.00)}   \\
 Haberman 	     & \textbf{76.61 (1.46)} &        {74.76 (1.84)} & 		  {76.60 (1.04)}   \\
 Heart 		 &        {83.86 (2.40)} &        {91.85 (0.79)} & \textbf{92.77 (0.68)}   \\
 Ionosphere 	 &        {87.34 (1.53)} & \textbf{92.32 (0.77)} &        {92.11 (0.75)}   \\
 Laryngeal1 	 &        {84.81 (2.38)} &        {88.55 (0.94)} & \textbf{89.83 (1.14)}   \\
 Laryngeal3 	 &        {73.98 (1.99)} & \textbf{80.28 (1.72)} &        {78.37 (2.01)}   \\
 Liver 		 &        {70.62 (2.91)} &        {79.69 (1.26)} & \textbf{84.10 (1.01)}   \\
 Mammographic   & \textbf{82.83 (1.54)} &        {80.82 (1.15)} &        {82.07 (0.85)}   \\
 Monk2 		 &        {91.82 (3.61)} & \textbf{96.63 (0.27)} & 		  {96.44 (0.32)}   \\
 Phoneme 	     &        {89.48 (0.44)} & \textbf{91.93 (0.32)} &        {91.31 (0.23)}   \\
 Pima 		     &        {76.02 (1.67)} &        {84.07 (0.48)} & \textbf{87.68 (0.27)}   \\
 Sonar 		 &        {83.46 (3.45)} &        {93.37 (1.08)} & \textbf{94.27 (0.96)}   \\
 Vehicle 	     & \textbf{77.98 (1.57)} & 		  {77.11 (0.78)} &        {76.60 (0.77)}   \\
 Vertebral 	 &        {84.33 (2.32)} & \textbf{89.87 (1.02)} &        {89.85 (1.00)}   \\
 Weaning 	     &        {84.38 (1.72)} & \textbf{95.17 (0.51)} &        {95.07 (0.36)}   \\ \hline
Average         &  81.88       &   85.63      &   \textbf{86.21}       \\ \hline
Wilcoxon        &  n/a      &  +            &  +            \\ \hline
\end{tabular}}
		}
		\end{table}	
		
\begin{table}[!htb]
	\ContinuedFloat
	\phantomcaption
	\centering
		\subfloat[]{
		\scriptsize
		    \label{table:hit-rate-mcb}
			\scalebox{0.9}{
\begin{tabular}{|c|ccc|}
\hline
\textbf{Dataset} & $\mathbf{GP}$ & $\mathbf{LP}$ & $\mathbf{LP^{e}}$ \\ \hline
Adult 		 &        {87.14 (0.73)} &         {87.35 (1.19)} &    \textbf{89.76 (0.59)}   \\
Blood 		 &        {79.61 (0.51)} &         {74.17 (1.48)} &    \textbf{79.67 (0.81)}   \\
CTG 		     & \textbf{92.49 (0.63)} &         {92.14 (0.44)} &           {92.00 (0.37)}   \\
Faults 		 &        {76.87 (1.26)} &  \textbf{77.16 (0.84)} &           {76.73 (0.49)}   \\
German 		 &        {71.23 (1.47)} &         {90.90 (0.61)} &    \textbf{91.93 (0.58)}   \\
Glass 		 & \textbf{76.27 (1.99)} &         {66.54 (1.44)} &           {69.98 (1.61)}   \\
Haberman 	     &        {76.35 (1.10)} &         {69.03 (1.65)} &    \textbf{76.77 (1.59)}   \\
Heart 		 &        {83.96 (1.72)} &         {89.10 (1.31)} &    \textbf{91.85 (1.17)}   \\
Ionosphere 	 &        {86.43 (1.43)} &         {88.65 (0.77)} &    \textbf{92.14 (0.75)}   \\
Laryngeal1 	 &        {84.75 (1.93)} &         {86.72 (1.11)} &    \textbf{88.71 (0.90)}   \\
Laryngeal3 	 &        {74.85 (2.90)} &         {78.97 (1.48)} &    \textbf{80.79 (1.37)}   \\
Liver 		 &        {67.34 (1.28)} &         {79.60 (1.24)} &    \textbf{83.68 (1.10)}   \\
Mammographic   & \textbf{82.68 (0.73)} &         {71.21 (1.48)} &           {82.56 (1.04)}   \\
Monk2 		 &        {86.67 (4.48)} &  \textbf{95.69 (0.28)} &           {95.35 (0.31)}   \\
Phoneme 	     &        {87.40 (0.47)} &         {89.12 (0.27)} &    \textbf{90.10 (0.18)}   \\
Pima 		     &        {75.82 (1.83)} &         {83.52 (0.79)} &    \textbf{87.54 (0.31)}   \\
Sonar 		 &        {80.19 (3.63)} &         {92.51 (0.91)} &    \textbf{93.89 (1.03)}   \\
Vehicle 	     &        {76.20 (1.51)} &  \textbf{77.90 (0.78)} &           {76.05 (0.76)}   \\
Vertebral 	 &        {82.39 (2.19)} &         {88.06 (1.27)} &    \textbf{89.63 (1.24)}   \\
Weaning 	     &        {83.36 (1.20)} &  \textbf{94.06 (0.51)} &           {93.88 (0.72)}   \\ \hline
Average         &  80.60       &   83.62     &    \textbf{86.15}      \\ \hline
Wilcoxon          &  n/a      &  $\sim$           &  +            \\ \hline
\end{tabular}}
		}
\end{table}	

%

\subsubsection{Performance in Generalization}
\label{sec:acc-rate}

\par
\begin{figure*}[!htb]
		\centering
		\centerline{		
		\includegraphics[width=1\textwidth]{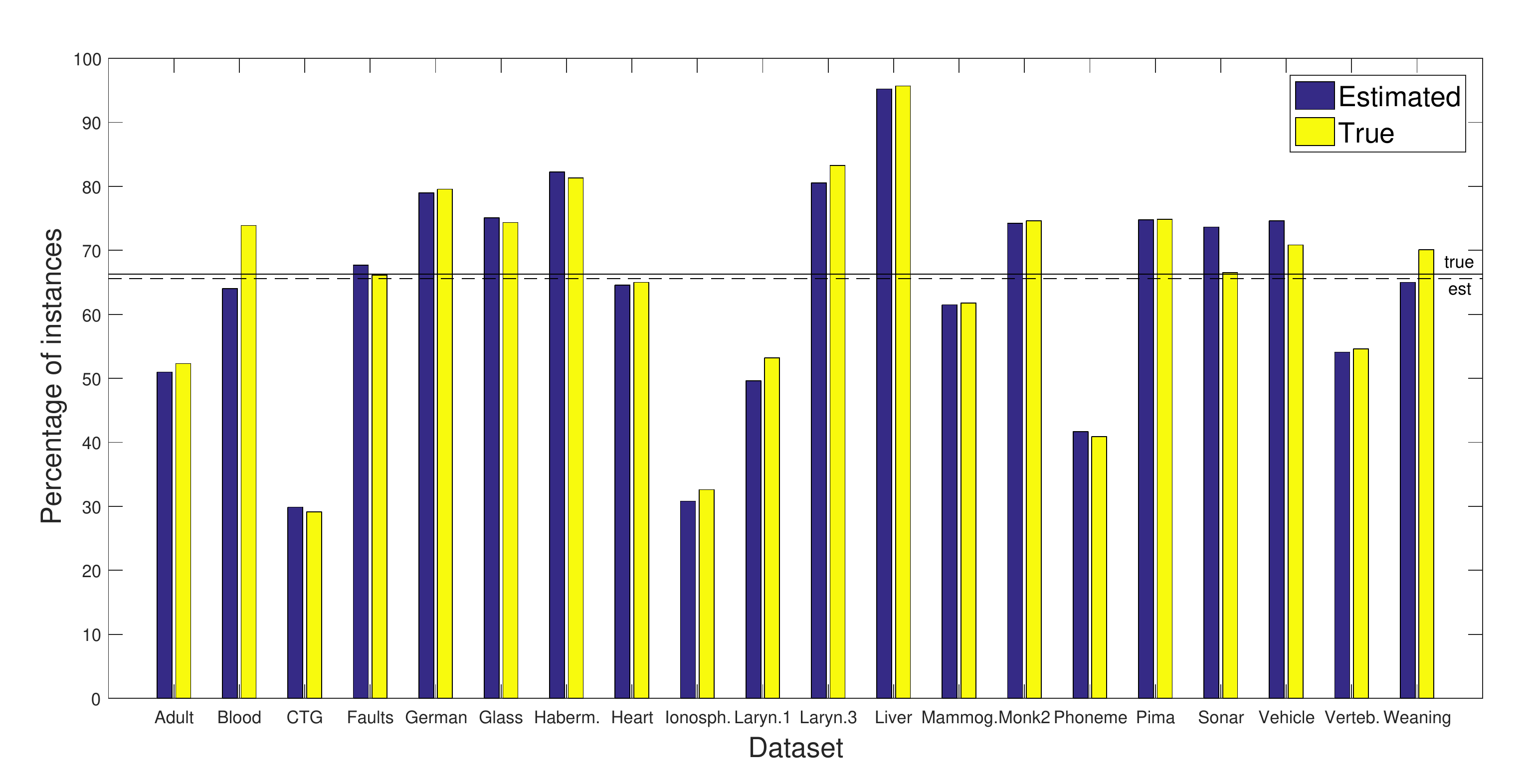}}
		\caption{Mean percentage of instances in difficult regions for all datasets from Table \ref{table:datasets}. 
		The \textit{Estimated} bar indicates the times the local pool was used to classify an instance, while the \textit{True} bar indicates the percentage of instances with true kDN value above zero. 
		The lines $true$ and $est$ indicate the averaged values of all datasets for the estimated and true percentage of hard instances, respectively.}
		\label{fig:true-est-kdn}
\end{figure*}

 The mean percentage of test instances with true hardness value above zero is depicted in the \textit{True} bars of Figure \ref{fig:true-est-kdn} for all datasets. 
The true hardness value is obtained observing the neighborhood of each test instance over the entire dataset. 
The mean percentage of test instances deemed hard by the proposed method is also depicted in Figure \ref{fig:true-est-kdn} (\textit{Estimated} bars). 
That is, the \textit{Estimated} bars show the frequency at which the proposed method generated and used local pools, whilst the \textit{True} bars show the actual proportion of instances in difficult regions for each problem. 
It can be observed that, though the proportion of instances in difficult regions varies greatly from problem to problem, the proposed method was mostly able to identify in which cases the query sample was truly located in a difficult region and thus generated a local pool to handle them.

\par The averaged value of the true and estimated percentage of hard instances is also indicated in Figure \ref{fig:true-est-kdn} by the $true$ and $est$ lines, respectively.  
It can be observed that the mean percentage of test instances truly located near the borders was $65.04\%$, while the proposed method generated local pools for $64.46\%$ of the test instances, on average. 

\par The accuracy rate of Bagging, FIRE-DES, the GP configuration and the proposed configurations were evaluated with OLA, LCA and MCB, and the results are presented in Table \ref{table:acc-rate-dcs}.
It can be observed that the proposed configurations ($LP$ and $LP^{e}$) obtained an average accuracy rate greater than Bagging, FIRE-DES and the $GP$ configuration for all DCS techniques. 
A Wilcoxon signed-rank test with a significance level of $\alpha = 0.05$ was also performed for comparing the accuracy of the evaluated techniques. 
We chose the Wilcoxon signed-rank test due to its robustness, as its result do not depend on the algorithms originally included in the comparison \cite{benavoli16a}.
It can be observed from the Wilcoxon rows of Table \ref{table:acc-rate-lca} that both proposed configurations were significantly superior to Bagging, and the $LP$ configuration significantly outperformed the $GP$ configuration \looseness=-1  using LCA. 

\begin{table}[!htb]
\centering
\caption{Mean and standard deviation of the accuracy rate of  using (a) OLA, (b) LCA and (c) MCB for a pool with 100 Perceptrons generated using Bagging (column Bagging), a pool of 100 Perceptrons generated using Bagging and pruned with the DFP method (column FIRE-DES), the $GP$ configuration, the $LP$ configuration and the $LP^{e}$ configuration.
The row \textit{Wilcoxon (Bagging)} shows the result of a Wilcoxon signed rank test over the mean accuracy rates of Bagging and each remaining method. 
The same test was performed in comparison with the FIRE-DES configuration and the $GP$ configuration (rows \textit{Wilcoxon (FIRE)} and \textit{Wilcoxon (GP)}, respectively).
The significance level was $\alpha = 0.05$, and the symbols $+$, $-$ and $\sim$ indicate the method is significantly superior, inferior or not significantly different, respectively.
The row \textit{Avg rank} shows the resulting mean ranks of a Friedman test with a significance level of $\alpha = 0.05$, and the p-value of the test is shown in row \textit{p-value}. 
In $LP^{mc}$, the $LP^{e}$ is used for 2-class problems, while the $LP$ is used for multi-class ones.
Best results \looseness=-1 are in bold.}
\label{table:acc-rate-dcs}
\subfloat[]{
	\label{table:acc-rate-ola}
	\scriptsize
	\scalebox{0.9}{
	\begin{tabular}{|c|ccccc|c|}
	\hline
	\textbf{Dataset}             & \textbf{Bagging \cite{bagging}} & \textbf{FIRE-DES \cite{dayvid}} & \textbf{GP} & $\mathbf{LP}$ & $\mathbf{LP^{e}}$ & $\mathbf{LP^{mc}}$ \\ \hline
	Adult 		                 &        {84.97 (2.50)} &         {83.84 (3.37)} & \textbf{88.15 (2.93)} &         {83.32 (3.63)} &          {87.75 (2.17)} &            {87.75 (2.17)} \\
	Blood 		                 &        {75.48 (2.31)} &         {69.28 (3.90)} &        {75.53 (1.14)} &         {73.78 (2.80)} &   \textbf{77.21 (1.57)} &     \textbf{77.21 (1.57)} \\
	CTG 		                 &        {88.83 (1.26)} &         {88.50 (1.36)} &        {90.24 (0.77)} &  \textbf{92.20 (1.10)} &          {89.98 (0.80)} &     \textbf{92.20 (1.10)} \\
	Faults 	             	     &        {66.52 (1.65)} &         {65.33 (1.95)} &        {71.91 (1.60)} &  \textbf{72.40 (1.29)} &          {65.93 (1.29)} &     \textbf{72.40 (1.29)} \\
	German 	                     &        {70.34 (1.88)} &         {68.56 (1.89)} &        {70.04 (2.35)} &         {72.16 (2.12)} &   \textbf{74.04 (1.88)} &     \textbf{74.04 (1.88)} \\
	Glass 		                 &        {61.42 (4.22)} &         {59.43 (5.66)} &        {66.79 (4.17)} &  \textbf{67.83 (3.94)} &          {60.75 (2.17)} &     \textbf{67.83 (3.94)} \\
	Haberman 	                 &        {70.79 (5.12)} &         {66.78 (4.81)} &        {71.58 (5.24)} &         {68.95 (4.19)} &   \textbf{72.43 (2.24)} &            {72.43 (2.24)} \\
	Heart 		                 &        {82.35 (3.44)} &         {82.21 (4.32)} & \textbf{86.62 (2.18)} &         {81.99 (4.79)} &          {83.68 (3.27)} &            {83.68 (3.27)} \\
	Ionosphere                   &        {86.70 (3.04)} &         {86.53 (2.84)} &        {87.16 (2.76)} &         {91.76 (1.95)} &   \textbf{91.99 (2.16)} &     \textbf{91.99 (2.16)} \\
	Laryngeal1                   & \textbf{82.92 (3.54)} &         {81.89 (5.62)} &        {80.38 (4.26)} &         {77.74 (5.53)} &          {80.57 (5.87)} &            {80.57 (5.87)} \\
	Laryngeal3                   &        {70.73 (5.79)} &         {66.46 (4.73)} & \textbf{72.25 (1.71)} &         {71.74 (2.73)} &          {64.66 (1.34)} &            {71.74 (2.73)} \\
	Liver 		                 &        {64.59 (4.18)} &         {65.00 (3.85)} &        {58.37 (3.53)} &         {60.12 (4.99)} &   \textbf{67.21 (1.67)} &     \textbf{67.21 (1.67)} \\
	Mammographic                 &        {82.57 (2.02)} &         {79.06 (3.62)} & \textbf{82.60 (2.47)} &         {75.53 (2.60)} &          {82.36 (1.81)} &            {82.36 (1.81)} \\
	Monk2 		                 &        {87.87 (3.97)} &         {87.92 (3.13)} &        {86.20 (3.74)} &  \textbf{94.91 (0.97)} &          {94.07 (0.76)} &            {94.07 (0.76)} \\
	Phoneme 	                 &        {80.31 (0.68)} &         {76.02 (1.17)} &        {86.74 (0.73)} &  \textbf{88.58 (0.63)} &          {86.60 (0.67)} &            {86.60 (0.67)} \\
	Pima 		                 &        {72.40 (2.73)} &         {68.78 (3.03)} &        {72.29 (2.39)} &         {72.03 (1.68)} &   \textbf{76.77 (2.26)} &     \textbf{76.77 (2.26)} \\
	Sonar 		                 &        {80.96 (4.04)} &         {79.90 (4.02)} &        {80.00 (3.33)} &  \textbf{83.27 (5.79)} &          {75.00 (4.14)} &            {75.00 (4.14)} \\
	Vehicle 	                 &        {73.61 (2.56)} &  \textbf{74.43 (1.95)} &        {70.09 (2.57)} &         {74.39 (2.09)} &          {69.74 (1.66)} &            {74.39 (2.09)} \\
	Vertebral 	                 &        {85.38 (4.04)} &         {84.49 (4.70)} &        {81.41 (2.06)} &         {85.19 (2.14)} &   \textbf{86.47 (2.65)} &     \textbf{86.47 (2.65)} \\
	Weaning 	                 &        {77.50 (3.36)} &         {77.57 (3.40)} &        {78.68 (3.71)} &  \textbf{86.05 (1.73)} &          {85.66 (2.37)} &            {85.66 (2.37)} \\ \hline
	\textbf{Average}             &         77.31         &          75.59    	  &         77.85         &  \textbf{78.69}        &            78.64         &             80.02         \\ \hline
	\textbf{Wilcoxon (Bagging)}  &           n/a         &            -           &      $\sim$           &           $\sim$        &          $\sim$           &          +             \\ \hline
	\textbf{Wilcoxon (FIRE)} &           +           &     n/a               &        $\sim$          &           +            &            +              &             +              \\ \hline
	\textbf{Wilcoxon (GP)}      &         $\sim$         &     $\sim$             &      n/a			&           $\sim$         &        $\sim$        &            +              \\ \hline
	\textbf{Avg rank}            &        3.850        &      4.900           &      3.450       &        3.375        &        3.125         &      2.300              \\ \hline
	\textbf{p-value}             & \multicolumn{6}{c|}{$6.09 \times 10^{-4}$}                                                                                                             \\ \hline
	\end{tabular}
	}}
\end{table}

\begin{table}[!htb]
\ContinuedFloat
\phantomcaption
\scriptsize
\centering
\subfloat[]{
	\label{table:acc-rate-lca}
	\scalebox{0.9}{
	\begin{tabular}{|c|ccccc|c|}
	\hline
	\textbf{Dataset}             & \textbf{Bagging \cite{bagging}} & \textbf{FIRE-DES \cite{dayvid}} & \textbf{GP} & $\mathbf{LP}$ & $\mathbf{LP^{e}}$ &$\mathbf{LP^{mc}}$ \\ \hline
	Adult 		                 &        {86.88 (3.17)} &         {85.72 (3.59)} & \textbf{87.40 (2.82)} &         {84.71 (3.73)} &         {87.11 (2.40)} &             {87.11 (2.40)} \\
	Blood 		                 &        {76.14 (2.24)} &         {71.06 (3.44)} &        {75.74 (1.04)} &  \textbf{77.95 (2.51)} &         {76.89 (1.67)} &             {76.89 (1.67)} \\
	CTG 		                 &        {88.38 (1.37)} &         {88.18 (1.36)} &        {90.30 (0.84)} &  \textbf{92.22 (1.10)} &         {90.58 (0.39)} &      \textbf{92.22 (1.10)} \\
	Faults 	             	     &        {66.00 (1.69)} &         {65.67 (2.23)} &		   {71.99 (1.53)} &  \textbf{73.20 (1.22)} &         {66.28 (1.15)} &      \textbf{73.20 (1.22)} \\
	German 	                     &        {70.66 (2.06)} &         {70.40 (1.22)} &        {70.84 (1.87)} &         {72.88 (2.37)} &  \textbf{74.08 (1.84)} &      \textbf{74.08 (1.84)} \\
	Glass 		                 &        {56.13 (5.47)} &         {56.04 (5.41)} & \textbf{69.43 (3.33)} &         {67.45 (2.73)} &         {62.55 (4.83)} &             {67.45 (2.73)} \\
	Haberman 	                 &        {73.03 (3.58)} &         {69.87 (4.87)} &        {71.05 (1.91)} &         {70.79 (3.71)} &  \textbf{72.11 (2.12)} &      \textbf{72.11 (2.12)} \\
	Heart 		                 &        {82.35 (4.84)} &         {82.21 (4.77)} & \textbf{86.47 (2.85)} &         {82.50 (5.54)} &         {83.09 (3.32)} &             {83.09 (3.32)} \\
	Ionosphere                   &        {86.14 (4.90)} &         {86.19 (4.70)} &        {87.27 (3.21)} &         {91.53 (1.45)} &  \textbf{92.44 (2.56)} &      \textbf{92.44 (2.56)} \\
	Laryngeal1                   &        {81.23 (2.70)} &         {80.09 (3.80)} & \textbf{80.94 (4.70)} &         {79.25 (5.05)} &         {80.57 (5.87)} &             {80.57 (5.87)} \\
	Laryngeal3                   &        {71.57 (5.25)} &         {68.88 (6.19)} &        {72.58 (2.14)} &  \textbf{73.48 (2.48)} &         {67.42 (1.86)} &      \textbf{73.48 (2.48)} \\
	Liver 		                 &        {64.59 (4.87)} &         {66.74 (2.70)} &        {58.37 (2.81)} &         {62.21 (5.26)} &  \textbf{66.98 (1.79)} &      \textbf{66.98 (1.79)} \\
	Mammographic                 &        {82.00 (3.18)} &         {78.89 (4.09)} &        {81.63 (3.06)} &         {80.05 (1.84)} &  \textbf{82.57 (1.77)} &             {82.57 (1.77)} \\
	Monk2 		                 &        {86.06 (3.06)} &         {85.88 (3.16)} &        {90.28 (2.18)} &  \textbf{94.91 (0.97)} &         {94.07 (0.76)} &             {94.07 (0.76)} \\
	Phoneme 	                 &        {80.78 (0.65)} &         {77.26 (0.89)} &        {87.01 (0.77)} &  \textbf{89.18 (0.50)} &         {86.62 (0.69)} &             {86.62 (0.69)} \\
	Pima 		                 &        {74.66 (2.39)} &         {72.19 (3.63)} &        {73.23 (3.39)} &         {73.46 (1.01)} &  \textbf{76.74 (2.24)} &      \textbf{76.74 (2.24)} \\
	Sonar 		                 &        {76.35 (5.41)} &         {75.87 (4.93)} &        {78.08 (5.01)} &  \textbf{82.98 (5.27)} &         {76.35 (3.64)} &             {76.35 (3.64)} \\
	Vehicle 	                 &        {72.03 (1.63)} &         {72.41 (1.86)} &        {70.75 (2.22)} &  \textbf{73.51 (1.64)} &         {71.34 (1.28)} &      \textbf{73.51 (1.64)} \\
	Vertebral 	                 &        {84.55 (3.42)} &         {85.51 (3.30)} &        {82.31 (1.93)} &         {85.32 (2.68)} &  \textbf{86.47 (2.65)} &      \textbf{86.47 (2.65)} \\
	Weaning 	                 &        {73.88 (2.78)} &         {73.75 (3.51)} &        {78.82 (3.05)} &  \textbf{86.51 (1.90)} &         {85.66 (2.37)} &             {85.66 (2.37)} \\ \hline
	\textbf{Average}             &         76.67    &               75.64 &         		78.22   &                \textbf{79.70}    &               78.99      &                 80.08     \\ \hline
	\textbf{Wilcoxon (Bagging)}  &           n/a         &            -           &      $\sim$           &           +        &          +           &          +             \\ \hline
	\textbf{Wilcoxon (FIRE)} &           +           &     n/a               &        +          &           +            &            +              &             +              \\ \hline
	\textbf{Wilcoxon (GP)}      &         $\sim$         &     -             &      n/a			&           +         &        $\sim$        &            +              \\ \hline
	\textbf{Avg rank}            &        4.150         &    5.350          &     3.550        &        2.925          &     2.850               &           2.175         \\ \hline
	\textbf{p-value}             & \multicolumn{6}{c|}{$4.63 \times 10^{-7}$}                                                                                                             \\ \hline
	\end{tabular}
	}}
\end{table}

\begin{table}[!htb]
\ContinuedFloat
\phantomcaption
\scriptsize
\centering
\subfloat[]{
	\label{table:acc-rate-mcb}
	\scalebox{0.9}{
	\begin{tabular}{|c|ccccc|c|}
	\hline
	\textbf{Dataset}             & \textbf{Bagging \cite{bagging}} & \textbf{FIRE-DES \cite{dayvid}} & \textbf{GP} & $\mathbf{LP}$ & $\mathbf{LP^{e}}$ & $\mathbf{LP^{mc}}$ \\ \hline
	Adult 		                 &         {85.00 (2.53)} &        {83.70 (3.25)} & \textbf{88.15 (2.93)} &        {84.45 (3.98)} &         {87.75 (2.18)} &              {87.75 (2.18)} \\
	Blood 		                 &         {75.16 (2.07)} &        {68.88 (3.24)} &        {75.53 (1.14)} & \textbf{76.99 (2.15)} &         {76.57 (1.90)} &              {76.57 (1.90)} \\
	CTG 		                 &         {88.87 (1.24)} &        {88.61 (1.54)} &        {90.24 (0.77)} & \textbf{92.10 (1.20)} &         {90.23 (0.36)} &       \textbf{92.10 (1.20)} \\
	Faults 	             	     &         {66.58 (1.37)} &        {65.77 (2.32)} &   	   {71.91 (1.60)} & \textbf{72.80 (1.29)} &         {66.24 (0.86)} &       \textbf{72.80 (1.29)} \\
	German 	                     &         {70.16 (2.41)} &        {68.94 (2.72)} &        {70.52 (2.08)} &        {72.84 (2.36)} &  \textbf{74.08 (1.84)} &       \textbf{74.08 (1.84)} \\
	Glass 		                 &         {60.00 (5.83)} &        {60.19 (6.45)} & \textbf{66.79 (4.17)} &        {66.42 (4.27)} &         {61.04 (2.89)} &              {66.42 (4.27)} \\
	Haberman 	                 &         {72.17 (5.87)} &        {68.36 (4.95)} &        {71.71 (4.91)} &        {69.80 (2.85)} &  \textbf{72.37 (2.67)} &              {72.37 (2.67)} \\
	Heart 		                 &         {81.10 (4.11)} &        {81.32 (4.82)} & \textbf{86.18 (2.36)} &        {82.06 (4.86)} &         {83.09 (3.32)} &              {83.09 (3.32)} \\
	Ionosphere                   &         {88.24 (2.56)} &        {86.36 (2.55)} &        {87.16 (2.71)} &        {91.48 (1.40)} &  \textbf{92.16 (2.39)} &       \textbf{92.16 (2.39)} \\
	Laryngeal1                   &  \textbf{83.02 (4.29)} &        {81.51 (6.44)} &        {80.57 (4.59)} &        {78.30 (5.03)} &         {80.47 (5.75)} &              {80.47 (5.75)} \\
	Laryngeal3                   &         {70.96 (5.62)} &        {67.19 (4.71)} &        {71.80 (1.58)} & \textbf{72.19 (2.65)} &         {66.18 (1.41)} &       \textbf{72.19 (2.65)} \\
	Liver 		                 &         {61.98 (4.86)} &        {63.49 (5.01)} &        {58.37 (3.49)} &        {61.34 (4.71)} &  \textbf{67.03 (1.32)} &              {67.03 (1.32)} \\
	Mammographic                 &         {82.31 (2.32)} &        {79.01 (3.40)} & \textbf{82.60 (2.47)} &        {78.87 (2.55)} &         {82.52 (1.65)} &              {82.52 (1.65)} \\
	Monk2 		                 &         {88.06 (4.18)} &        {87.69 (4.29)} &        {87.96 (3.80)} & \textbf{94.91 (0.97)} &         {94.07 (0.76)} &              {94.07 (0.76)} \\
	Phoneme 	                 &         {80.53 (0.79)} &        {76.12 (1.05)} &        {86.73 (0.73)} & \textbf{88.98 (0.56)} &         {86.68 (0.74)} &              {86.68 (0.74)} \\
	Pima 		                 &         {72.73 (2.60)} &        {68.36 (2.96)} &        {72.71 (2.67)} &        {72.92 (1.56)} &  \textbf{76.74 (2.28)} &       \textbf{76.74 (2.28)} \\
	Sonar 		                 &         {80.67 (4.11)} &        {80.29 (4.13)} &        {79.81 (3.09)} & \textbf{83.08 (5.42)} &         {76.15 (3.33)} &              {76.15 (3.33)} \\
	Vehicle 	                 &         {73.30 (2.54)} &        {74.58 (2.45)} &        {70.14 (2.52)} & \textbf{74.88 (1.57)} &         {70.75 (1.65)} &       \textbf{74.88 (1.57)} \\
	Vertebral 	                 &         {84.55 (4.75)} &        {85.38 (4.86)} &        {82.69 (2.22)} &        {85.58 (2.38)} &  \textbf{86.41 (2.71)} &       \textbf{86.41 (2.71)} \\
	Weaning 	                 &         {76.38 (2.43)} &        {76.05 (3.10)} &        {79.21 (3.30)} & \textbf{86.38 (1.72)} &         {85.59 (2.36)} &              {85.59 (2.36)} \\ \hline
	\textbf{Average}             &          77.08        &          75.59 		    &       78.03		  &         \textbf{79.31}         &          78.80           &             80.00         \\ \hline
	\textbf{Wilcoxon (Bagging)}  &           n/a         &            -           &      $\sim$           &          +       &          $\sim$           &          +             \\ \hline
	\textbf{Wilcoxon (FIRE)} &           +           &     n/a               &        $\sim$          &           +            &            +              &             +              \\ \hline
	\textbf{Wilcoxon (GP)}      &         $\sim$         &     $\sim$             &      n/a			&           $\sim$         &        $\sim$        &            +              \\ \hline
	\textbf{Avg rank}            &       4.200          &      5.050            &     3.550         &       2.825           &        3.075             &       2.300              \\ \hline
	\textbf{p-value}             & \multicolumn{6}{c|}{$2.07 \times 10^{-5}$}                                                                                                             \\ \hline
	\end{tabular}
	}}
\end{table}

%
\par Also, it can be observed in Table \ref{table:acc-rate-dcs} that, for two-class problems with high percentage of difficult instances such as German, Liver, Monk2, Pima and Sonar (Figure \ref{fig:true-est-kdn}), the use of local pools fairly increased the accuracy rate in comparison with the other configurations for the three DCS techniques, further suggesting the advantage of such pools over the global one for instances in \looseness=-1 difficult regions.

\par
\begin{figure*}[!htbp]
		\centering
		\centerline{
		\subfloat[]{
			\includegraphics[scale=0.56]{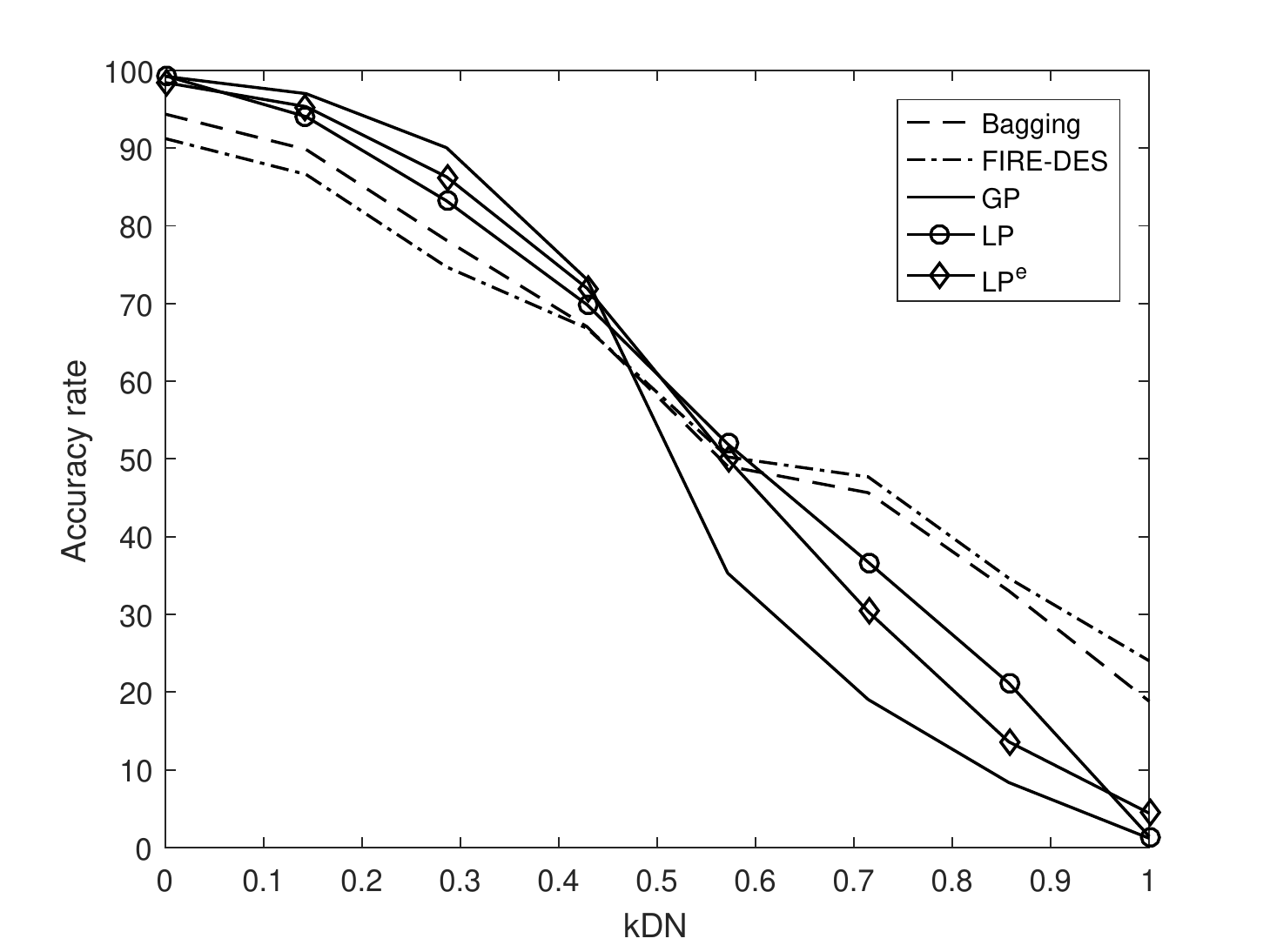}
		}
		\subfloat[]{
			\includegraphics[scale=0.56]{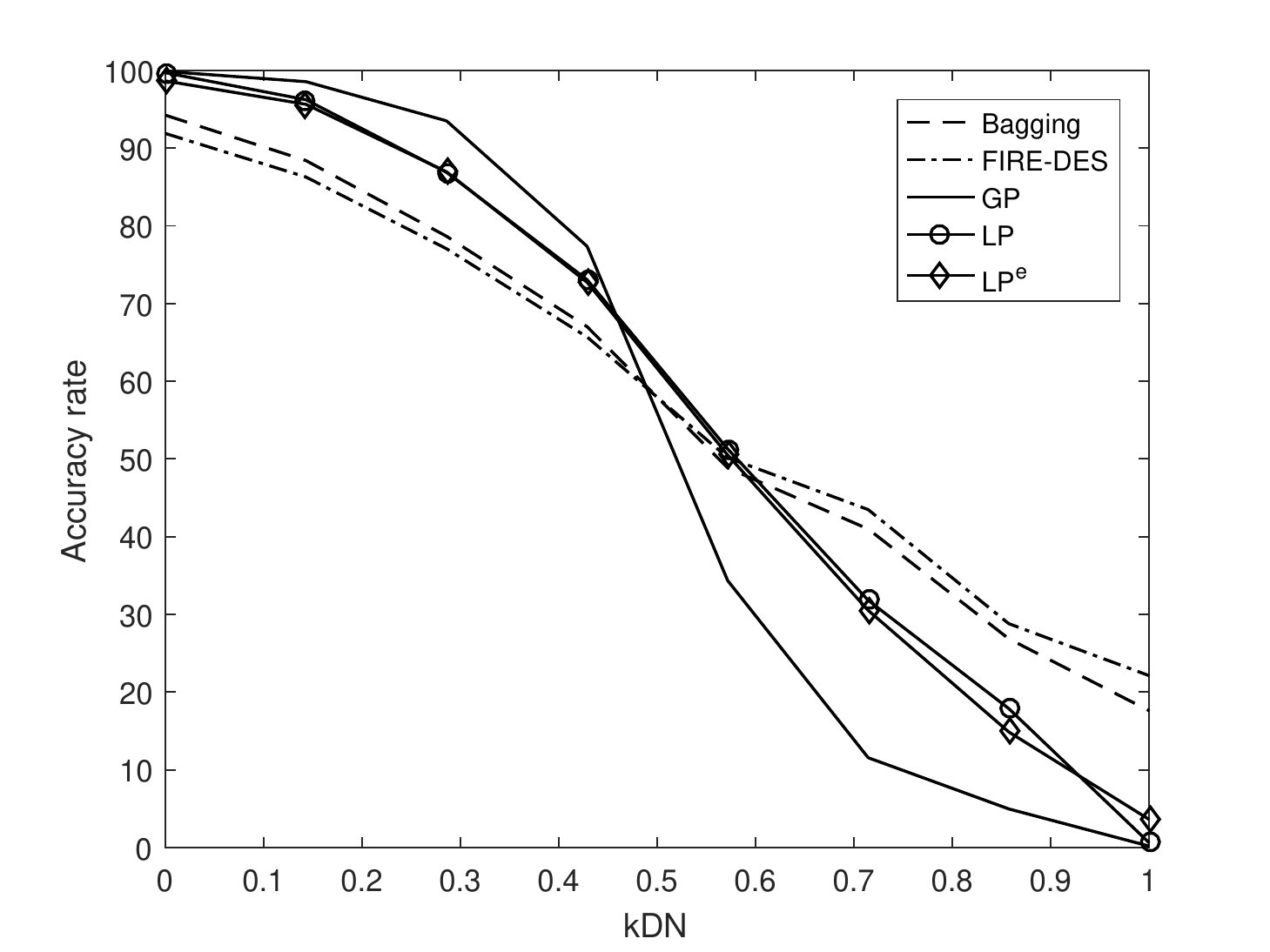}
		}}
		\subfloat[]{
			\includegraphics[scale=0.56]{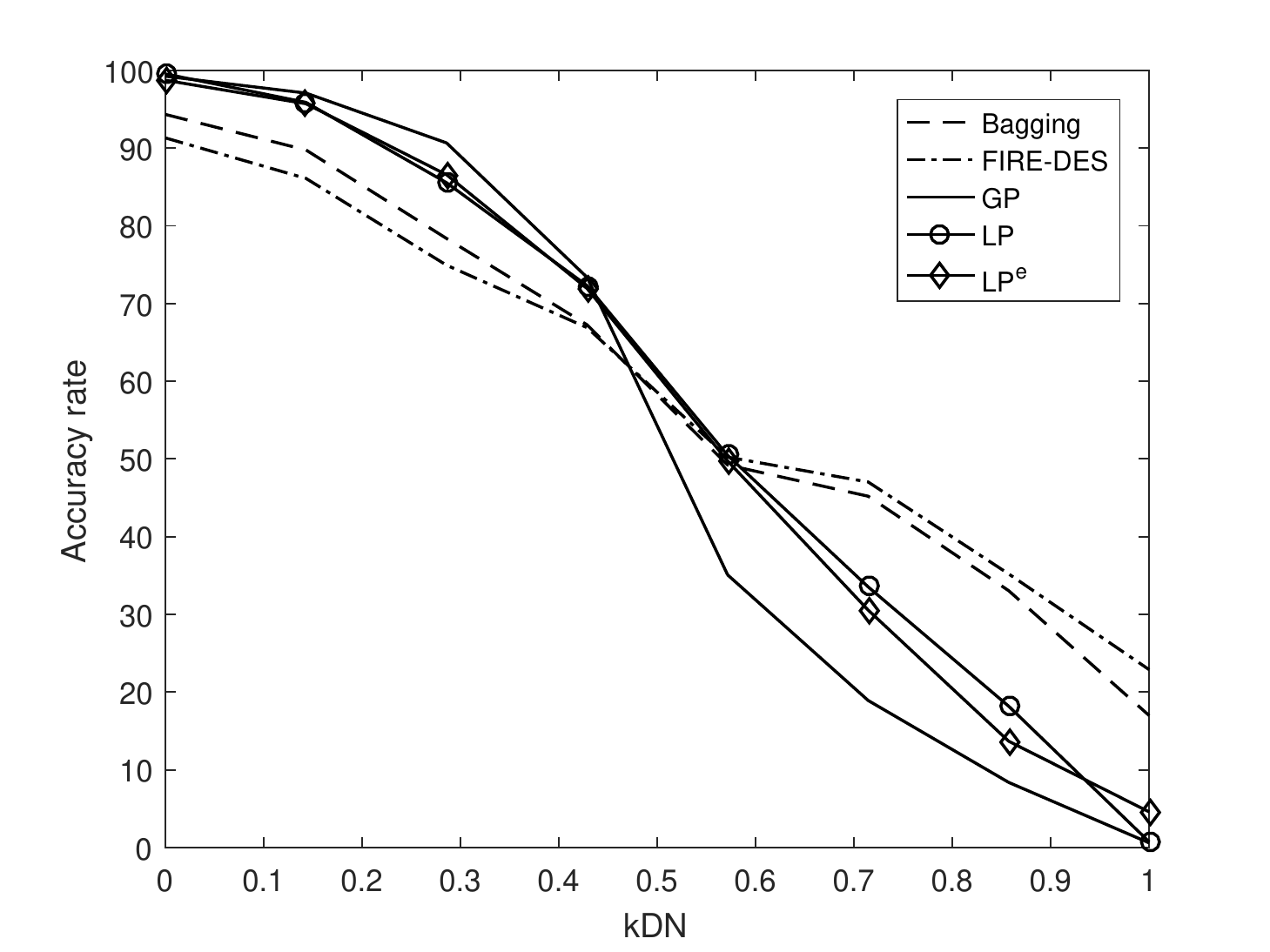}
		}
		\caption{Mean accuracy rate using (a) OLA, (b) LCA and (c) MCB with the Bagging, the FIRE-DES, the $GP$, the $LP$ and the $LP^{e}$ configurations for each group of kDN value, for all datasets from Table \ref{table:datasets}.}
		\label{fig:acc-rate-kdn}
\end{figure*}

 To analyze the distribution of error by hardness value using the three DCS techniques, Figure \ref{fig:acc-rate-kdn} shows the accuracy rate by true kDN value for the five selected configurations. 
Two behaviors can be observed in this analysis. 
The configurations that use Bagging for pool generation misclassify more of the easier instances ($kDN \leq 0.5$) than the proposed configurations, though they correctly classify more of the harder ones ($kDN > 0.5$). 
It can be observed that the gain in performance for the very hard instances is accompanied by a considerable decrease in performance over the very easy ones. 
This behaviour suggests a slight overfitting of the classifiers, which is difficult to adjust for these techniques. 
Since hard instances are usually more scarce than easy ones, it is understandable that the proposed method performs better overall in comparison to the Bagging-generated configurations. 

\par On the other hand, in comparison with the GP configuration, the proposed technique was able to increase the percentage of correctly classified instances with $kDN > 0.5$ with little effect on the performance over the easier ones ($kDN \leq 0.5$). 
This behavior suggests that, not only does the use of local pools is advantageous for hard instances, the RoC evaluation performed on the online phase of the proposed technique allows for a more controlled delimitation of where the classifiers should overfit or not. 

\subsubsection{Discussion}
\label{sec:disc}

\par
\begin{figure*}[!htb]
		\centering
		\centerline{
		\subfloat[]{
			\label{fig:lim-nn-2c}
			\includegraphics[width=0.45\textwidth]{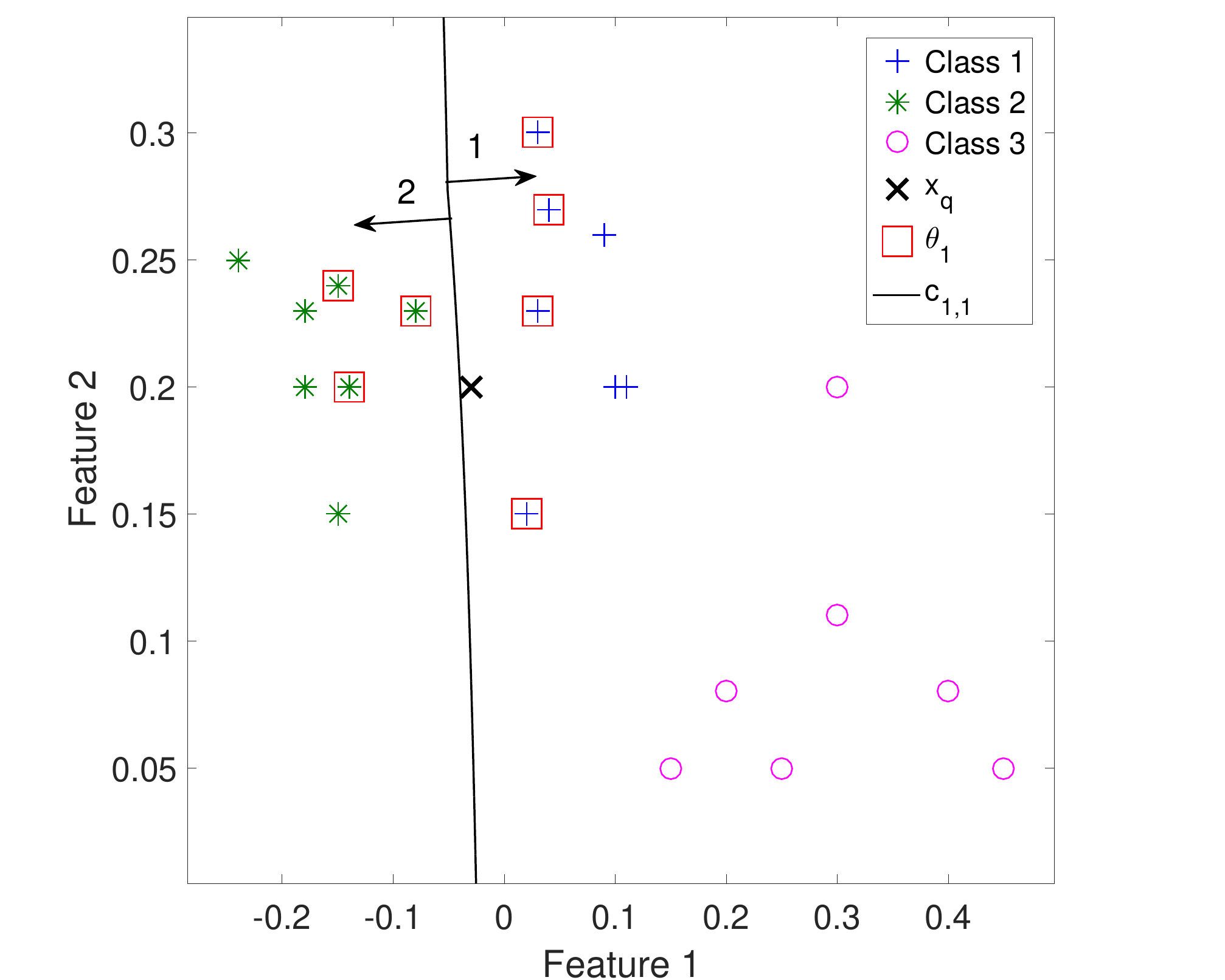}
		}
		\subfloat[]{
			\label{fig:lim-nne-2c}
			\includegraphics[width=0.45\textwidth]{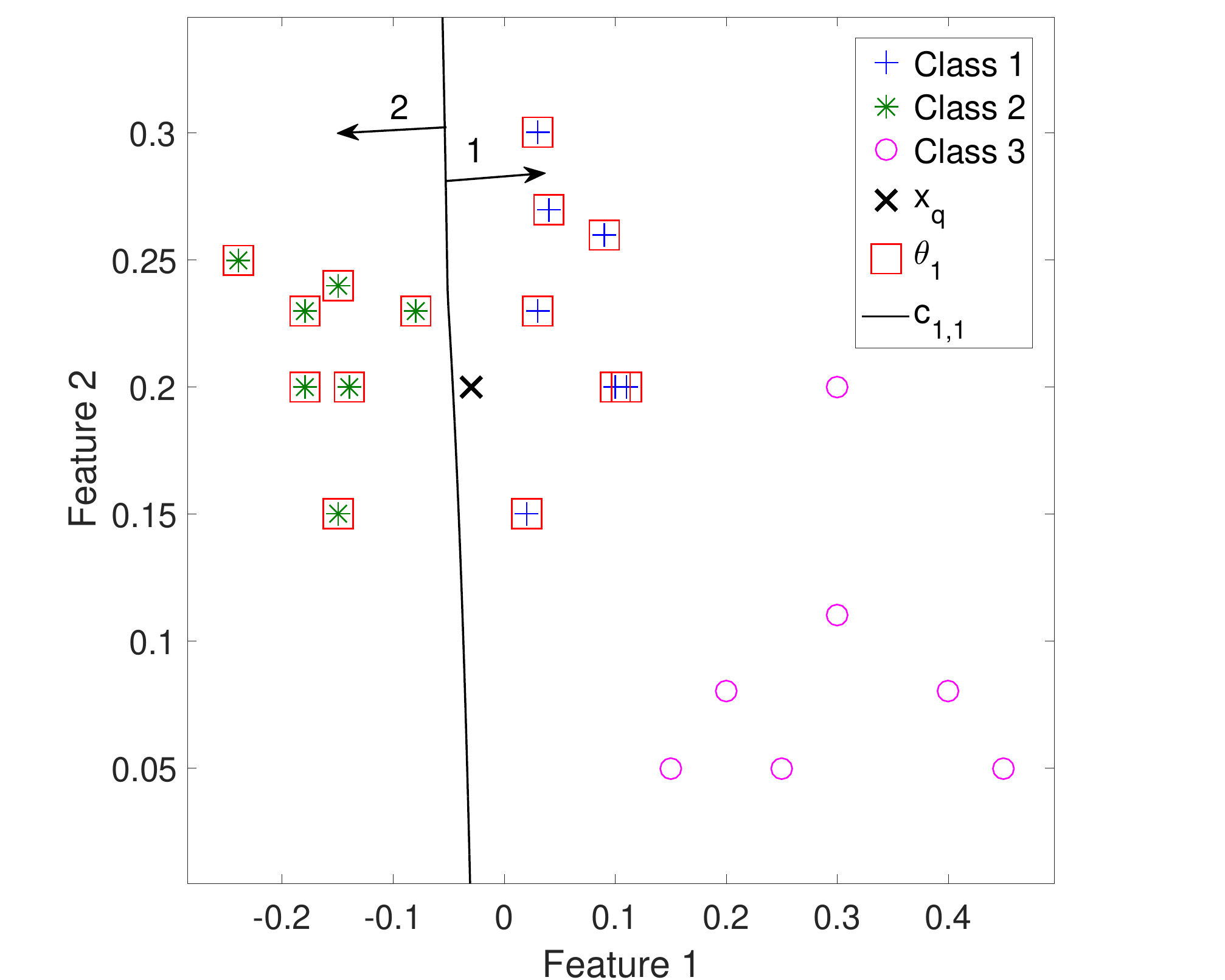}
		}}		
		\centerline{
		\subfloat[]{
		\label{fig:lim-nn-3c}
			\includegraphics[width=0.45\textwidth]{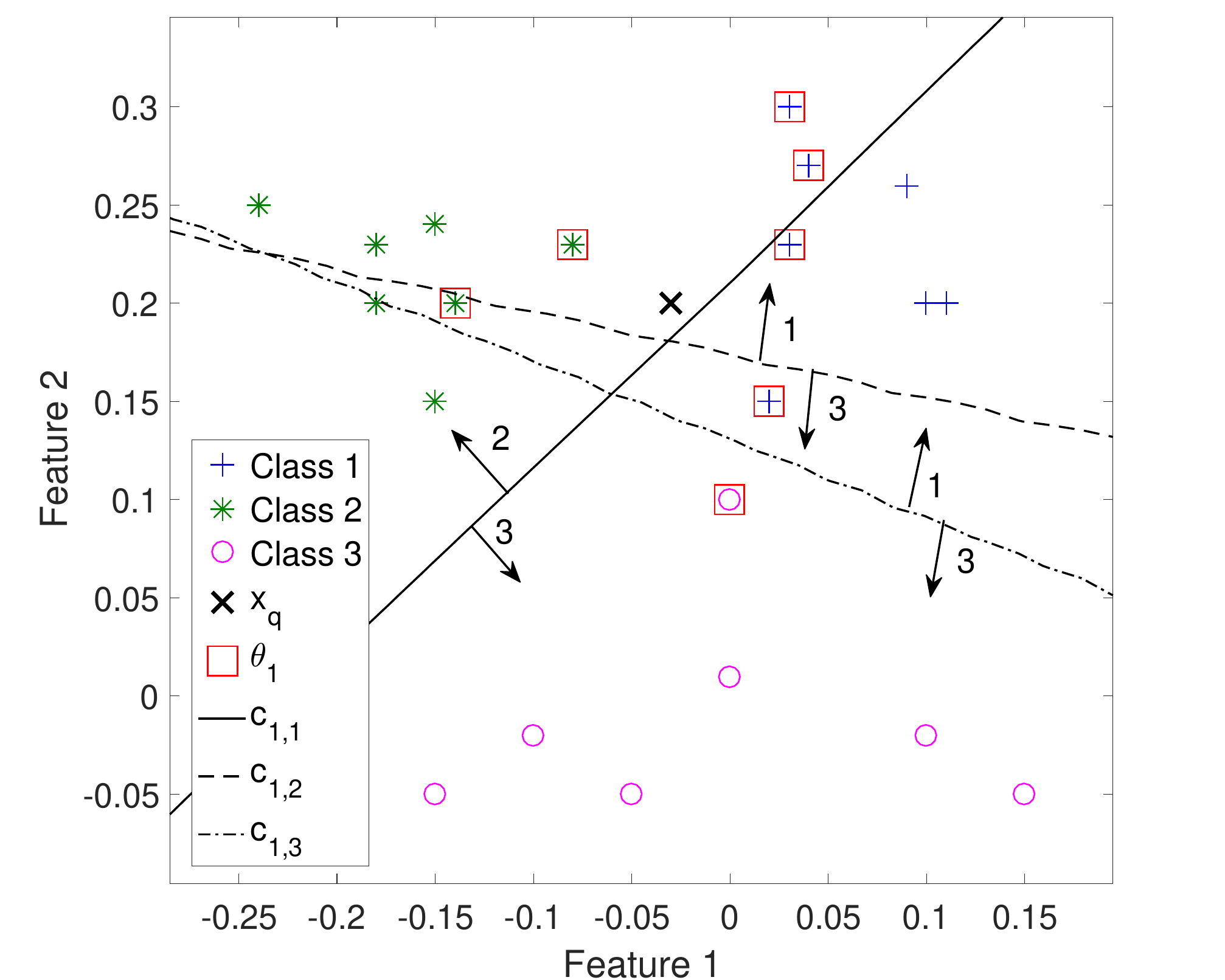}
		}
		\subfloat[]{
		\label{fig:lim-nne-3c}
			\includegraphics[width=0.45\textwidth]{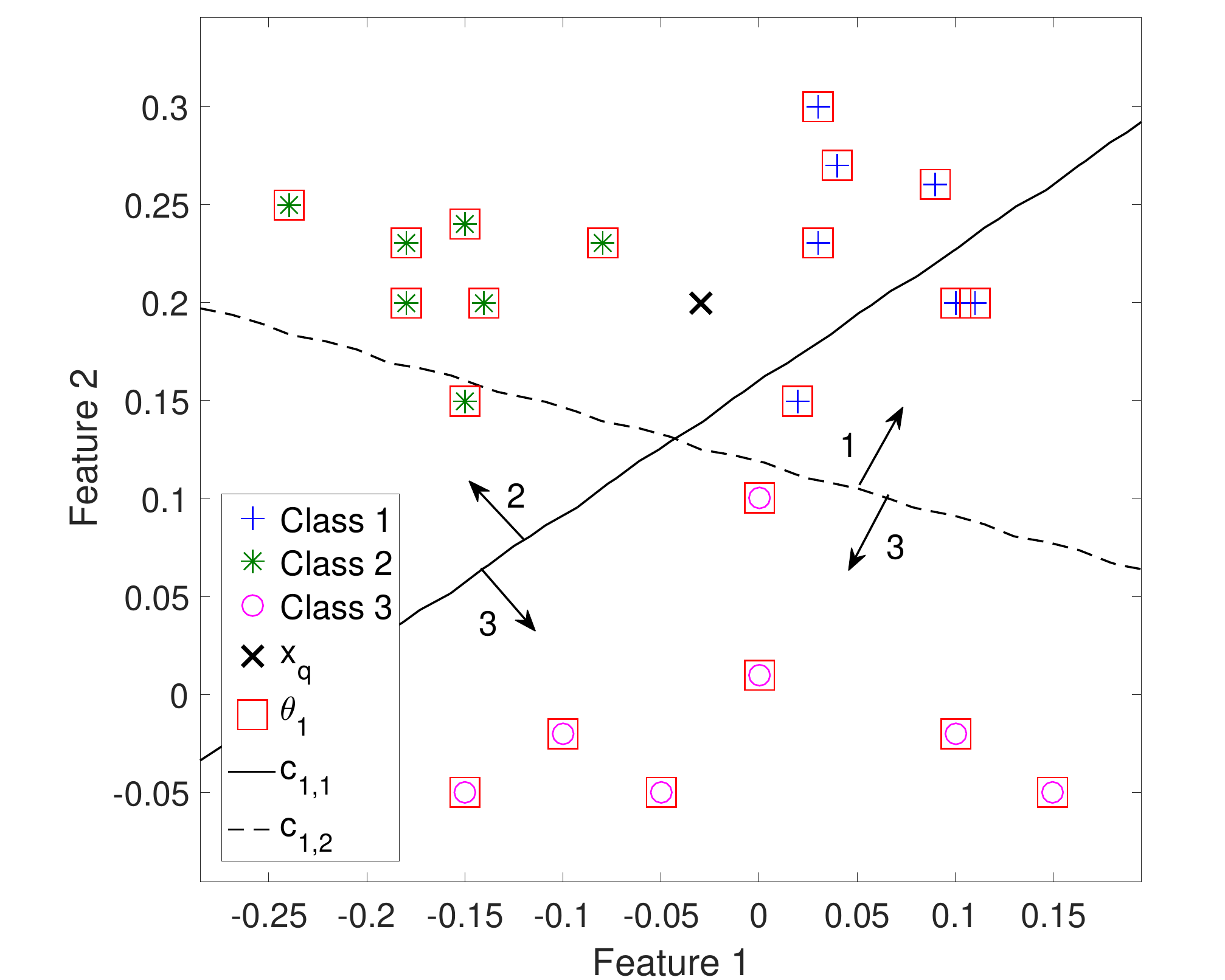}
		}}		
\caption{Example of pool generation for multi-class problems. 
In all scenarios, $x_{q}$ belongs to Class 1. 
In (a) and (c), the query instance's ($\mathbf{x_{q}}$) neighborhood $\theta_{1}$ was obtained using k-NN with $k_{1} = 7$.   
In (b) and (d), $\theta_{1}$ was obtained using a version of k-NNE with $k_{1} = 7$ as well.
These neighborhoods were used as input to the SGH method, which yielded the corresponding subpool of classifiers depicted in the images.}
\label{fig:lim}
\end{figure*}

 From Table \ref{table:acc-rate-dcs}, it can be observed that the two evaluated configurations of the proposed method yielded quite distinct results: the $LP$ configuration always surpassed, by far most of the times, the $LP^{e}$ configuration for the multi-class problems for both DCS techniques. 
The reason for this difference in performance lies in the neighborhood selection schemes used in the online phase of the proposed method, as it can be observed in Figure \ref{fig:lim}, in which two multi-class toy problems are  \looseness=-1 depicted. 

\par In Figure \ref{fig:lim-nn-2c}, the neighborhood $\theta_{1}$ of the query instance $\mathbf{x_{q}}$ was obtained using the regular k-NN rule. It can be observed that, since the border contains only two classes (Class 1 and Class 2), this is also the case for all two-class problems. 
Therefore, the SGH method, which generates only two-class classifiers, returns a pool with only one classifier ($c_{1,1}$) that cover the entire neighborhood $\theta_{1}$. 
Figure \ref{fig:lim-nne-2c} shows the same scenario, but with $\theta_{1}$ being obtained using the version of k-NNE used in this work, which returns the same amount of neighboring instances for all classes in the original k-NN neighborhood. 
That is, the instances from classes too far from the query sample are not included in this method, as Figure \ref{fig:lim-nne-2c} shows. 
The generated pool also contains only one classifier ($c_{1,1}$) that cover the instances in $\theta_{1}$. 
In both presented cases, the DCS technique would select the correct classifier for this query sample, which belongs to Class 1, though the classifier from Figure \ref{fig:lim-nne-2c} seems better adjusted than the one from Figure \ref{fig:lim-nn-2c}.

\par On the other hand, Figure \ref{fig:lim-nn-3c} shows a similar situation, but with Class 3 much closer to the other two classes. 
In this case, the neighborhood $\theta_{1}$ returned by k-NN contains instances from the three classes in the problem. 
Since the SGH method only generates two-class classifiers, the coverage of $\theta_{1}$ is incomplete. 
This is due to the fact that the most distant class in the input set is selected more frequently to draw the hyperplanes. 
It can be observed in Figure \ref{fig:lim-nn-3c} that Class 3, which is the farthest class and thus the least relevant one, is much better covered, with all classifiers recognizing it, than the other two classes. 
In fact, there is not one classifier that separates Class 1 from Class 2 in the generated pool. 
However, since the DCS technique evaluates the classifiers competence over $\theta_{1}$ in the proposed technique, Class 3 only possesses one instance, therefore its weight is much smaller than the remaining two classes in the classifiers' score. 
That way, the classifier $c_{1,3}$ would be selected by OLA, for instance, which would yield the correct label of $\mathbf{x_{q}}$. 

\par Figure \ref{fig:lim-nne-3c} depicts the same scenario from Figure \ref{fig:lim-nn-3c}, but with $\theta_{1}$ obtained using k-NNE. 
Since the original k-NN neighborhood already contained an instance from Class 3, this class is also included in $\theta_{1}$. 
This leads to the neighborhood containing $k_{1} = 7$ instances of each of the three classes of the problem. 
The SGH method generates then two classifiers ($c_{1,1}$ and $c_{1,2}$), and, as in the previous case, the most distant and least relevant class (Class 3) is favoured by the method, since all classifiers recognize it. 
The other two classes, which are closer to $\mathbf{x_{q}}$, do not have a classifier in this subpool to distinguish among themselves. 
However, as opposed to the previous case, the amount of instances of the farthest class is the same as the other two classes, which makes its as relevant as the closer classes for the DCS techniques, since the classifiers are evaluated over the entire $\theta_{1}$. 
In this example, as both classifiers correctly label two out of three classes in the neighborhood, the DCS technique would choose one of them randomly, which would in turn fairly degrade the performance of the system.

\par Therefore, a better approach for multi-class problems is to use the $LP$, which evaluates over the original neighborhood and is likely to give less weight to less relevant classes in the border region. 
Hence, the $LP^{mc}$ column in Table \ref{table:acc-rate-dcs} shows the result of the combined $LP^{e}$ and $LP$ configurations, in which the k-NNE is used for 2-class problems and the k-NN for the multi-class problems. 
It can be observed from the Wilcoxon rows that this scheme is significantly better than Bagging, FIRE-DES and the $GP$ configurations for all DCS techniques. 
The Friedman test was also performed for the three DCS techniques and the resulting average rank of the configurations can be observed in Table \ref{table:acc-rate-dcs}.
The $LP^{mc}$ configuration also obtained the highest average rank in this test. 

\par
\begin{figure*}[!htbp]
		\centering
		\subfloat[]{
			\includegraphics[scale=0.45]{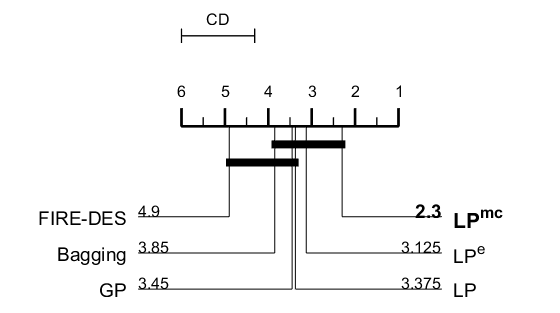}
		}
		\subfloat[]{
			\includegraphics[scale=0.45]{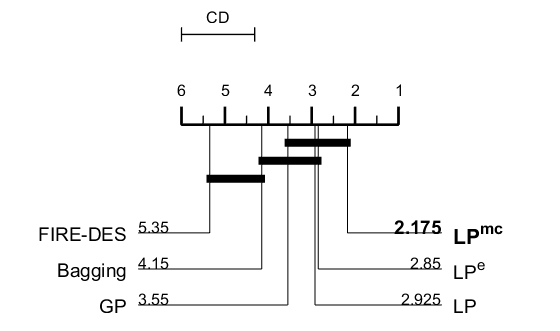}
		}
		\\
		\subfloat[]{
			\includegraphics[scale=0.45]{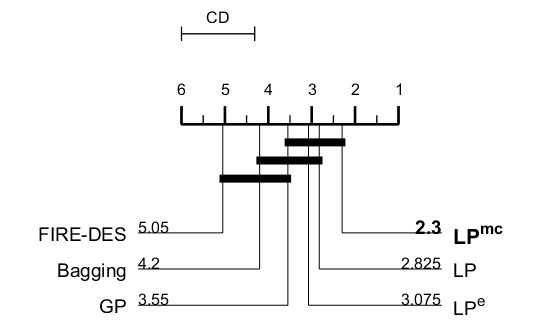}
		}
		\caption{Critical difference diagram representing the results of a post-hoc Bonferroni-Dunn test on the accuracy rates of the methods from Table \ref{table:acc-rate-dcs} for (a) OLA, (b) LCA and (c) MCB. 
The calculated critical difference value was $CD = 1.6861$.
The values near the methods' labels indicate their average rank. 
Statistically similar methods are connected by an horizontal line, while statistically different ones are disconnected. 
}
\label{fig:cd}
\end{figure*}

 Since the resulting p-values of the Friedman test indicate that there is a significant difference between the performances of the evaluated configurations for all three DCS techniques, a post-hoc Bonferroni-Dunn test was performed afterwards to obtain a pairwise comparison between the configurations. 
Two configurations are significantly different if the difference between their average rank is greater than the critical difference $CD$. 
The critical difference diagrams \cite{demvsar2006statistical} depicted in Figure \ref{fig:cd}, show the results of the post-hoc tests for each DCS technique. 
The configurations with no significant difference are connected by a bar, whilst significantly different ones are not intersected in the diagram.

\par The critical difference value for this test was $CD = 1.6861$. 
It can be observed that the $LP^{mc}$ configuration is significantly superior to the FIRE-DES scheme for all DCS techniques, which suggests that generating locally accurate classifiers is a better strategy than pruning a large pool in search of such classifiers for instances in difficult regions, at least for balanced and moderately imbalanced problems, as used in the experiments. 
The $LP^{mc}$ configuration also yielded a significant increase in performance in comparison with Bagging for all DCS techniques but OLA. 
Since the $LP^{mc}$ configuration partially solves the limitations of the SGH method and performed the best for all three DCS techniques, it is used in the following comparative study with the state-of-the-art models in the literature.


\subsection{Comparison with State-of-the-art Models}
\label{sec:comp-models}

\par A comparative study on the performances of the proposed method and nine state-of-the-art models is presented in this section. 
The purpose of this study is to assess whether the proposed method achieves similar recognition rates to the most well-performing models in the literature, considering single models and other MCS.

\par Five static state-of-the-art classifiers feature in the comparative study: the Multi-layer Perceptron (MLP) model with the Levenberg-Marquadt algorithm, the Support Vector Machine (SVM) model with a Gaussian Kernel, the Random Forest (RF) \cite{randomforest} classifier, the AdaBoost \cite{adaboost} classifier, and the Oblique Decision Tree (DT) ensemble  \cite{zhang2017benchmarking}. 
These models belong to the best performing families of classifiers, according to \cite{fernandez2014we}, and also were among the best performing models in \cite{zhang2017benchmarking}. 
Since static models do not need a DSEL dataset, it was used as a validation set for the MLP classifier and added to the training set for the remaining  \looseness=-1 static models. 

\par Furthermore, four state-of-the-art DS techniques were also included in this analysis: the Randomized Reference Classifier (RRC) \cite{rrc}, the META-DES \cite{metades}, the META-DES.Oracle (META-DES.O) \cite{cruz2017meta} and the FIRE-KNORA-U (F-KNU) \cite{dayvid}. 
The latter consists of the FIRE-DES framework coupled with the K-Nearest Oracles Union (KNORA-U) \cite{knora} selection technique. 
The same Bagging-generated pool of 100 Perceptrons used in the previous section was used for these techniques. 
The region of competence size was also set to 7, as in the previous experiments.

\par All classifiers were evaluated using the MATLAB PRTOOLS toolbox \cite{prtools}, and the parameters of the static models were set to the default. 
Moreover, the proposed method's configuration used for comparison in this analysis was the $LP^{mc}$ with LCA, since it yielded the highest mean accuracy rate in the previous experimental study. 

\par Table \ref{table:acc-static} shows the mean accuracy rate of the static classification models and the proposed method, for all datasets from Table \ref{table:datasets}. 
It can be observed that the proposed configuration yielded a higher overall accuracy rate than all static models but the Oblique DS ensemble. 
It is important to remember, though, that no fine tuning of parameters was performed, and that stands for all static models as well as the proposed technique. 
A Wilcoxon signed-rank test was also performed over the results (row Wilcoxon), and it can be observed that the performance of the proposed configuration was significantly superior to the MLP and SVM models. 

\begin{table}[!htb]
\caption{Mean and standard deviation of the accuracy rate of MLP, SVM, RF, AdaBoost, Oblique DT ensemble and the $LP^{mc}$ configuration.
The row \textit{Wilcoxon} shows the result of a Wilcoxon signed rank test over the mean accuracy rates of the proposed configuration and each of the remaining methods. 
The significance level was $\alpha = 0.05$, and the symbols $+$, $-$ and $\sim$ indicate if the compared method is significantly superior, inferior or not significantly different from the proposed method, respectively. 
Best results are in bold.}
\label{table:acc-static}
\scriptsize
\centerline{
\scalebox{0.9}{
\begin{tabular}{|c|ccccc|c|}
\hline
\textbf{Dataset}  & \textbf{MLP} & \textbf{SVM} & \textbf{RF \cite{randomforest}} & \textbf{AdaBoost \cite{adaboost}} &	\textbf{Oblique DT ens. \cite{zhang2017benchmarking}} & $\mathbf{LP^{mc}}$ \\ \hline
Adult 		 &        {82.83 (3.61)} &        {73.99 (2.88)} &        {67.83 (8.34)}  & 	   {88.44 (2.05)}  & \textbf{88.76 (1.43)}   &	         {87.11 (2.40)} \\
Blood 		 &        {78.11 (1.63)} & \textbf{78.14 (1.08)} &        {72.07 (3.00)}  &        {77.29 (1.74)}  & 	    {77.23 (0.95)}   &	         {76.89 (1.67)} \\
CTG 		 &        {89.52 (1.46)} &        {84.51 (0.53)} &        {91.20 (0.94)}  &        {92.42 (1.69)}  & \textbf{93.96 (0.68)}   &	         {92.22 (1.10)} \\
Faults 	     &        {68.79 (4.45)} &        {49.59 (0.30)} &        {70.86 (2.35)}  &        {54.66 (2.43)}  & \textbf{77.04 (2.06)}   &	         {73.20 (1.22)} \\
German 	     &        {70.24 (3.08)} &        {70.12 (0.19)} &        {38.28 (6.78)}  &        {74.82 (1.94)}  & \textbf{75.18 (1.88)}   &	         {74.08 (1.84)} \\
Glass 		 &        {61.32 (7.20)} &        {62.83 (3.47)} &        {71.04 (4.59)}  &        {48.30 (5.98)}  & \textbf{78.58 (3.14)}   &	         {67.45 (2.73)} \\
Haberman 	 &        {70.72 (2.33)} & \textbf{74.34 (2.02)} &        {69.41 (4.24)}  &        {68.62 (3.48)}  & 	    {72.04 (3.69)}   &	         {72.11 (2.12)} \\
Heart 		 &        {75.15 (5.02)} &        {58.82 (1.79)} &        {62.94 (6.45)}  &        {84.26 (4.80)}  & \textbf{86.32 (3.20)}   &	         {83.09 (3.32)} \\
Ionosphere   &        {87.84 (4.30)} &        {68.07 (2.24)} &        {93.81 (2.16)}  & \textbf{96.36 (1.50)}  & 	    {95.17 (1.32)}   &	         {92.44 (2.56)} \\
Laryngeal1   &        {79.53 (6.16)} &        {76.23 (3.26)} &        {81.79 (4.55)}  &        {81.60 (3.29)}  & \textbf{85.94 (2.90)}   &	         {80.57 (5.87)} \\
Laryngeal3   &        {67.87 (5.36)} &        {68.88 (2.78)} &        {72.53 (2.93)}  &        {70.11 (2.88)}  & \textbf{73.54 (3.25)}   &	         {73.48 (2.48)} \\
Liver 		 &        {68.31 (4.80)} &        {64.53 (3.82)} &        {69.30 (4.52)}  &        {69.71 (3.49)}  & \textbf{71.05 (4.11)}   &	         {66.98 (1.79)} \\
Mammographic &        {84.45 (2.94)} &        {83.46 (2.87)} &        {59.28 (8.08)}  &        {80.70 (2.25)}  & \textbf{84.57 (1.45)}   &	         {82.57 (1.77)} \\
Monk2 		 &        {99.49 (1.06)} &        {95.28 (1.37)} &        {89.54 (2.93)}  & \textbf{100.0 (0.00)}  & 	    {96.90 (1.28)}   &	         {94.07 (0.76)} \\
Phoneme 	 &        {83.51 (1.07)} &        {87.43 (0.43)} &        {90.34 (0.49)}  & \textbf{90.66 (0.55)}  & 	    {89.65 (0.52)}   &	         {86.62 (0.69)} \\
Pima 		 &        {74.35 (3.63)} &        {71.46 (2.20)} &        {76.25 (2.67)}  &        {75.42 (1.94)}  & \textbf{77.21 (1.37)}   &	         {76.74 (2.24)} \\
Sonar 		 &        {78.08 (5.49)} &        {81.15 (4.02)} &        {83.75 (5.38)}  & \textbf{85.38 (5.16)}  & 	    {83.56 (5.56)}   &	         {76.35 (3.64)} \\
Vehicle 	 & \textbf{76.91 (2.38)} &        {63.63 (2.85)} &        {73.77 (2.21)}  &        {68.94 (3.30)}  & 	    {74.27 (2.43)}   &	         {73.51 (1.64)} \\
Vertebral 	 &        {81.03 (4.15)} &        {85.00 (2.76)} &        {85.45 (3.63)}  &        {84.04 (2.48)}  & 	    {85.96 (3.80)}   &	  \textbf{86.47 (2.65)} \\
Weaning 	 &        {78.42 (5.20)} &        {69.74 (6.80)} &        {86.97 (2.73)}  & \textbf{87.76 (1.87)}  & 	    {86.38 (2.06)}   &	         {85.66 (2.37)} \\ \hline
\textbf{Average}  &    77.82         &    73.36              &   75.32                & 78.97                  &  \textbf{82.66}	     &    	80.08 \\ \hline
\textbf{Wilcoxon} &      -           &     -                 &     $\sim$             &   $\sim$ 	           &	     +		         &         n/a 			 \\ \hline
\end{tabular}}
}                                                                                                    
\end{table}

\par Table \ref{table:acc-des} shows the mean accuracy rate of the four state-of-the-art DES techniques and the proposed configuration.
It can be observed that the proposed technique obtained a greater mean accuracy rate in comparison with three of the four DES techniques. 
Moreover, according to a Wilcoxon signed-rank test with significance level of $\alpha = 0.05$ (Wilcoxon row), the proposed configuration obtained a significantly superior performance to the F-KNU technique. 
In comparison to the remaining DES techniques, the proposed method yielded a statistically similar \looseness=-1 performance.

\begin{table}[!htb]
\centering
\caption{Mean and standard deviation of the accuracy rate of the Randomized Reference Classifier (RRC), the META-DES, the META-DES.Oracle (META-DES.O), the FIRE-KNORA-U (F-KNU) and the $LP^{mc}$ configuration.
The row \textit{Wilcoxon} shows the result of a Wilcoxon signed rank test over the mean accuracy rates of the proposed configuration and each of the remaining methods. 
The significance level was $\alpha = 0.05$, and the symbols $+$, $-$ and $\sim$ indicate if the compared method is significantly superior, inferior or not significantly different from the proposed method, respectively. 
Best results \looseness=-1 are in bold.}
\label{table:acc-des}
\scriptsize
\scalebox{0.9}{
\begin{tabular}{|c|cccc|c|}
\hline
\textbf{Dataset}  & \textbf{RRC \cite{rrc}} & \textbf{META-DES \cite{metades}} & \textbf{META-DES.O \cite{cruz2017meta}} & \textbf{F-KNU \cite{dayvid}} & $\mathbf{LP^{mc}}$ \\ \hline
Adult 		 & \textbf{88.87 (2.27)} &        {84.45 (6.41)} &        {80.78 (7.33)} &        {84.86 (2.85)} &         {87.11 (2.40)}  \\
Blood 		 &        {76.30 (1.41)} & \textbf{77.98 (1.81)} & \textbf{77.98 (1.20)} &        {64.87 (2.56)} &         {76.89 (1.67)}  \\
CTG 		 &        {89.41 (0.71)} &        {91.49 (0.66)} &        {92.10 (1.12)} &        {88.34 (0.94)} &  \textbf{92.22 (1.10)}  \\
Faults 	     &        {70.35 (1.06)} & \textbf{73.58 (1.57)} &        {73.39 (1.61)} &        {67.44 (1.74)} &         {73.20 (1.22)}  \\
German 	     & \textbf{76.42 (2.14)} &        {75.70 (1.69)} &        {74.76 (1.82)} &        {70.30 (1.01)} &         {74.08 (1.84)}  \\
Glass 		 &        {65.19 (4.39)} & \textbf{70.19 (3.44)} &        {69.53 (5.17)} &        {65.47 (3.73)} &         {67.45 (2.73)}  \\
Haberman 	 &        {74.08 (1.71)} &        {73.82 (5.79)} & \textbf{75.26 (2.36)} &        {57.24 (4.91)} &         {72.11 (2.12)}  \\
Heart 		 &        {86.62 (1.42)} & \textbf{86.47 (3.53)} &        {81.10 (4.35)} &        {85.51 (2.35)} &         {83.09 (3.32)}  \\
Ionosphere   &        {88.75 (2.24)} &        {88.47 (2.19)} &        {85.17 (5.10)} &        {88.35 (1.91)} &  \textbf{92.44 (2.56)}  \\
Laryngeal1   & \textbf{85.19 (3.08)} &        {80.28 (4.95)} &        {78.21 (6.14)} &        {80.94 (5.30)} &         {80.57 (5.87)}  \\
Laryngeal3   & \textbf{74.27 (3.40)} &        {73.54 (3.31)} &        {73.48 (3.63)} &        {66.24 (4.13)} &         {73.48 (2.48)}  \\
Liver 		 &        {65.81 (4.34)} & \textbf{68.95 (3.25)} &        {67.21 (3.57)} &        {60.58 (3.99)} &         {66.98 (1.79)}  \\
Mammographic & \textbf{85.77 (2.08)} &        {73.13 (15.9)} &        {72.36 (18.2)} &        {78.53 (2.58)} &         {82.57 (1.77)}  \\
Monk2 		 &        {85.23 (2.71)} & \textbf{96.76 (1.22)} & \textbf{96.76 (1.22)} &        {85.32 (2.57)} &         {94.07 (0.76)}  \\
Phoneme 	 &        {74.08 (1.57)} &        {87.49 (0.82)} & \textbf{89.34 (0.69)} &        {73.89 (1.61)} &         {86.62 (0.69)}  \\
Pima 		 &        {76.95 (2.33)} & \textbf{77.40 (1.94)} &        {76.98 (2.49)} &        {67.29 (2.73)} &         {76.74 (2.24)}  \\
Sonar 		 &        {80.96 (2.92)} &        {82.98 (3.38)} & \textbf{83.75 (2.96)} &        {81.35 (2.93)} &         {76.35 (3.64)}  \\
Vehicle 	 &        {75.40 (1.97)} &        {75.94 (2.21)} &        {75.12 (2.11)} & \textbf{76.56 (1.84)} &         {73.51 (1.64)}  \\
Vertebral 	 &        {85.19 (3.01)} &        {86.22 (3.45)} &        {85.77 (2.79)} & \textbf{87.12 (3.98)} &         {86.47 (2.65)}  \\
Weaning 	 &        {81.84 (3.27)} &        {84.28 (3.37)} &        {82.89 (3.28)} &        {81.12 (3.29)} &  \textbf{85.66 (2.37)}  \\ \hline
\textbf{Average}  &    79.33       &   \textbf{80.45}      &     79.59        &  75.50             &   80.08          \\ \hline
\textbf{Wilcoxon} &    $\sim$      &   $\sim$      &    $\sim$          &     -              &      n/a              \\ \hline
\end{tabular}
}
\end{table}

\subsection{Computational Complexity}
\label{sec:cost}

In this section, we analyze the computational complexity of the proposed method versus the complexity of different dynamic selection techniques. We use the Big-O ($\mathcal{O}$) and Big-Omega ($\Omega$) notations~\cite{knuth1976big} to represent the worst and best running time scenarios, respectively.

The analysis is made taking into account the dataset size $n$, the classifiers pool size $m$, the dimensionality of the dataset $d$ and the neighborhood size $k$. For the sake of simplicity, we consider that all base classifiers are from the same model (Perceptron), and the cost associated with the training of the base classifier is denoted by $l$.

\subsubsection{Complexity in Memorization}

The proposed method does not train a pool of classifiers. During the memorization phase, it only calculates the hardness level of each instance. The basis of the hardness calculation is the k-NN method, which is known to be linear with the size of the dataset and the dimensionality, $\mathcal{O}(nd)$. Since the k-nearest neighbors of each training sample needs to be calculated, the final cost grows quadratically with the dataset size $n$, and linearly with the dimensionality $d$, $\mathcal{O}(n^{2}d)$.

The training steps for DS techniques involves the generation of the pool and pre-processing the base classifiers outputs for each instance in the dataset. In this work, the Bagging algorithm was used to generate a pool composed of $m$ base classifiers. So, the cost is $m$ times the cost associated to train the base classifier, $\mathcal{O}(ml)$. Then, the outputs of each base classifier for each training sample is pre-calculated, giving a total of $\mathcal{O}(mn)$ calculations. Thus, the final cost associated with DS techniques is of order $\mathcal{O}(ml + mn)$. 

Another important aspect is the memory cost associated with each technique. The proposed method requires only storing the hardness level of each sample, $\mathcal{O}(n)$. For the DS techniques, they keep a matrix with the outputs of each base classifiers for each data point, $\mathcal{O}(mn)$, as well as the parameters (weights) of each base classifier. The number of weights in a Perceptron classifier is the number of dimensions in the dataset. Thus, having a final memory cost of order $\mathcal{O}(nm + md)$.

\subsubsection{Complexity in Generalization}

The analysis in generalization is conducted by dividing each algorithm into three steps according to DS the taxonomy proposed in~\cite{cruz2017dynamic}: definition of the region of competence, estimation of the competence level and the classification. Then, the computational complexity of each part is analyzed individually. Table~\ref{table:complexity} shows the computational complexity using the Big O notation for each dynamic selection algorithm. 

\begin{table}[h!] 
	\centering 
	\caption{Computational cost in generalization of each method studied.} 
	\label{table:complexity}  
	
	\begin{tabular}{l| r r r}  
		\hline  
		Algorithm & Region of competence & Competence estimation & Classification \\
		
		\hline  
		\textbf{Proposed} & $\mathcal{O}(nd)$ 					   & $\mathcal{O}(m''dkm')$ 	&   $\mathcal{O}(m'd)$					  \\
		OLA \cite{ola}& $\mathcal{O}(nd)$ 					       & $\mathcal{O}(mk)$ 	& 	$\mathcal{O}(d)$ 	  \\
		LCA \cite{ola}& $\mathcal{O}(nd)$						  	   & $\mathcal{O}(mk)$   &  $\mathcal{O}(d)$ 	  \\
		MCB \cite{mcb}& $\mathcal{O}(nd)$ 					  	   & $\mathcal{O}(mk)$   & 	$\mathcal{O}(d)$ 	  \\
		KNU \cite{dayvid}& $\mathcal{O}(nd)$ 						   & $\mathcal{O}(mk)$   &	$\mathcal{O}(md)$	  \\ 
		RRC \cite{rrc}& $\mathcal{O}(nd)$ 						   & $\mathcal{O}(mn)$  & 	$\mathcal{O}(md)$ 	  \\
		META-DES \cite{metades}& $\mathcal{O}(nd + nm)$			 	   & $\mathcal{O}(2mk)$   & 	$\mathcal{O}(md)$  	  \\
		META-DES.O \cite{cruz2017meta}& $\mathcal{O}(nd + nm)$ 			   & $\mathcal{O}(2mk)$   & 	$\mathcal{O}(md)$  	  \\
		\hline
	\end{tabular}
\end{table}

For the estimation of the region of competence, the proposed method as well as all DS methods studied in this work are based on the k-NN algorithm. Thus, it is of order $\mathcal{O}(nd)$ The only exception is the two versions of the META-DES framework which requires two estimations of the local competences: one in the feature space and another in the decision space. In the case of the decision space, the number of dimensions is equal to the number of base classifiers in the pool, $m$. Hence, the computational complexity estimating the local region is $\mathcal{O}(nd + nm)$ for these techniques.

In the competence level estimation, the cost involved to calculate the competence level of a single base classifier is equal to get its performance on its k-Nearest neighbors $\mathcal{O}(k)$. However, since the competence level needs to be calculated for each expert in the pool, the final complexity is linear with both $m$ and $k$ $\mathcal{O}(mk)$. The only exception is for the Randomized Reference Classifier (RRC) algorithm. This method uses the whole dataset rather than only the k-Nearest Neighbors to calculate the competence level of a base classifier. So, the computational cost of this techniques is of order $\mathcal{O}(nm)$, making this step very costly when dealing with large datasets. 

In contrast, the proposed method generates the local pool in this step rather than only selecting the most competent classifiers. The local pool is generated iteratively using the SGH method, which depends on the dimensionality of the problem to find the center of each class and generate the hyperplane. In each iteration, the SGH method is applied over a different neighborhood, which results in $m''$ classifiers, with $m''$ being less than three, on average. These classifiers are then evaluated over a neighborhood of size $k$, so, for each iteration, $\mathcal{O}(km''d)$. The best classifier produced in each iteration is added to the local pool. Thus, for a local pool size of $m'$ hyperplanes, the method iterates $m'$ times. The cost involved in this part is then $\mathcal{O}(m''dkm')$. 

The classification step using the proposed method is performed by applying the majority voting rule over the $m'$ generated local Perceptron classifiers. Hence, the computational cost involved is of order $\mathcal{O}(m'd)$. 
It is important to observe that the size of the local pool is much smaller than the one generated with Bagging, $m' = 7$ versus $m = 100$. So, since there are fewer base classifiers to evaluate in the proposed method, this step is usually faster in execution when compared to DES techniques. In contrast, the DCS techniques use only one classifier for classification. 
Hence, the computational complexity of this step is equal to the cost of the base classifier alone, $\mathcal{O}(d)$. The DES techniques can select an arbitrary number of base classifiers. However, in the worst case scenario (i.e., when all base classifiers in the pool are selected), it has a complexity of order $\mathcal{O}(md)$, which increases linearly as the number of classifiers in the pool increases. In the best case scenario, when a single classifier is selected, the computational cost is equal to the cost of a single base classifier, $\Omega(d)$. 
	
This analysis explains why the proposed method was much faster in generalization than the DS techniques using a pool of classifiers generated with Bagging, especially when compared with DES techniques. In the best case scenario, when the sample is located in an easy region, there is no need to generate a local pool. So, its cost is equal to simply applying the k-NN classifier, $\Omega(nd)$. In the worst case, the computational cost is the sum of the three steps: $\mathcal{O}(nd + m''dkm' + m'd)$. In contrast, DCS and DES techniques always require the computation of the three steps regardless of whether or not the query sample is located in an easy region.
	
For this reason, the average running time of the proposed method was three times faster than the DCS techniques using the pool generated with Bagging, eight times faster than the KNORA-U and also around 14 times faster than the RRC and META-DES techniques. However, the DCS techniques using the GP were 10 times faster than the proposed method. This can be explained due to its use of even fewer base classifiers. The global pool has, on average, less than 4 base classifiers, which is even less than the number of the base classifiers generated online using the proposed LP method. That reduces considerably the computational cost of applying DS methods.
	
It is important to note that the estimation of the nearest neighbors is usually the most computationally expensive part of the proposed method as well as DS algorithms, especially when dealing with large sample size datasets. However, this cost can be reduced either by using different implementation of the k-NN technique such as kd-trees, which has a computational complexity of $\mathcal{O}(d \; log_{2}n)$, or by reducing the dataset size and dimensionality of the problem through prototype and feature selection techniques, respectively. The other factor that increases the computational complexity of DS techniques is the pool size. However, as reported in~\cite{roy2016meta2}, DS techniques do not require large pool sizes in order to achieve good classification performance. In fact, the best classification accuracy is usually obtained with less than 50 base classifiers in the pool. By reducing the pool size, the computational cost of DS technique can be further reduced, especially for the DES ones.


\section{Conclusion}
\label{sec:concl}

In \cite{mariana}, it was shown that the DCS techniques had difficulty in selecting a competent classifier even though the presence of such a classifier in the pool was assured. 
The generation method used in that work guaranteed an Oracle accuracy of 100\%. 
It was concluded that the Oracle model, being performed globally, did not help in the search for a good pool of classifiers for DCS techniques, because the latter use only local data to select a competent classifier for any given \looseness=-1 instance. 

In this work, an instance hardness analysis was performed in order to draw a correlation between hardness measures and the error rates of DCS techniques. 
Based on that relationship, an online pool generation scheme was proposed with the purpose of increasing the accuracy rates of the instances the DCS techniques had difficulty in labelling. 
The proposed technique involved generating subpools for each identified difficult region in the feature space, so that another, more locally accurate pool could be used for hard instances, in hopes that, by fully covering these regions with a locally specialist pool, it would be easier for the DS techniques to select the best classifiers for these samples. 
The instances deemed ``easy", however, would be classified using a simple \looseness=-1 k-NN rule.

Experiments were performed over 20 public datasets, and two configurations of the proposed scheme were analyzed. 
It was shown that the use of local pools increased the hit rate of the global pool (GP) for most datasets, suggesting the use of such pools indeed helps the DS in selecting the most competent ones for a given query instance.  
The overall performances of both configurations were compared with a pool of 100 Perceptrons generated using Bagging, with an online pruned pool of originally 100 Bagging-generated Perceptrons and with the GP. 
It was observed that a combination of both proposed configurations yields a significantly increase in accuracy rate compared to the other tested methods for the three evaluated DCS techniques, suggesting that, not only do the DCS techniques select the best classifier more frequently, but also the recognition rates of the DCS techniques fairly increase when using a local perspective during generation. 
The choice of which proposed configuration to use is based on the characteristics of each problem, and this selection is necessary due to a limitation in the SGH method. 
%
Furthermore, the proposed technique was compared to nine state-of-the-art classification models, including five static and four DES techniques, and it yielded a significant superior performance to three of the models and a statistically similar performance to five of them.


\par Improvements to the proposed technique may involve developing an automatic scheme for defining the input parameters
and an adaptation for better dealing with multi-class problems. 
Moreover, the impacts of data preprocessing on the performance of DS techniques over imbalanced problems have been analyzed in \cite{anandarupimbal}. 
Thus, a study on the robustness of the proposed method to class imbalance and the suitability of using data preprocessing techniques for imbalance learning may also be performed in future works.
Several recent works have also delved into neighborhood characterization and RoC definition for improving the recognition rates of DS techniques, yielding good results specially for imbalanced problems \cite{firedes++,knorabbi}. 
So, a study on neighborhood estimation and characterization with respect to the proposed technique would also be an interesting future research.


\section*{Acknowledgements}
The authors would like to thank Brazilian agencies: CAPES (Coordena\c{c}\~{a}o de Aperfei\c{c}oamento de Pessoal de N\'{i}vel Superior), CNPq (Conselho Nacional de Desenvolvimento Cient\'{i}fico e Tecnol\'{o}gico) and FACEPE (Funda\c{c}\~{a}o de Amparo \`{a} Ci\^{e}ncia e Tecnologia de Pernambuco).

\section*{References}
\looseness=-1
\bibliography{references}

\end{document}